%% file: main.tex
\documentclass[manuscript,screen]{acmart}

\AtBeginDocument{%
  }

\usepackage{multirow}
\usepackage{tabularx}
\usepackage{colortbl}
\usepackage{tcolorbox}
\usepackage{pifont}
\tcbuselibrary{skins}

\definecolor{tblshade}{gray}{0.94}
\definecolor{attrfg}{RGB}{24,72,140}
\definecolor{boxframe}{gray}{0.55}
\definecolor{statgray}{gray}{0.40}
\newcommand{\attr}[1]{\textcolor{attrfg}{\textbf{#1}}}
\newtcolorbox{promptbox}[1][]{enhanced, colback=white, colframe=boxframe,
  boxrule=0.4pt, arc=2pt, left=6pt, right=6pt, top=4pt, bottom=4pt,
  fonttitle=\small\bfseries, coltitle=black, colbacktitle=gray!12,
  colbacklower=gray!6, segmentation style={dashed, draw=boxframe},
  fontupper=\small, fontlower=\small, #1}

\graphicspath{{figures/}}

\newcommand{\dataname}{\textsc{IndiPersona}}
\newcommand{\toolname}{\textsc{Addis}}
\newcommand{\metricname}{\textsc{Ard}}

\setcopyright{none}
\renewcommand\footnotetextcopyrightpermission[1]{}
\acmConference[FAccT '27]{ACM Conference on Fairness, Accountability, and Transparency}{}{}

\begin{document}

\title{LLM GerryMandering: Separating Interactions from Additive Effects in LLM Audits}

\author{George Rahul}
\authornote{Both authors contributed equally to this research.}
\affiliation{%
  \institution{International Institute of Information Technology Hyderabad}
  \city{Hyderabad}
  \country{India}}
\email{george.rahul@research.iiit.ac.in}

\author{Ronit Jalihal}
\authornotemark[1]
\affiliation{%
  \institution{International Institute of Information Technology Hyderabad}
  \city{Hyderabad}
  \country{India}}
\email{ronit.jalihal@students.iiit.ac.in}

\author{Bhaskara Hanuma Vedula}
\affiliation{%
  \institution{International Institute of Information Technology Hyderabad}
  \city{Hyderabad}
  \country{India}}
\email{vedula.hanuma@research.iiit.ac.in}

\author{Ponnurangam Kumaraguru}
\affiliation{%
  \institution{International Institute of Information Technology Hyderabad}
  \city{Hyderabad}
  \country{India}}
\email{pk.guru@iiit.ac.in}

\author{Pushkar Shukla}
\affiliation{%
  \institution{Cloudphysician.ai}
  \country{USA}}

\input{sections/00_abstract}



\maketitle

\input{sections/01-intro_combined}
\input{sections/02_related_work}
\input{sections/03_methodology}
\input{sections/04_results_discussion}

\input{sections/05_limitations_future_work}

\input{sections/06_ethics_statement}

\bibliographystyle{ACM-Reference-Format}
\bibliography{references}

\appendix
\input{sections/07_appendix}

\end{document}

%% file: sections/00_abstract.tex
\begin{abstract}

Intersectional bias arises when a model treats people who hold several marginalised identities at once differently from what their separate identities would predict. Auditing such intersectional subgroups remains a challenge: their number grows quadratically, and there is no agreed baseline for what independent attributes would predict. Without such a baseline, many existing audits flag a subgroup when it exceeds each attribute alone, a condition that attributes acting independently already satisfy. We test this standard criterion on 150 synthetic datasets built with no interaction. With the shuffle null, it flags 66.1 false subgroups per dataset on average, and at least one in every dataset. When we plant interactions of known size into these datasets, a repaired null lets the criterion recover only 32\% of them. \\
We introduce \toolname{} (Additive Intersectional Screen), which fits an additive model over all attributes, scores each subgroup by its Additive Residual Distance (\metricname{}) from that model’s prediction, and tests it against a Freedman–Lane null. On our synthetic benchmark, \toolname{} flags 0.09 false subgroups per dataset and recovers 84$\%$
of the same planted interactions. We then build \dataname{}, 10,000 synthetic Indian personas with six identity attributes, and audit 13 open language models. The standard criterion flags 7.1$\%$ of the 738 tested subgroups on average; \toolname{} flags 3.0$\%$. Gender identity crossed with sexual orientation is the largest pairing in every model, and counterfactual personas confirm it and additionally reveal a caste × class interaction in all but one model that the observational test lacked the data to detect. Much of what audits report as intersectional bias can thus arise from independent single-attribute effects alone, yet a small core of interactions survives.

 \end{abstract}

%% file: sections/01-intro_combined.tex
\section{Introduction}

Language models are increasingly deployed in systems that judge people, from ranking job applicants to answering a patient's medical questions~\cite{10.1145/3689904.3694699, testoni2026calibrated}. Audits report that responses depend on identity, and that some combinations are treated differently from either identity alone~\cite{an-etal-2024-large, gupta2026compoundingdisadvantageauditingintersectional, phutane2026ableist, testoni2026calibrated}. Most audits nevertheless examine one identity at a time. Intersectionality theory makes a specific claim about people who hold several identities at once: the harm they face is not reducible to the harms attached to each identity on its own, or even to their sum. Crenshaw writes that the intersectional experience is ``greater than the sum of racism and sexism''~\cite[p.~140]{crenshaw1989demarginalizing}, so the quantity to be exceeded is the sum of the parts. We operationalize this claim by asking whether a model's response to a combination of identities departs from what its separate identities predict.

Consider a model that attributes anxiety and depression to personas, and rates a woman three points higher on anxiety and a Muslim four points higher on depression. A Muslim woman then lies five points from the reference, farther than under either identity alone (Figure~\ref{fig:geometry}a). Yet, it is exactly where the two effects add, so no interaction is involved. Only a departure from this sum, such as the 1.5 points in Figure~\ref{fig:geometry}b, indicates an interaction, so exceeding each part is not equivalent to exceeding the \emph{sum} of the parts.

Detecting such interactions at scale is challenging, because candidate subgroups multiply quadratically with the number of identity attributes while the observations available for each decrease~\cite{pmlr-v80-kearns18a, himmelreich2025intersectionality}. Testing hundreds of subgroups simultaneously also inflates false discoveries unless the error rate is controlled across all of them. Automated subgroup discovery makes this search systematic~\cite{zhang2016identifying, pmlr-v80-kearns18a}, and the baseline against which it ranks subgroups determines which ones can be found. A common baseline in intersectional audits of language models, which we term the \emph{single-attribute criterion}, reports a combination when it fares worse than each of its attributes alone~\cite{gupta2026compoundingdisadvantageauditingintersectional, phutane2026ableist, testoni2026calibrated}. As the example shows, merely additive effects satisfy this criterion.  Conversely, it can miss real interactions when the two effects point in opposite directions, or when the interaction is perpendicular to their sum.

We address this problem with \toolname{} (Additive Intersectional Screen), which screens every pair of identity attributes for interactions in how language models treat people. We replace the single-attribute baseline with an additive prediction. We fit a linear regression over the whole dataset, giving every level of every attribute its own coefficient. It has no interaction terms. A subgroup's prediction is the sum of the coefficients of the attributes it carries. We define the \emph{additive residual distance} (\metricname{}) $\delta$ as the distance between a subgroup's observed mean response profile and this prediction. It is equal for interactions of equal magnitude in any direction. Figure~\ref{fig:geometry} contrasts the two baselines on our example. We test each subgroup against a null that keeps every attribute's effect and removes only the interaction effect, so the test asks whether two identities interact, not whether identity matters at all. We control the family-wise error rate at 5\% across the search (\S\ref{sec:auditing_framework}).

\input{sections/motiv}

In a natural sample of data, the attributes are correlated, so a pair's effect cannot be separated from its parts', and some subgroups are too small to study. So, we generate the personas we test, and sample their attributes independently of one another. The exception to this is caste, which is paired only with religions it actually occurs in. The dataset, \dataname{}, is ten thousand Indian personas crossing six attributes: gender identity, age, sexual orientation, socioeconomic status, caste and religion, each described in one of 50 everyday scenes. Fourteen models answer the same 82 questionnaire items about every persona. We test all 738 pairs of identity values with at least 50 personas, and use within-persona counterfactuals for combinations the main sample cannot estimate.

We test both baselines on data we build ourselves. It contains no interaction, so every subgroup a method flags in it is a false positive. We build it from the real design and the models' own answers, keeping each attribute's own effect and destroying only intersectional effects. Across 150 such datasets, each with 738 subgroups, \toolname{} flags 0.09 false subgroups per dataset on continuous answers. The single-attribute baseline flags 66.1, and fires on every dataset. We also synthetically add real intersectional effects into the dataset, of sizes we choose. \toolname{} recovers 84\% of them, against 32\% for the single-attribute baseline given the same null. On answers rounded to the Likert scale, rounding itself creates interactions and raises every method's error rate. We ran every other published method through the same test: per-item interaction regressions~\cite{3540261.3540461}, MAIHDA~\cite{evans2018multilevel} and PERMANOVA~\cite{doi:https://doi.org/10.1002/9781118445112.stat07841}. Each raises a false flag in at least 73\% of datasets, compared to our 20\%. With interactions planted, the precision of \toolname{} is 0.72, compared to 0.02--0.28 for the alternatives.

After running \toolname{} on the 13 open models, an average of only 3.0\% of subgroups are found to be intersectional on scale scores. By contrast, given the same statistical treatment, the single-attribute baseline flags 7.1\% (52 subgroups per model), of which only about nine coincide with ours. Gender identity crossed with sexual orientation is the single largest pairing, in every model we audit. Within-persona counterfactuals reproduce that effect in all 13 open models, and find a caste $\times$ class interaction in 12 of them, a result the observational test could not find due to lack of data. The location of these interactions holds up when we reword the prompt, draw a fresh sample of personas, or adjust for the template, though their direction depends on the wording. \metricname{} measures interaction in what models attribute to people, not harm, and our data come from one country, so we read the flagged combinations as candidates for closer scrutiny, not as estimates of how much discrimination exists in the world. Our contributions are as follows.

\begin{enumerate}
  \item \textbf{A metric for intersectional subgroup discovery.} We compare each subgroup against the sum of its attributes' separate effects, and test the difference against a null built for it.
  \item \textbf{A benchmark for intersectional metrics.} We build synthetic data with controllable interactions between the subgroups, so that the ground truth is known. On questionnaire-style answers, rounding weakens the error control of every method, but every published alternative we tried makes false positives far more often than ours.
  \item \textbf{A factorial persona dataset and an audit of fourteen models.} Ten thousand Indian personas whose identity attributes vary nearly independently, an audit of thirteen open models and one closed model, and the finding that only a small minority of subgroups survive a calibrated test, concentrated on a single pairing of attributes.
\end{enumerate}

%% file: sections/motiv.tex
\definecolor{womanfg}{HTML}{D9732B}
\definecolor{muslimfg}{HTML}{1F8A80}
\definecolor{gapred}{HTML}{C8102E}
\begin{figure}[t]
  \centering
  \begin{tikzpicture}[scale=0.36, >=stealth, font=\scriptsize]
    \foreach \sx/\ox/\oy/\lab/\ver/\dl in {
        0/3/4/{(a) additive}/{excess $=1.0$, flagged}/{$\delta=0$},
        10.5/3.9/5.2/{(b) along the sum}/{excess $=2.5$, flagged}/{$\delta=1.5$},
        21/1.8/4.9/{(c) at a right angle}/{excess $=1.2$, flagged}/{$\delta=1.5$}} {
      \begin{scope}[shift={(\sx,0)}]
        \fill[gray!10] (0,0) -- (4,0) arc (0:90:4) -- cycle;
        \draw[gray!55, dashed] (4,0) arc (0:90:4);
        \draw[->, gray] (-0.3,0) -- (6.2,0) node[below left] {anxiety};
        \draw[->, gray] (0,-0.3) -- (0,6.4) node[above right=-2pt] {depression};
        \draw[gray, densely dotted] (3,0) -- (3,4) -- (0,4);
        \draw[->, very thick, womanfg] (0,0) -- (3,0) node[below, pos=0.5] {woman};
        \draw[->, very thick, muslimfg] (0,0) -- (0,4) node[left, pos=0.5, rotate=90, anchor=south] {Muslim};
        \draw[->, thick] (0,0) -- (\ox,\oy);
        \draw[gapred, very thick] (3,4) -- (\ox,\oy);
        \draw[gray!70!black, thick, fill=white] (3,4) circle (0.2);
        \fill (\ox,\oy) circle (0.17);
        \node[align=center] at (3.0,-2.7) {\textbf{\lab}\\[1pt]\ver\\[1pt]\textcolor{gapred}{\dl}};
      \end{scope}
    }
    \node[anchor=west] at (3.35,4.45) {observed $=$ \textcolor{gray!70!black}{sum}};
    \node[anchor=west] at (11.0+3.5,5.55) {observed};
    \node[gray!70!black, anchor=north west] at (10.5+3.1,3.85) {sum};
    \node[gapred, anchor=west] at (10.5+3.55,4.45) {$\delta$};
    \node[gapred, anchor=south] at (21+2.45,4.6) {$\delta$};
    \node[gray!60!black, anchor=north east, align=right] at (-0.2,-0.15) {Hindu\\man};
    \node[gray!60!black, align=center, font=\tiny] at (2.05,0.95) {not\\flagged};
    \begin{scope}[shift={(35,0)}]
      \fill[gray!10] (3,0) arc (0:180:3) -- cycle;
      \draw[gray!55, dashed] (3,0) arc (0:180:3);
      \draw[->, gray] (-3.6,0) -- (6.2,0) node[below left] {anxiety};
      \draw[->, gray] (0,-0.3) -- (0,4.4) node[above right=-2pt] {depression};
      \draw[gray, densely dotted] (3,0) -- (0.402,1.5) -- (-2.598,1.5);
      \draw[->, very thick, womanfg] (0,0) -- (3,0) node[below, pos=0.45] {woman};
      \draw[->, very thick, muslimfg] (0,0) -- (-2.598,1.5);
      \node[muslimfg, anchor=south east] at (-1.6,1.6) {Muslim};
      \draw[->, thick] (0,0) -- (1.902,1.5);
      \draw[gapred, very thick] (0.402,1.5) -- (1.902,1.5);
      \draw[gray!70!black, thick, fill=white] (0.402,1.5) circle (0.2);
      \fill (1.902,1.5) circle (0.17);
      \node[gapred, anchor=south] at (1.15,1.55) {$\delta$};
      \node[align=center] at (0.3,-2.7) {\textbf{(d) opposed shifts}\\[1pt]excess $=-0.6$, missed\\[1pt]\textcolor{gapred}{$\delta=1.5$}};
    \end{scope}
  \end{tikzpicture}

\caption{The single-attribute criterion cannot tell addition from interaction. Relative to a Hindu man, woman adds 3 points of attributed anxiety and Muslim 4 of depression. Points beyond the dashed arc are flagged, and red marks \metricname{} $\delta$ from their sum (open circle). (a) Without interaction, the Muslim woman is still flagged. (b, c) The same interaction raises the excess more along the sum than at a right angle. (d) If Muslim lowers anxiety instead, the shifts nearly cancel and a real interaction stays inside the arc.}

  \label{fig:geometry}
\end{figure}
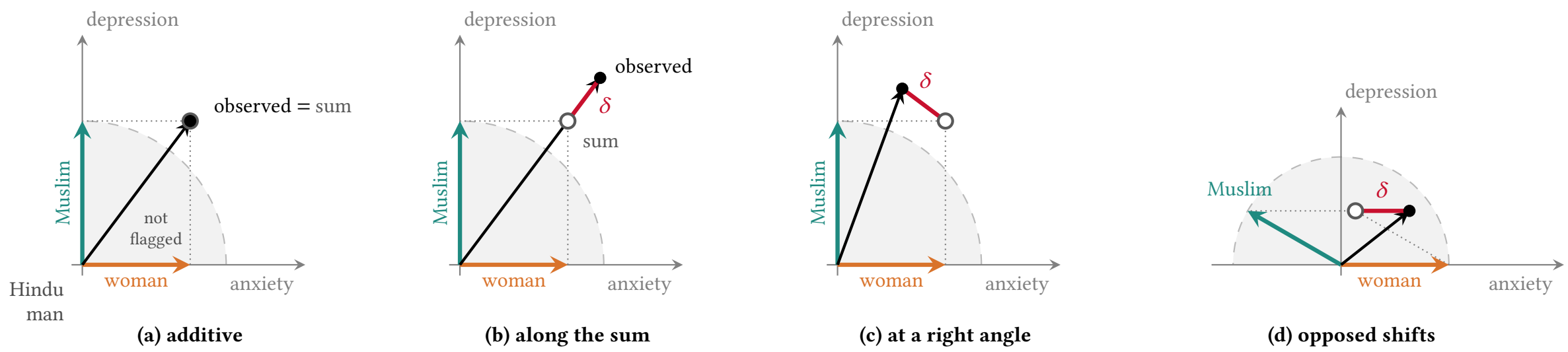

%% file: sections/02_related_work.tex
\section{Related Work}
\label{sec:related_work}

\noindent\textbf{Intersectional bias in language models.} Bias in language models is often strongest where identities meet, in embeddings, generated text, question answering and hiring decisions~\cite{10.1145/3461702.3462536, 3540261.3540461, lalor-etal-2022-benchmarking, ma-etal-2023-intersectional, parrish-etal-2022-bbq, an-etal-2024-large, 3716662.3716799}.
Fairness for each group does not imply fairness for their combinations~\cite{pmlr-v81-buolamwini18a, pmlr-v80-kearns18a}.
Most of this work uses Western categories such as race and gender~\cite{parrish-etal-2022-bbq, 10.1145/3461702.3462536}, while social hierarchies and the biases attached to them differ across countries, languages and regions~\cite{10.1145/3442188.3445896, ramesh-etal-2023-fairness, jha-etal-2023-seegull, bhutani-etal-2024-seegull, mukherjee-etal-2023-global}.

\noindent\textbf{Bias evaluation in India.} Work on India began with word embeddings, which encode caste, religious and gender associations in Hindi and other Indian languages~\cite{malik-etal-2022-socially, kirtane-anand-2022-mitigating, 10.1145/3465336.3475118}. Studies of Hindi then found that detecting gender bias needs community input~\cite{10.1145/3630106.3659017} and that models are more biased when generating Hindi than English~\cite{10.1145/3715275.3732208}.
Stereotype benchmarks for language models followed, covering Indian axes of disparity~\cite{bhatt-etal-2022-contextualizing}, caste and religion~\cite{10.1145/3677525.3678666, ijcai2025p1100}, several social dimensions~\cite{sahoo-etal-2024-indibias}, community-sourced stereotypes~\cite{NEURIPS2023_0dc91de8}, Indian languages~\cite{tomar-etal-2025-bharatbbq} and many identity groups~\cite{nawale-etal-2025-fairi}.
Benchmarks of this kind test whether a model repeats known associations, so they depend on curated inventories whose validity is often unclear~\cite{blodgett-etal-2021-stereotyping}, and they miss harms outside them.
Later work therefore studied open-ended generation~\cite{dammu-etal-2024-uncultured, seth2025deep, 10.1145/3593013.3594016, 3716662.3716702} and decisions.
Decision audits hold a task fixed and change only the identity, in hiring~\cite{rao2025invisiblefiltersculturalbias}, economic guidance~\cite{nair2026rupeebiasauditingdemographicbias}, matchmaking~\cite{naik2026simaaiuntycasteaudit} and education~\cite{gupta2026compoundingdisadvantageauditingintersectional}.
Because only the identity changes, they measure how a model treats a person rather than which words it associates, and our audit follows the same design.

\noindent\textbf{How intersections are tested.} Few of these audits test intersections systematically.
Stereotype benchmarks include a few predefined pairs, mostly with gender~\cite{sahoo-etal-2024-indibias, tomar-etal-2025-bharatbbq}, and decision audits change one attribute at a time~\cite{nair2026rupeebiasauditingdemographicbias} or a few within one task~\cite{naik2026simaaiuntycasteaudit, gupta2026compoundingdisadvantageauditingintersectional}, as do audits outside India with race, gender and age~\cite{tamkin2023evaluatingmitigatingdiscriminationlanguage, salinas2025whatsnameauditinglarge, 10.1145/3689904.3694699, an-etal-2024-large}.
When an audit compares a combination with its parts, it usually asks whether the combination fares worse than each part alone~\cite{tan2019assessing, lalor-etal-2022-benchmarking, gupta2026compoundingdisadvantageauditingintersectional, phutane2026ableist, testoni2026calibrated}, the single-attribute baseline that \S\ref{sec:naive} shows addition alone can satisfy.
Other audits fit an interaction term for one outcome at a time~\cite{3540261.3540461, 10.1145/3689904.3694699}.
Recent work widens the set of combinations, by pairing multi-attribute instances with their single-attribute sources~\cite{zhang2026comfair} or by flagging failures that appear only when two identity words are swapped together~\cite{souani2025hinter}, but it fits no additive baseline across all attributes and does not control false positives across cells.
Two studies come closest to our critique. Kanubala and Valera~\cite{kanubala2025misalignment} show that intersectional fairness metrics for classifiers respond more to additive bias than to interaction, and Rennard and Xypolopoulos~\cite{rennard2026simulate} compare simulated personas with an additive prediction and find that models represent intersectional identities in only one or two dimensions.
Searching over many subgroups is also established for classifiers, through subgroup fairness audits~\cite{pmlr-v80-kearns18a} and scan statistics~\cite{zhang2016identifying}.
Table~\ref{tab:related} compares these methods with ours. Unlike this work, we test every pairwise cell of a multivariate answer profile against an additive baseline fitted over all six attributes, control the family-wise error rate across cells, and confirm the main cells with counterfactuals.

\newcommand{\cmark}{\ding{51}}
\newcommand{\xmark}{\ding{55}}

\begin{table*}
\caption{\textbf{\toolname{} combines properties that prior methods address separately.} \emph{India}: Indian identity axes. \emph{Treatment}: only identity varies. \emph{All cells}: every pairwise cell. \emph{Additive}: additive baseline over all attributes. \emph{Profile}: answers tested jointly. \emph{Cell}: names the interacting cell. \emph{FWER}: error controlled across cells. \emph{Planted}: validated on planted interactions. \emph{Counterf.}: confirmed by counterfactuals. \cmark{} yes, \xmark{} no, $\sim$ partly, -- n/a.}
  \label{tab:related}
  \footnotesize
  \begin{tabular}{@{}l cc cccc ccc@{}}
    \toprule
    & \multicolumn{2}{c}{Setting} & \multicolumn{4}{c}{Intersections} & \multicolumn{3}{c}{Inference}\\
    \cmidrule(lr){2-3}\cmidrule(lr){4-7}\cmidrule(l){8-10}
    Method & India & Treatment & All cells & Additive & Profile & Cell & FWER & Planted & Counterf.\\
    \midrule
    Stereotype benchmarks, India~\cite{sahoo-etal-2024-indibias, tomar-etal-2025-bharatbbq} & \cmark & \xmark & \xmark & \xmark & \xmark & -- & \xmark & \xmark & \xmark\\
    Decision audits, India~\cite{rao2025invisiblefiltersculturalbias, nair2026rupeebiasauditingdemographicbias, naik2026simaaiuntycasteaudit} & \cmark & \cmark & \xmark & \xmark & \xmark & -- & \xmark & \xmark & $\sim$\\
    Single-attribute baseline~\cite{gupta2026compoundingdisadvantageauditingintersectional, lalor-etal-2022-benchmarking, phutane2026ableist, testoni2026calibrated} & $\sim$ & \cmark & $\sim$ & \xmark & \xmark & \cmark & \xmark & \xmark & \xmark\\
    Per-item interaction regression~\cite{3540261.3540461, 10.1145/3689904.3694699} & \xmark & \cmark & $\sim$ & $\sim$ & \xmark & \cmark & $\sim$ & \xmark & \xmark\\
    MAIHDA~\cite{evans2018multilevel} & \xmark & -- & \cmark & \cmark & \xmark & \cmark & \xmark & \xmark & \xmark\\
    PERMANOVA~\cite{doi:https://doi.org/10.1002/9781118445112.stat07841} & \xmark & -- & $\sim$ & $\sim$ & \cmark & \xmark & \xmark & \xmark & \xmark\\
    Multi-attribute pairing (CoMFair)~\cite{zhang2026comfair} & \xmark & \cmark & $\sim$ & \xmark & \xmark & \cmark & \xmark & \xmark & \xmark\\
    Metamorphic testing (HInter)~\cite{souani2025hinter} & \xmark & \cmark & $\sim$ & \xmark & \xmark & \cmark & \xmark & \xmark & \cmark\\
    Subgroup search for classifiers~\cite{pmlr-v80-kearns18a, zhang2016identifying} & \xmark & -- & \cmark & \xmark & \xmark & \cmark & $\sim$ & \xmark & \xmark\\
    \midrule
    \textbf{\toolname{} (ours)} & \cmark & \cmark & \cmark & \cmark & \cmark & \cmark & \cmark & \cmark & \cmark\\
    \bottomrule
  \end{tabular}
\end{table*}

%% file: sections/03_methodology.tex
\section{Methodology}
\label{sec:methodology}

\begin{figure}[t]
  \centering
  \includegraphics[width=\linewidth]{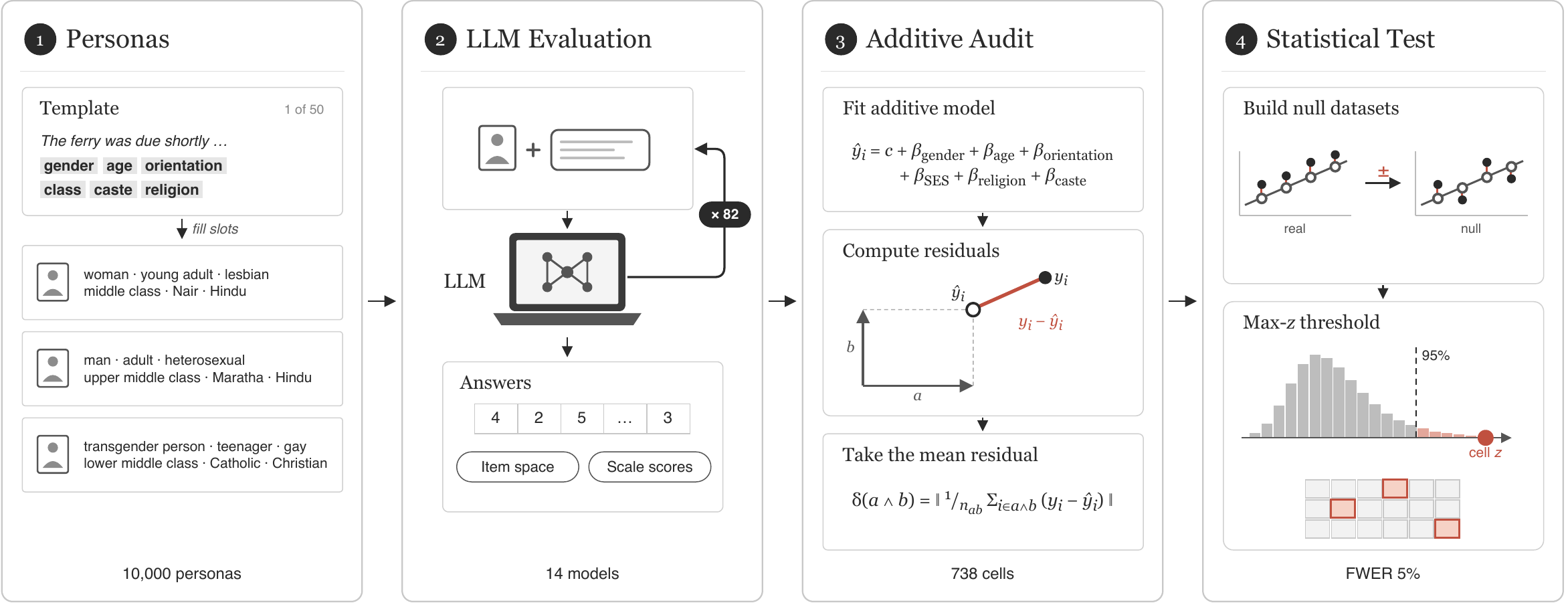}
  \caption{\textbf{The audit pipeline.} \dataname{} personas (\S\ref{sec:dataset_construction}) are answered item by item by each model. An additive model fits the answers (\S\ref{sec:auditing_framework}), and a Freedman--Lane null controls the family-wise error rate across all 738 cells (\S\ref{sec:error_control}).}
  \label{fig:pipeline}
\end{figure}

\toolname{} searches for intersectional subgroups in how a language model treats people.
It needs two parts.
A \emph{baseline} predicts where a subgroup should sit if its attributes do not interact.
A \emph{null} tells how far a subgroup can stray from the baseline by chance.
Section~\ref{sec:task} states the task.
Section~\ref{sec:naive} shows that addition alone satisfies the single-attribute baseline of current audits, and \S\ref{sec:auditing_framework} replaces it with an additive prediction.
Sections~\ref{sec:error_control} and \ref{sec:benchmark} add error control and a benchmark with known truth.
Section~\ref{sec:dataset_construction} describes \dataname{}, our dataset of personas, and \S\ref{sec:counterfactuals} describes its counterfactual quartets.
Figure~\ref{fig:pipeline} shows the full pipeline.

\subsection{The Audit Task}
\label{sec:task}

\noindent\textbf{Notation.}
Each of $N=10{,}000$ personas holds one value of each of six identity attributes.
A model rates persona $i$ on one questionnaire item at a time.
The answers form the persona's \emph{attributed profile} $y_i\in\mathbb{R}^d$, either its item answers ($d\le 82$) or $d=9$ scale scores computed from them (\S\ref{sec:error_control}).
For a value $a$, let $\bar y_a$ be the mean profile of the personas that hold $a$, $\bar y$ the population's mean profile, and $u_a=\bar y_a-\bar y$ the \emph{shift} of $a$.
Two values $a$ and $b$ of different attributes define a \emph{cell} $a\wedge b$, the set of $n_{ab}$ personas that hold both, with mean profile $\bar y_{a\wedge b}$.
We test all 738 cells with $n_{ab}\ge 50$, denoted $\mathcal{C}$.

\noindent\textbf{Target.}
If $a$ and $b$ do not interact, the cell sits at the additive prediction $\bar y+u_a+u_b$, and its \emph{interaction} is its departure from that prediction, $\Delta_{ab}=\bar y_{a\wedge b}-(\bar y+u_a+u_b)$.
Section~\ref{sec:auditing_framework} refines the additive prediction to account for the other four attributes.
We call a cell intersectional when $\Delta_{ab}\neq 0$, thus treating intersectionality as an interaction on the additive scale~\cite{vanderweele2014tutorial}, a working definition rather than the theory's own~\cite{crenshaw1989demarginalizing}.

\noindent\textbf{The audit.}
An audit returns a set of flagged cells $F\subseteq\mathcal{C}$.
It should flag the cells with $\Delta_{ab}\neq 0$ while keeping the family-wise error rate (FWER), the probability of flagging any cell with $\Delta_{ab}=0$, at 5\% or lower.
The flagged set depends on the baseline that the audit compares each cell with.

\subsection{The Single-Attribute Baseline Is Satisfied by Addition}
\label{sec:naive}

Current audits measure how far a subgroup sits from a reference, such as the population's mean or the most privileged group.
They call a cell $a\wedge b$ intersectional when it sits farther from this reference than each of its two values~\cite{tan2019assessing, gupta2026compoundingdisadvantageauditingintersectional}.
With the population's mean $\bar y$ as the reference, the rule reads:
\begin{equation}
  \mathrm{excess}(a\wedge b)=\lVert\bar y_{a\wedge b}-\bar y\rVert-\max\bigl(\lVert u_a\rVert,\,\lVert u_b\rVert\bigr) > 0 .
  \label{eq:excess}
\end{equation}
We call this rule the \emph{single-attribute baseline}, and the left side of Equation~\ref{eq:excess} its \emph{excess}.
The rule asks whether a combination fares worse than its parts. A positive excess can also arise by chance.
Audits that test their intersectional claims ask whether a subgroup differs from its comparison group at all.
They use permutation tests~\cite{tan2019assessing}, $t$-tests~\cite{gupta2026compoundingdisadvantageauditingintersectional}, rank tests~\cite{phutane2026ableist} or tests against an unperturbed prompt~\cite{testoni2026calibrated}.
All of these tests share one null hypothesis: identity has no effect.
We test the excess against the permutation version of this null, which we call the \emph{shuffle null}.
It shuffles the identity labels across personas.
Each persona keeps its answers but receives another persona's identities.
We then recompute the excess on each shuffled dataset.
Shuffling cuts the link between every identity and the answers, so shuffled data contain no main effects and no interaction.
The shuffle null therefore asks whether identity matters at all, not whether two values interact.
Real answers do contain main effects, so the excess of a real cell easily beats its shuffled values, even when the two values only add.
The test then reports interactions that are not there.
Section~\ref{sec:res_naive} shows how often. The excess has a second problem, even before any test.
It is positive for many cells in which nothing interacts.
Suppose a cell sits exactly at its additive prediction, so that $\Delta_{ab}=0$.

\begin{proposition}
\label{prop:additive}
If $\Delta_{ab}=0$, then $\mathrm{excess}(a\wedge b)>0$ if and only if $\langle u_a,u_b\rangle > -\tfrac{1}{2}\min\bigl(\lVert u_a\rVert^2,\lVert u_b\rVert^2\bigr)$.
\end{proposition}

We prove Proposition~\ref{prop:additive} in Appendix~\ref{sec:app_proof}. The sign of the excess depends on the angle between $u_a$ and $u_b$, not on any interaction.
Take two shifts of equal length.
The excess is then positive whenever the angle between them is less than $120^\circ$.
The single-attribute baseline flags such cells even when nothing interacts (Fig.~\ref{fig:geometry}(a)).
When a real interaction exists, the verdict also depends on which way the interaction points (Fig.~\ref{fig:geometry}(b, c)).
When the shifts point apart at more than $120^\circ$, the excess turns negative, and a small real interaction stays hidden (Fig.~\ref{fig:geometry}(d)).
The shuffle null fixes neither failure.
A cell flagged by the single-attribute baseline may hold only two ordinary biases added together.
A cell it misses may hold a real interaction.
We therefore replace the baseline itself.

\subsection{\metricname{}: Distance From the Additive Prediction}
\label{sec:auditing_framework}

\metricname{} compares each cell with the position that addition alone predicts.
To get this position, we fit a linear regression, the \emph{additive model}, to the attributed profiles of all personas:
\begin{equation}
  \hat{y}_i = c + \beta_{\text{gender}(i)} + \beta_{\text{age}(i)} + \beta_{\text{orientation}(i)} + \beta_{\text{SES}(i)} + \beta_{\text{religion}(i)} + \beta_{\text{caste}(i)} .
  \label{eq:additive}
\end{equation}
Here $\hat{y}_i$ is the predicted profile of persona $i$ and $c$ is a constant.
Each $\beta$ is the least-squares effect of one of the persona's six values, with its own coefficient for each value. For religion, for example, $\beta_{\text{religion}(i)}$ is $\beta_{\text{Christian}}$ for a Christian persona and $\beta_{\text{Hindu}}$ for a Hindu one.

The additive model has no interaction terms.
Any effect of a pair of values beyond the sum of their single effects therefore stays in the residual $y_i-\hat{y}_i$.

\metricname{}, the additive residual distance $\delta$, is the length of the cell's mean residual, the red segment in Figure~\ref{fig:geometry}:
\begin{equation}
  \delta(a\wedge b)=\Bigl\lVert \tfrac{1}{n_{ab}}\textstyle\sum_{i\in a\wedge b}\bigl(y_i-\hat{y}_i\bigr)\Bigr\rVert .
  \label{eq:delta}
\end{equation}
Take a simple case.
The personas differ only in the two attributes of the cell, and every value of one attribute appears equally often with every value of the other (for example, every gender with every orientation).
Then $\delta$ equals the length of the interaction $\Delta_{ab}$ from \S\ref{sec:task}, the standard interaction contrast~\cite{vanderweele2014tutorial}:
\begin{equation}
  \delta(a\wedge b)=\lVert\Delta_{ab}\rVert=\lVert\bar{y}_{a\wedge b}-\bar{y}_a-\bar{y}_b+\bar{y}\rVert .
  \label{eq:delta_balanced}
\end{equation}
If $\Delta_{ab}=0$, $\delta$ is zero up to random noise.
Otherwise it grows with $\lVert\Delta_{ab}\rVert$, whichever way the cell moves (Figure~\ref{fig:geometry}b,c).
Unlike the excess, $\delta$ does not depend on the angle between $u_a$ and $u_b$.

Our personas differ in all six attributes, so the additive model includes all six.
If a cell happened to hold more poor personas than usual, a model of its two attributes alone would mistake the shift of being poor for an interaction.
The full model removes that shift, and stays correct when cells differ in size (Appendix~\ref{sec:app_six}).

\subsection{Error Control Across Many Sparse Cells}
\label{sec:error_control}

We test 738 cells at once, and their sizes $n_{ab}$ differ more than twentyfold.
Two problems follow.
Small cells have larger variance, so a single threshold on raw $\delta$ would be set by the small cells.
And with hundreds of tests, some cell will exceed any fixed threshold by chance.
We therefore need a null that asks the right question, a score comparable across cells, and a rule that controls the error rate over all cells jointly.

\noindent\textbf{Hypotheses.}
For each cell $a\wedge b$, we ask whether the two values interact.
The null hypothesis $H_0\colon\Delta_{ab}=0$ states that the cell sits exactly at its additive prediction, while each value may still shift the answers on its own by any amount, and the alternative $\Delta_{ab}\neq 0$ states that the cell sits elsewhere.
This differs from the shuffle null of \S\ref{sec:naive}, whose null hypothesis is that identity has no effect at all.

\noindent\textbf{Procedure.} The test has four steps.
\begin{enumerate}
  \item \emph{Build null datasets.} Following Freedman and Lane~\cite{freedman1983nonstochastic} and Winkler et al.~\cite{winkler2014permutation}, we keep each persona's fitted part $\hat y_i$, flip the sign of its residual $y_i-\hat y_i$ at random, and add the two back together. The main effects are untouched, while any interaction, which lives in the residuals of its cell, is cancelled by the flips, and each persona's noise keeps its size (Appendix~\ref{sec:app_null}). We call this the \emph{Freedman--Lane null}.
  \item \emph{Recompute the metric.} On each of 2{,}000 null datasets, we refit the additive model and recompute $\delta$ for every cell.
  \item \emph{Standardize.} We convert each cell's observed $\delta$ into $z_{ab}=\bigl(\delta(a\wedge b)-m_{ab}\bigr)/s_{ab}$, where $m_{ab}$ and $s_{ab}$ are the mean and standard deviation of the observed $\delta$ pooled with its 2{,}000 null values, which removes the dependence on cell size.
  \item \emph{Apply the max-$z$ rule.} In each null dataset, we record the largest $z$ over all cells, and a cell is significant if its observed $z_{ab}$ exceeds 95\% of these maxima.
\end{enumerate}
\toolname{} is \metricname{} together with the Freedman--Lane null and the max-$z$ rule.

\noindent\textbf{Why this controls the error rate.}
The threshold is the 95th percentile of the largest $z$ over all cells, so under $H_0$ the chance that any cell exceeds it, the FWER, is designed to be at most 5\%.
This single-step rule approximates the min-$P$ procedure of Westfall and Young~\cite{westfall1993resampling} and rests on two assumptions, namely that residuals of different personas are independent and that each is as likely positive as negative (Appendix~\ref{sec:app_null}).
We therefore check the FWER empirically.
It stays near 5\% on continuous answers and on discrete Likert noise, but not when an additive signal is rounded to a few answer categories, because rounding creates real interactions on the answer scale, which every method detects (\S\ref{sec:res_benchmark}).

\noindent\textbf{Tested cells.}
Small intersectional groups make fairness hard to verify statistically~\cite{himmelreich2025intersectionality, ferrara2025sizeadaptive}, which is why we standardize each cell, require $n_{ab}\ge 50$, and control the error rate across all of $\mathcal{C}$.
The 47 caste $\times$ socioeconomic cells below this minimum are not tested (Appendix~\ref{sec:app_six}), and the counterfactual quartets of \S\ref{sec:counterfactuals} test one caste $\times$ class interaction directly.

\noindent\textbf{Representations.}
A \emph{representation} turns a model's answers about persona $i$ into the profile $y_i$ on which we compute $\delta$, and we use two.
In the \emph{item space}, each question's answer becomes one dimension, z-scored across personas, and profiles are compared with the distance of Cronbach and Gleser~\cite{cronbach1953assessing}.
Items to which a model gives the same answer for every persona carry no information about identity and are dropped, so $d$ ranges from 66 to 82 depending on the model.
In \emph{scale scores}, each of the $d=9$ dimensions is one trait scored with the standard key of its questionnaire, namely the five Big Five traits, need for cognition, empathic concern, anxiety and depression (Appendix~\ref{sec:app_alpha} reports each scale's internal consistency per model).
Scale scores carry less signal, so \toolname{} flags fewer cells on them, and a finding that holds in both representations does not depend on how the answers are summarized.

\subsection{A Benchmark With Known Truth}
\label{sec:benchmark}

Real answers have no ground truth.
When a method flags a cell, we cannot tell whether the flag is right.
To compare methods, we need data in which we know which cells interact.
Our benchmark builds such data from real answers with the plasmode approach~\cite{vaughan2009use, franklin2014plasmode}.
The plasmode approach starts from real data, keeps their structure, and plants a known truth.

\begin{figure}[t]
  \centering
  \includegraphics[width=\linewidth]{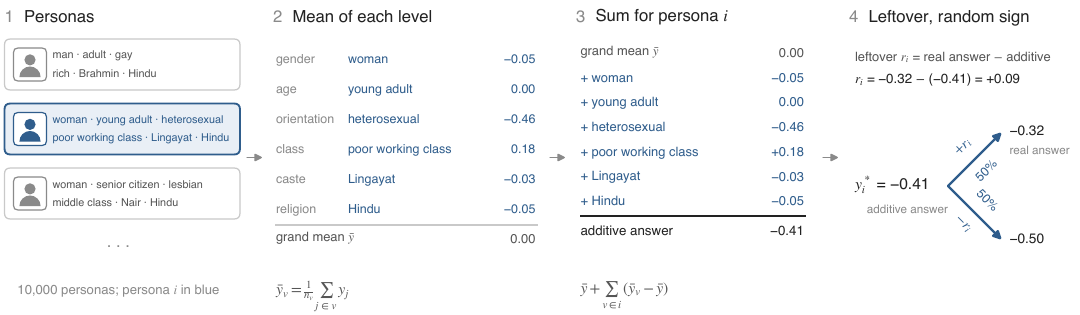}
  \caption{\textbf{The group-mean generator keeps every main effect and removes every interaction.} Real answers of Llama~3 8B on the anxiety scale, in standard deviations. Each persona's synthetic answer is the grand mean plus the shift of each of its six values, plus its own leftover with a random sign.}
  \label{fig:benchmark_gen}
\end{figure}

We build every dataset three times, once from the answers of each of three models from different families (Llama~3 8B, Gemma~4 E4B, Ministral 8B).
Each model thus contributes equally.
Our main \emph{noise generator}, \emph{group-mean}, rebuilds each persona's answer from group averages, the grand mean plus the shift of each of its six values, and adds back its own leftover with a random sign (Figure~\ref{fig:benchmark_gen}).
It never fits the additive model (Equation~\ref{eq:additive}), which makes it an impartial check of \toolname{}.
Two further generators start from that model's fit $\hat y_i$.
\emph{Flip} flips the signs of the residuals $y_i-\hat y_i$ at random, as in the Freedman--Lane null, and \emph{permute} shuffles them across personas.
Every generated dataset has real main effects and no interaction.
We then \emph{plant} interactions by shifting every persona in a few chosen cells by a known amount, measured in residual standard deviations.
In some datasets we plant nothing.

When nothing is planted, every flag is a false positive, so we measure each method's FWER.
Answers rounded to the Likert scale are the one exception (Appendix~\ref{sec:app_likert}).
When interactions are planted, we know which cells a method should flag, so we also measure its recall and precision.
We follow the ADEMP structure for simulation studies~\cite{morris2019using}.

A scenario fixes how the answers are built and where interactions are planted.
We report four scenarios, each testing a different risk: answers rounded to the Likert scale, an interaction in a single item, several interactions in one attribute pair, and the real gender identity $\times$ orientation pattern (Appendix~\ref{sec:app_scenarios}).

The benchmark tests whether a method can find planted interactions.
To find real interactions in language models, we need a dataset of personas.

\subsection{The \dataname{} Dataset}

\label{sec:dataset_construction}

\dataname{} is a set of synthetic Indian personas that combine six identity attributes: gender identity, age, sexual orientation, socioeconomic status, religion, and caste and community. Table~\ref{tab:demographics} in Appendix~\ref{sec:app_attributes} lists the values of each attribute.
Each cell needs a large $n_{ab}$ to estimate $\delta$.
We therefore assign the attributes independently of one another, except that religion follows caste.
This trades realism for estimability.
Real populations are correlated, so the personas show how models treat each combination of identity labels, not how bias falls on India's population.

\noindent\textbf{Personas.}
We generate 10{,}000 personas.
Gender, age, sexual orientation, socioeconomic status and caste are each assigned by quota, independently of one another, so every value appears equally often.
Religion is then chosen among the religions in which the persona's caste occurs~\cite{sachar2006, scorder1950}.
This rule avoids implausible combinations such as a Brahmin Muslim.
Our caste list is a simplified working representation of caste, not a complete one.
Assigning attributes independently also produces combinations that are contradictory, such as a man described as lesbian, or sensitive, such as a child with a sexual orientation. We keep them so that every cell can be estimated, but a flag in such a cell may reflect the model's reaction to the contradiction rather than an intersectional effect. None of the three main cells (\S\ref{sec:res_rare}) is such a combination.

\noindent\textbf{Templates.}
Each persona appears in a short scene, such as a queue at a public office or a ferry landing (Fig.~\ref{fig:template_example} in Appendix~\ref{sec:app_prompts}).
We use 50 templates, so that no single scene's wording drives the answers.
Each has a fixed opening and one sentence per attribute.
Gender comes first, and the other five attributes follow in a random order for each persona.
A few templates add details that could hint at some traits, so we do not use those templates with the questionnaires they could affect (Appendix~\ref{sec:app_prompts}).
Templates are assigned at random, independently of identity.

\noindent\textbf{Questions and answer order.}
The model rates each persona on 82 items from five questionnaires, one item per prompt (Appendix~\ref{sec:app_prompts}, Table~\ref{tab:items}).
Every item is rewritten in the third person, so that the model rates a described person instead of reporting on itself.
``I am the life of the party'' becomes ``This person is the life of the party''.
Answer options always appear in ascending order.

\noindent\textbf{Control arms.}
\dataname{} also holds three control arms for our robustness checks: a fresh sample of personas, two rewordings of every prompt, and personas without identity attributes (Appendix~\ref{sec:app_arms}).

\subsection{Counterfactual Quartets}
\label{sec:counterfactuals}

\toolname{} shows which cells of the main personas interact, but not why.
Each cell mixes the effect of its two labels with the other four values of its personas.
Counterfactual quartets remove this mixing.
We take 250 base personas per contrast and write each one four times, in a $2\times2$ design that changes two labels and nothing else.
For example, one base persona is written as a rich Brahmin, a rich Dalit, a poor working class Brahmin and a poor working class Dalit, with every other detail the same.
The interaction is the difference between two changes: the move from rich to poor working class for the Dalit version, minus the same move for the Brahmin version.
The rest of the persona stays fixed, so this within-persona difference-in-differences estimates the causal interaction of the two labels in the prompt~\cite{egami2019causal}.
Quartets also help where the main personas are too sparse.
With 31 castes, the cell Dalit $\times$ poor working class has $n_{ab}$ of only about 50 to 80.
\toolname{} cannot detect most interactions in a cell this small, but the quartet above tests the two labels directly.
We use four contrasts: the main gender identity $\times$ orientation interaction, a check that this interaction is specific to the transgender label, a caste interaction, and a calibration contrast where we expect little interaction (Appendix~\ref{sec:app_contrasts}).

\section{Experimental Setup}
\label{sec:experimental_setup}

We run \toolname{} on 14 models and check four things in \S\ref{sec:results_discussion}.
We check whether \toolname{} keeps its FWER near 5\% on the real design (Appendix~\ref{sec:app_calibration}), and whether the interactions it finds are caused by the identity labels in the prompt rather than by the other values of the personas in a cell (\S\ref{sec:res_causal}).
We also check whether its findings survive changes to the prompt, the sample and the representation (Appendix~\ref{sec:app_robust}), and how it compares with the published alternatives (\S\ref{sec:res_benchmark} and Appendix~\ref{sec:app_baselines_real}).

\noindent\textbf{Models.}\label{sec:setup_models}
We audit 13 open instruct models from the Gemma, Llama, Ministral, Phi and Qwen families, with up to 24B parameters, and DeepSeek V4.1 Flash (all 14 are listed in Figure~\ref{tab:headline}).
Each model answers at temperature 0, with thinking off, and returns only a number.
Because a model whose answers barely change with the persona cannot show bias, we analyze a model only if it passes a \emph{persona-sensitivity check}, which all 14 pass (Appendix~\ref{sec:app_setup}).

\noindent\textbf{Published alternatives.}\label{sec:setup_alternatives}
We compare our test with three kinds of published methods, each run as published with its own null, correction and decision rule.
The \emph{single-attribute baseline} (Equation~\ref{eq:excess}) is run in two forms, flagging every cell with a positive excess or only cells whose excess is significant under the shuffle null (\S\ref{sec:naive}); in \S\ref{sec:res_naive} we also pair the excess with our Freedman--Lane null to separate the effect of the null from that of the statistic.
\emph{Tests of departure from addition} test for an interaction, as our test does.
Per-item regression fits an interaction term for each item, as in LLM audits~\cite{3540261.3540461}, with a Bonferroni correction across items and cells, and we run it with classical standard errors, as published, and with HC3 standard errors~\cite{mackinnon1985some}, which allow the noise to differ between cells.
MAIHDA, a standard intersectional method in social epidemiology, adds a random intercept for every subgroup to the additive model~\cite{evans2018multilevel}.
PERMANOVA, a standard permutation test for multivariate data, tests the whole interaction term of each attribute pair~\cite{doi:https://doi.org/10.1002/9781118445112.stat07841}, so it cannot say which cell in the pair interacts.
\emph{Metamorphic testing} with HInter reports a hidden failure when swapping two identity words changes the answer but swapping either word alone does not~\cite{souani2025hinter}; because our counterfactual quartets have the same $2\times2$ structure, we apply HInter's rule to every item of the quartets (Appendix~\ref{sec:app_baselines_real}).

%% file: sections/04_results_discussion.tex
\section{Results and Discussion}
\label{sec:results_discussion}

Our results support three claims.
First, \toolname{} keeps its FWER near 5\% on continuous answers.
The single-attribute baseline has a high error rate, because it flags cells whose attributes only add (\S\S\ref{sec:res_naive}--\ref{sec:res_benchmark}).
Second, few cells pass \toolname{} on real answers.
In every model, the flagged cells concentrate on gender identity $\times$ sexual orientation (\S\ref{sec:res_rare}).
Third, changing the labels reproduces the gender identity $\times$ orientation interaction, but the direction of the interaction depends on the prompt (\S\S\ref{sec:res_rare}--\ref{sec:res_causal}).
Appendix~\ref{sec:app_baselines_real} compares the published alternatives on real answers.

\subsection{The Single-Attribute Baseline on Synthetic Data}
\label{sec:res_naive}

We first test the single-attribute baseline on synthetic data.
Section~\ref{sec:res_fpr_excess} measures its false positive rate and the role of its null.
Section~\ref{sec:res_direction_blind} measures its recall and precision, and how both depend on the direction of the interaction.

\subsubsection{Where the Single-Attribute Baseline's False Positives Come From}
\label{sec:res_fpr_excess}

We use the 500 datasets with no interaction from Appendix~\ref{sec:app_calibration}.
These datasets are built from the real answers of Llama~3 8B with the group-mean generator.
As Proposition~\ref{prop:additive} predicts, the excess (Equation~\ref{eq:excess}) is positive for 89.5\% of cells in the item space and for 75.5\% on scale scores.
The excess is therefore a biased measure of interaction.
With no interaction present, its expected value is positive, not zero. To correct this bias, the null distribution must carry the same positive bias.
The shuffle null does not, because it removes the main effects that make the excess positive.
On the benchmark, the excess with the shuffle null flags at least one cell in every dataset, and 66.1 cells per dataset on average (Table~\ref{tab:naive_null}).

\begin{table*}
  \footnotesize
  \setlength{\tabcolsep}{4pt}
  \begin{minipage}[t]{0.56\linewidth}
    \caption{\textbf{The Freedman--Lane null removes the single-attribute baseline's false positives, but not its misses.} Scale scores, group-mean generator. \emph{Flagged}, \emph{Cells}: datasets flagged and false cells per dataset, nothing planted. Recall and precision: the real gender identity $\times$ orientation pattern at one residual SD.}
    \label{tab:naive_null}
    \centering
    \begin{tabular}{@{}l l r r r r@{}}
      \toprule
      Statistic & Null & Flagged & Cells & Recall & Prec.\\
      \midrule
      Excess (Eq.~\ref{eq:excess}) & Shuffle & 100\% & 66.1 & 47.9\% & 0.06\\
      Excess (Eq.~\ref{eq:excess}) & Freedman--Lane & 2.7\% & 0.03 & 32.4\% & 0.99\\
      \metricname{} ($\delta$) & Freedman--Lane & 8.7\% & 0.09 & 84.4\% & 0.99\\
      \bottomrule
    \end{tabular}
  \end{minipage}\hfill
  \begin{minipage}[t]{0.4\linewidth}
    \caption{\textbf{The excess finds an interaction along the cell's shift and misses one at a right angle.} Recall on scale scores, group-mean generator, Freedman--Lane null. Planted size in residual SDs.}
    \label{tab:direction}
    \centering
    \begin{tabular}{@{}l l r r@{}}
      \toprule
      Size & Direction & Excess & \metricname{}\\
      \midrule
      0.4 & Along the shift & 43\% & 38\%\\
      0.4 & At a right angle & 0\% & 53\%\\
      0.8 & Along the shift & 93\% & 90\%\\
      0.8 & At a right angle & 35\% & 97\%\\
      \bottomrule
    \end{tabular}
  \end{minipage}
\end{table*}

With the Freedman--Lane null, the result reverses.
The Freedman--Lane null keeps every main effect, so the bias appears in the null too and cancels.
The excess now flags only 2.7\% of datasets instead of all of them.
Appendix~\ref{sec:app_likert_results} reports the same comparison on Likert-rounded answers.
There, the repaired excess keeps a higher precision than \toolname{}, because it finds few interactions of any kind. Repairing the null fixes the false positive rate (Appendix~\ref{sec:app_ablation} tests a second simple repair).
It does not change what the excess measures: distance from the population mean, not distance from the additive prediction (Fig.~\ref{fig:geometry}c,d).
Section~\ref{sec:res_direction_blind} measures how many interactions the repaired excess misses as a result.

\subsubsection{Recall by the Direction of the Interaction}
\label{sec:res_direction_blind}

Even with the Freedman--Lane null, the excess misses many interactions.
On the real gender identity $\times$ orientation pattern, the repaired excess recovers 32.4\% of the planted cells, against 84.4\% for $\delta$ (Table~\ref{tab:naive_null}).
Both keep a precision near 0.99, so only recall separates them. The direction of the planted interaction explains the gap.
We plant the same interaction twice in the same cell (Table~\ref{tab:direction}).
The first time, it points along the cell's additive shift from the population mean. The second time, it points at a right angle to that shift.
The excess detects the first interaction but misses most of the second.
$\delta$ detects both.

The right-angle case is the one intersectionality theory is about.
An interaction at a right angle means that the combination does something neither part does.
In a high-dimensional space such as the nine scale scores, most directions are close to perpendicular.
The right-angle case is therefore also the most common one.

\subsection{Published Alternatives on Synthetic Data}
\label{sec:res_benchmark}

We next compare \toolname{} with the published alternatives (\S\ref{sec:setup_alternatives}) on the benchmark (\S\ref{sec:benchmark}).
Section~\ref{sec:res_fpr_methods} compares their false positive rates.
Section~\ref{sec:res_recall} compares their recall and precision when we plant known interactions.

\subsubsection{Comparing False Positive Rates Across Methods}
\label{sec:res_fpr_methods}

We use benchmark datasets with no interaction, built from the answers of three models with the group-mean generator.
On continuous answers, \toolname{} flags 8.7\% of datasets on scale scores(95\% CI 4.7--24.4) and 4.7\% in the item space (Table~\ref{tab:bench_fwer}a).
Per-item regression with HC3 standard errors is the exception.
It holds its FWER about as well as \toolname{}.
With classical standard errors, as published, per-item regression makes a false positive in every item-space dataset.
Appendix~\ref{sec:app_calibration} reports a finer calibration of \toolname{}.\label{sec:res_calibration}

\begin{table*}
  \caption{\textbf{\toolname{} keeps the highest precision in every setting.} \emph{Flagged}: datasets with any flag when nothing is planted. \emph{Recall / prec.}: interactions planted at one residual SD in (a), two in (b). Per-item regression uses Bonferroni. (a) Group-mean generator; real pattern on scale scores, single item in the item space. (b) Permute generator (Appendix~\ref{sec:app_likert}). MAIHDA and PERMANOVA ran here only, on 30 datasets, as they are slow (Appendix~\ref{sec:app_baselines_real}).}
  \label{tab:bench_fwer}
  \footnotesize
  \setlength{\tabcolsep}{4pt}
  \begin{minipage}[t]{0.6\linewidth}
    \centering
    \textbf{(a) Continuous answers}\\[2pt]
    \begin{tabular}{@{}l rr rr@{}}
      \toprule
      & \multicolumn{2}{c}{Flagged} & \multicolumn{2}{c}{Recall / prec.}\\
      \cmidrule(lr){2-3}\cmidrule(l){4-5}
      Method & Scale & Item & Real pattern & Single item\\
      \midrule
      \toolname{} & 8.7\% & 4.7\% & 84.4\% / 0.99 & 53.1\% / 0.98\\
      Per-item HC3 & 9.3\% & 3.3\% & 75.6\% / 0.98 & 97.1\% / 0.53\\
      Per-item classical & 36.0\% & 100\% & 77.6\% / 0.93 & 99.2\% / 0.15\\
      \bottomrule
    \end{tabular}
  \end{minipage}\hfill
  \begin{minipage}[t]{0.38\linewidth}
    \centering
    \textbf{(b) Likert-rounded answers}\\[2pt]
    \begin{tabular}{@{}l r r@{}}
      \toprule
      Method & Flagged & Recall / prec.\\
      \midrule
      \toolname{} & 20.0\% & 78.1\% / 0.72\\
      Per-item HC3 & 99.3\% & 58.8\% / 0.20\\
      Per-item classical & 100\% & 85.9\% / 0.20\\
      MAIHDA & 96.7\% & 76.0\% / 0.28\\
      PERMANOVA & 73.3\% & 72.7\% / 0.02\\
      \bottomrule
    \end{tabular}
  \end{minipage}
\end{table*}

\noindent\textbf{Likert-rounded answers.}
Real answers are whole numbers on a four-point or five-point scale.
One benchmark scenario therefore rounds its continuous answers to these categories (\S\ref{sec:benchmark}).
Rounding a measurement scale is known to create departures from additivity~\cite{loftus1978interpretation, wagenmakers2012interpretation}.
A few cells therefore depart from the additive prediction even when nothing is planted.
In an average dataset, about 3 of the 738 cells depart enough for \toolname{} to detect.
Appendix~\ref{sec:app_likert} describes the construction and how we find these cells.

Rounding weakens the error control of every method we test.
\toolname{} weakens least.
It flags at least one cell in 20.0\% of the 150 rounded datasets.
Every published alternative we tried exceeds 96\%, except PERMANOVA at 73\% (Table~\ref{tab:bench_fwer}b).
On rounded data that copy the real design, the FWER of \toolname{} also rises above 5\%.
A control shows that rounding, not \toolname{}, causes these extra flags (Appendix~\ref{sec:app_likert_results}).

\subsubsection{Recall and Precision Across Methods}
\label{sec:res_recall}

\toolname{} has high precision and recall in every setting but one (Table~\ref{tab:bench_fwer}).
It loses recall when a single item carries the whole interaction.
\metricname{} measures one distance across all items, so a change in one item is diluted.
Per-item regression tests each item separately.
It finds more of these single-item cells, but at about half the precision of \toolname{}.

\subsection{Interactions Found in the Models' Answers}
\label{sec:res_rare}

We now run \toolname{} on the real answers of the 14 models about the \dataname{} personas (\S\ref{sec:dataset_construction}).

\noindent\textbf{How many cells interact.} 
\toolname{} flags few cells in every model.
Averaged over the 13 open models, it flags 22 of the 738 cells on scale scores (3.0\%) and 55 in the item space (7.5\%).
For DeepSeek V4.1 Flash, it flags 11 and 56 cells (Appendix~\ref{sec:app_all_cells}).
The single-attribute baseline reports far more cells.
Its excess is positive for 75--83\% of cells on scale scores.
This share is about the same as on synthetic data with no interaction (\S\ref{sec:res_fpr_excess}).
Under the shuffle null (\S\ref{sec:naive}), the baseline still flags 52 cells per model, more than twice as many as \toolname{}. The two tests also disagree on which cells interact.
Only about 9 cells per model are flagged by both.
Part of this gap lies out of the excess's reach.
In every model, some cells flagged by \toolname{} have a negative excess, as in Fig.~\ref{fig:geometry}(d).
No threshold on the excess can flag these cells.
In DeepSeek, one such cell is transgender person $\times$ senior citizen, its second-strongest interaction in the item space.

\noindent\textbf{Which cells interact.} 
\label{sec:res_location}
On scale scores, gender identity crossed with sexual orientation is the largest pairing in every model (Appendix~\ref{sec:app_all_cells}).
The 12 cells of this pairing are under 2\% of the tested cells but carry a third of all flags.
These flags are not independent findings, because one strong cell shifts the residuals of its row and column (Appendix~\ref{sec:app_echo}).
Three of these cells interact in nearly every model.
Transgender person $\times$ queer and transgender person $\times$ heterosexual interact in all 13 open models in both representations.
Man $\times$ queer interacts in 12, and DeepSeek shows all three.
We call these three cells the \emph{main cells}.
In the item space, each main cell sits two thirds to one residual standard deviation from its additive prediction.
To check that the main cells are not an accident of one sample, we rerun five of the models on a fresh sample of 5{,}000 personas drawn the same way (\S\ref{sec:dataset_construction}).
All three main cells are flagged again in all five models in the item space.
On scale scores, the two transgender cells are flagged again in all five models, and man $\times$ queer in four (Appendix~\ref{sec:app_all_cells}, Table~\ref{tab:effectsizes}).
Per-item regression and MAIHDA also flag all three main cells in at least 12 of the 13 models (Appendix~\ref{sec:app_baselines_real}).

The item space carries more information, so it shows a wider pattern.
In the item space, transgender person $\times$ poor working class, poor working class $\times$ heterosexual and ultra rich $\times$ heterosexual also interact in all 13 models.
Gender identity $\times$ orientation is the largest pairing in 8 of these models.
Appendix~\ref{sec:app_all_cells} lists every significant cell.

\begin{figure}[t]
  \centering
  \includegraphics[width=0.6\linewidth]{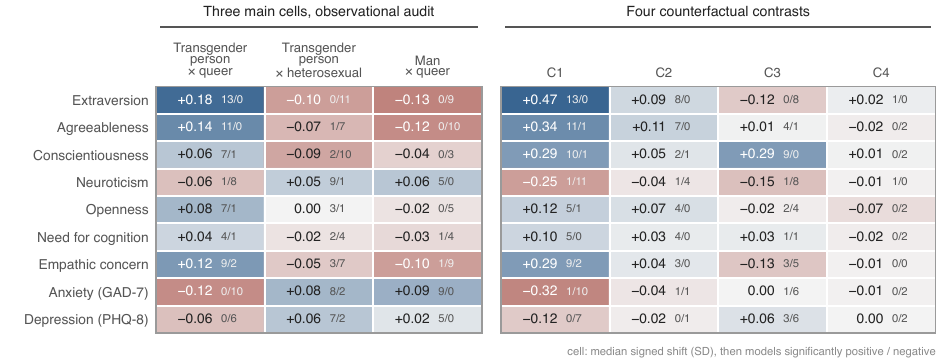}
  \caption{\textbf{C1 moves the same traits as transgender person $\times$ queer, and C3 moves conscientiousness, extraversion and neuroticism.} Median shift over 13 open models, in SD of each scale. Below: models significantly positive / negative. The two blocks are standardised separately, so compare signs, not sizes.}
  \label{tab:direction_main}\label{tab:cf_direction}
\end{figure}

\noindent\textbf{The direction of the interactions.} 
On scale scores, the departure measured by $\delta$ is a vector with one dimension per trait.
Each dimension is the cell's interaction on that trait.
For every model and cell, we can therefore see which traits carry the interaction (Figure~\ref{tab:direction_main}).
The excess compares two lengths and keeps no direction, so the single-attribute baseline cannot show which traits move.
Under our main wording, models attribute more extraversion, agreeableness and empathic concern to queer transgender personas than their two labels predict, and less anxiety.
The rise in extraversion is significant in all 13 open models.
The other two main cells lean the other way, toward less extraversion and more anxiety.

In questionnaire points, these shifts are small.
A typical model gives a queer transgender persona 0.28 fewer points of anxiety on the GAD-7 (range 0 to 21) than its two labels predict.
All 13 models move anxiety the same way.
$\delta$ is still large, because the departure spreads over many scales and items at once.

Not all of these results are robust.
When every prompt is reworded, the drop in anxiety for queer transgender personas weakens or disappears in most models.
Across all flagged cells, only about half reappear on the fresh sample (Appendix~\ref{sec:app_robust}).

\noindent\textbf{Caste.}
No caste cell interacts in more than 4 of the 13 open models.
Caste cells are too small for \toolname{} to measure well.
We therefore test caste with counterfactual quartets in \S\ref{sec:res_causal}.

\subsection{Results From the Counterfactual Quartets}
\label{sec:res_causal}
The counterfactual quartets (\S\ref{sec:counterfactuals}) change only two labels in each persona and hold everything else fixed.
Any interaction in a quartet is therefore caused by those two labels, within the model's prompts.

We use contrast C4 as the reference.
C4 changes adult to senior citizen and middle class to upper middle class.
These are two small steps between neighbouring levels, and we do not expect them to interact.
The C4 interaction is small but not zero.
We use the C4 interaction as a yardstick for the size of the other contrasts.

C1 changes man to transgender person and heterosexual to queer.
C1 thus tests the main interaction of \S\ref{sec:res_rare}, a cell \toolname{} had already flagged, so it confirms that finding rather than discovering a new one.
C1 reproduces this interaction in all 13 open models, with contrast values about five times as large as C4.
C1 also points the same way as the audit of the main personas: more extraversion in all 13 models and less anxiety in 10 (Figure~\ref{tab:cf_direction}).
C2 makes the same change with woman in place of transgender person.
C2 is larger than C4 with 95\% bootstrap confidence in only 5 of the 13 models, against all 13 for C1.

C3 changes Brahmin to Dalit and rich to poor working class.
The C3 interaction is about as large as the C1 interaction, and larger than C4 in 12 of the 13 models.
Models attribute more conscientiousness and less extraversion and neuroticism to a poor working-class Dalit persona than its two labels predict.
Anxiety and depression move in different directions in different models.
Caste cells showed no interaction in \S\ref{sec:res_rare} because they are small (\S\ref{sec:counterfactuals}), not because caste labels do not interact.
Auditors who care about caste should use targeted counterfactuals.

\section{Conclusion}
These results suggest three practices for intersectional audits of language models.

\emph{Use an additive baseline.}
Without an additive baseline, an audit mistakes cells whose attributes only add for interactions.
The audit also misses real interactions whose direction barely changes the cell's distance from the population mean.

\emph{Use a null that keeps the main effects.}
The shuffle null removes each attribute's own effect along with any interaction, so almost any cell can beat it.
A null that keeps the main effects, such as the Freedman--Lane null, avoids this problem.

\emph{Test small cells with counterfactuals.}
Attributes with many values, such as caste, give few personas per cell.
Counterfactual quartets can test the labels of such cells directly.

None of these practices measure harm.
Whether a flagged cell harms people depends on the outcome and the setting.
An audit has to judge harm separately.

%% file: sections/05_limitations_future_work.tex
\section{Limitations and Future Work}
\label{sec:limitations_future_work}

Our personas are synthetic, so they show how models treat each combination of identity labels, not how bias falls on India's population.
Whether two values add depends on the scale on which we measure the answers.
The share of intersectional cells therefore differs between the item space and scale scores, although the main interactions sit in the same place in both.
Flagged cells are non-additive on the answer scale, which can include effects of the answer format near its floor or ceiling, and a flagged pair may also involve a third attribute (Appendix~\ref{sec:app_robust}).
Our test is also not calibrated on Likert-rounded benchmark data, where its FWER is 20.0\%, so single flagged cells on Likert answers should be confirmed with an ordinal model.
Future work should extend the test from pairs of attributes to combinations of three or more, and apply it to real human answers to the same questionnaires, to compare how people and language models treat the same combinations of identities.

%% file: sections/06_ethics_statement.tex
\section*{Ethical Considerations and Adverse Impacts}
This work involves no human subjects and no personal data, since every persona is generated from templates and every answer comes from a language model.
The anxiety and depression items ask a model what a described person would answer, so they are attributions, not screenings, and must not be read as clinical claims about any group.
Our caste list is a simplified working representation and may not match how communities describe themselves.
The main risk is misreading our results: a flagged cell describes how models treat a combination of labels, not real people, and a cell our test does not flag is not shown to be free of harm, only to hold no interaction large enough to detect.
We will release our code and the model answers on GitHub upon publication, as audit evidence, not as descriptions of real groups or as training data.

\section*{Generative AI Usage Statement}
We used Claude Opus 5.5 in Claude Code to search the literature for Section~\ref{sec:related_work}, to prototype code, and to proofread and reword our own sentences.
We checked every cited paper, tested all code behind our results, and reviewed every edit.
The ideas, analyses and claims are ours, and we take full responsibility for them.

%% file: sections/07_appendix.tex
\section{Proof of Proposition~\ref{prop:additive}}
\label{sec:app_proof}

Let $u$ and $v$ be the shifts of values $a$ and $b$ from the population's mean profile.
Under exact additivity, the cell $a\wedge b$ shifts by $u+v$, so its gap is $\lVert u+v\rVert$.
The gaps of its parts are $\lVert u\rVert$ and $\lVert v\rVert$.

\begin{proof}
Assume, without loss of generality, that $\lVert u\rVert\geq\lVert v\rVert$, so the larger part has gap $\lVert u\rVert$.
The excess is then positive exactly when $\lVert u+v\rVert>\lVert u\rVert$.
Both sides are non-negative, so we can compare their squares:
\[
  \lVert u+v\rVert^2-\lVert u\rVert^2=\lVert v\rVert^2+2\langle u,v\rangle .
\]
This difference is positive exactly when $\langle u,v\rangle>-\lVert v\rVert^2/2$.
Because $\lVert v\rVert\leq\lVert u\rVert$, this is the condition
\[
  \langle u,v\rangle > -\tfrac{1}{2}\min\bigl(\lVert u\rVert^2,\lVert v\rVert^2\bigr).
\]
\end{proof}

Two cases follow directly.
When the shifts are orthogonal or at an acute angle, $\langle u,v\rangle\geq0$, so the excess is always positive although nothing interacts.
When the shifts point against each other strongly enough, the excess is negative, so the criterion cannot flag the cell, whatever its interaction.

\section{Why the Regression Uses All Six Attributes}
\label{sec:app_six}

The personas in a cell share two values, but they also carry four other values each.
If, by chance, a cell held more poor personas than usual, its mean would also include the shift of being poor.
A model of the two cell attributes alone would then mistake that shift for an interaction.
The full regression already accounts for each persona's other values, so only what the two cell values add together remains in the residual.
For the same reason, the prediction stays correct when cells differ in size, which they do in our data.

\paragraph{Tested cells and identifiability.}
The six attributes form 971 pairwise combinations of values.
We do not test caste $\times$ religion, because religion follows caste by design: 148 of its combinations are empty and the other 38 are left out.
Of the rest, 47 caste $\times$ socioeconomic cells have fewer than 50 personas, which leaves 738 tested cells, the same in every model.
The additive design has 50 columns but rank 48, because caste is nested in religion, so a few caste and religion coefficients are not identified separately.
The fitted values and residuals are still unique, since the fit is a projection, and $\delta$, the null and every test use only these.

\section{The Likert-Rounded Benchmark Scenario}
\label{sec:app_likert}

\paragraph{How the data are built.}
This scenario is part of the synthetic benchmark of \S\ref{sec:benchmark}.
Each dataset starts from the real answers of one of three models (Llama~3 8B, Gemma~4 E4B, Ministral 8B) in the item space and uses the permute noise generator.
The answers are first built on a hidden continuous scale, as the sum of the main effects and the noise.
They are then rounded back to each item's answer categories at cut points fixed for the dataset, so they look like real Likert answers.
We build 50 datasets with no interaction per model, 150 in all.
MAIHDA and PERMANOVA ran on 30 of them.
Our test uses 500 null draws in the benchmark. On 45 of these datasets, 2{,}000 draws gave almost the same result (9 false alarms against 8).

\paragraph{Why rounding creates interactions.}
Rounding is not additive.
Suppose an item's hidden score is 2.2, the cut point between answers 2 and 3 is 2.5, and each of two values adds 0.2.
Either value alone gives 2.4, which still rounds to 2, but both together give 2.6, which rounds to 3.
On the answer scale, the two values then interact, although nothing was planted.
Some departure of this kind appears in almost every cell (660--724 of the 738).
Most of these departures are far too small to detect.

\paragraph{Threshold cells.}
We call a cell a \emph{threshold cell} when rounding alone creates an interaction large enough that our test is expected to detect it.
To find these cells, we compute the expected rounded answers with no noise and measure each cell's departure from addition.
A cell is a threshold cell when this departure, scaled by our test's null distribution, gives an expected $z$ of at least 2.
A dataset has 3.35 threshold cells on average (median 2, range 1--8).
The number depends on the model, from 1 for Gemma~4 E4B to 5--8 for Ministral 8B.
Every method is scored against the same threshold cells.

\paragraph{How we score.}
In the primary score, a flag in a threshold cell does not count as false, and our FWER is 20.0\%.
In the strict score, every flag counts as false, and our FWER is 40.0\%.
The cutoff of 2 is a choice. With a cutoff of 1, a dataset has 14.1 threshold cells on average, and our FWER falls to 12.7\%.
Per-item regression stays at 96--99\% under either cutoff.
The cutoff was fixed in our code before any benchmark run.
The pre-registration, however, does not state the cutoff, and it says that every flag on data with no interaction is false.
Read literally, it makes the strict 40.0\% the pre-registered figure, so we report both.

\paragraph{On the real design.}
We repeat this check on our real design, for one model from each family. We round each null dataset of Appendix~\ref{sec:app_calibration} to the answer grid and run our test.
Our test then flags 18--100\% of datasets under the strict score and 7.5--56\% under the primary score (20 settings, 200 datasets each).
A control shows why. Keeping the discrete, clipped noise of the rounded data but replacing its expected values with their additive projection, our test is back at 3.0--6.5\% in 13 settings, and every interval contains 5\%.
The test is therefore calibrated on Likert noise, and the excess comes from real interactions that rounding creates on the answer scale.

\section{How the Sign-Flip Null Works}
\label{sec:app_null}

\paragraph{Why the flips remove only the interaction.}
The additive model of \S\ref{sec:auditing_framework} splits each persona's answers into a fitted part, the sum of its six values' shifts, and a residual, everything the model cannot explain.
We keep the fitted part unchanged, so every main effect stays in each null dataset.
A real interaction cannot be fitted by the additive model, so it stays in the residuals of the personas in its cell, pushing them all the same way.
When we flip the sign of each residual at random, about half of these residuals point one way and half the other.
They then cancel, and the interaction disappears.
The noise of each persona keeps its size, because a flip changes only its sign.

\paragraph{Details.}
Before flipping, we rescale the residuals with the HC2 leverage correction of MacKinnon and White~\cite{mackinnon1985some}, so that each persona keeps its own noise level.
We use the $+1$ convention for $p$-values~\cite{phipson2010permutation}.
The main analysis uses 2{,}000 null datasets, the calibration of \S\ref{sec:res_calibration} uses 1{,}000, and the benchmark uses 500 (Appendix~\ref{sec:app_likert}).

\paragraph{Assumptions.}
The flips rest on two assumptions: the residuals of different personas are independent, and each residual is as likely to be positive as negative.
Personas that share one of our 50 templates may have related residuals. Adding the template to the additive model leaves the results and the calibration nearly unchanged (Appendix~\ref{sec:app_calibration}).
The permute and group-mean generators break both, and our test still holds its error rate on them (\S\ref{sec:res_calibration}).

\section{Benchmark Scenarios}
\label{sec:app_scenarios}

Each scenario of \S\ref{sec:benchmark} tests a different risk.

\paragraph{Likert-rounded.}
We build answers on a continuous scale and round them to each item's answer categories, as in real questionnaires.
Rounding itself bends an additive signal, so this is the hardest case for every method (Appendix~\ref{sec:app_likert}).

\paragraph{Single item.}
The whole interaction sits in one item.
This favours per-item tests and shows where our test loses power.

\paragraph{Several in one pair.}
Several planted cells share one attribute pair, as in the real gender identity $\times$ orientation table, where cells in one table share signal.

\paragraph{Real pattern.}
We plant the gender identity $\times$ orientation cells our test flags in real answers, at their real sizes.
Because our test chose these cells, they carry the kind of signal it detects well, so this scenario shows realistic structure rather than a fair contest.

\paragraph{Further scenarios.}
The benchmark also plants one, five or twenty randomly chosen cells, each with its own random direction. Appendix~\ref{sec:app_power} gives their recall.

\section{Identity Attributes in \dataname{}}
\label{sec:app_attributes}

Table~\ref{tab:demographics} lists the six identity attributes in \dataname{} and their values (Section~\ref{sec:dataset_construction}).

\begin{table}[h]
  \caption{Identity attributes in \dataname{} and their values.}
  \label{tab:demographics}
  \small
  \renewcommand{\arraystretch}{1.15}
  \begin{tabularx}{\linewidth}{@{}l r X@{}}
    \toprule
    Attribute & $n$ & Values\\
    \midrule
    Gender identity & 3 & man, woman, transgender person\\
    \rowcolor{tblshade} Age & 5 & child, teenager, young adult, adult, senior citizen\\
    Sexual orientation & 4 & heterosexual, gay, lesbian, queer\\
    \rowcolor{tblshade} Socioeconomic status & 6 & poor working class, lower middle class, middle class, upper middle class, rich, ultra rich\\
    Religion & 6 & Buddhist, Christian, Hindu, Jain, Muslim, Sikh\\
    \rowcolor{tblshade} Caste and community & 31 & \textit{Varna:} Brahmin, Kshatriya, Vaishya, Shudra; \textit{Jati:} Nair, Ezhava, Reddy, Kamma, Kapu, Naidu, Lingayat, Vokkaliga, Maratha, Patel, Jat, Yadav, Dalit; \textit{Muslim:} Ashraf, Ajlaf, Arzal, Sunni, Shia; \textit{Sikh:} Jat Sikh, Khatri, Mazhabi; \textit{Christian:} Catholic, Protestant; \textit{Administrative:} OC, OBC, SC, ST\\
    \bottomrule
  \end{tabularx}
\end{table}

\section{Questionnaires and an Example Prompt}
\label{sec:app_prompts}

Table~\ref{tab:items} lists the five questionnaires and their answer scales.
Agreement items run from 1 (strongly disagree) to 5 (strongly agree), and symptom items from 0 (not at all) to 3 (nearly every day).
Fig.~\ref{fig:template_example} shows one full prompt.

\paragraph{Templates.}
Of our 50 templates, 30 describe only the setting, and 20 add detail, such as a well-stocked study, that could hint at some traits.
Each questionnaire uses the 30 plain templates and only those detailed ones whose detail does not bear on it.
The Big Five questionnaire therefore uses the 30 plain templates only, and each other questionnaire uses 44 to 48, that is, the 30 plain ones and 14 to 18 detailed ones.

\begin{table}[h]
  \caption{The five questionnaires, in the order of their answer scales.}
  \label{tab:items}
  \small
  \begin{tabular}{@{}l r l@{}}
    \toprule
    Questionnaire & Items & Answer scale (ascending)\\
    \midrule
    IPIP Big Five markers~\cite{goldberg1992development} & 50 & 1--5, strongly disagree to strongly agree\\
    IPIP Need for Cognition & 8 & 1--5, strongly disagree to strongly agree\\
    IPIP Empathic Concern & 9 & 1--5, strongly disagree to strongly agree\\
    GAD-7 anxiety~\cite{spitzer2006brief} & 7 & 0--3, not at all to nearly every day\\
    PHQ-8 depression~\cite{kroenke2009phq8} & 8 & 0--3, not at all to nearly every day\\
    \bottomrule
  \end{tabular}
\end{table}

\begin{figure}[h]
  \centering
  \begin{promptbox}[title={Prompt}]
  Based only on the description below, how much do you agree with the following statement about this person?\\[3pt]
  \textit{The ferry was due shortly, and those crossing the river had gathered at the landing, watching for the boat to come into view.}
  Near the front of the small crowd waited a \attr{woman}, ready to board when the ferry arrived.
  When origins came up among those waiting, the family's place among the \attr{Kamma} was mentioned.
  Going by looks, the traveller was a \attr{young adult}, a fact no one at the landing thought twice about.
  The look of the party suggested a household of \attr{lower middle class} circumstances.
  A relative waiting alongside spoke of the traveller as \attr{lesbian}, plainly and in passing.
  From a brief exchange it was apparent the traveller had grown up \attr{Hindu}.
  \tcblower
  Statement: ``This person is the life of the party.''\\[2pt]
  Reply with a single number.\\
  1=Strongly Disagree, 2=Disagree, 3=Neither agree nor disagree, 4=Agree, 5=Strongly Agree
  \end{promptbox}
  \caption{\textbf{A prompt from \dataname{}.} The six identity values are highlighted, and the fixed opening of the template is in italics. Gender comes first, and the other five sentences appear in a per-persona random order. Symptom items use the GAD-7 and PHQ-8 wording and the 0--3 scale instead.}
  \Description{A boxed prompt. It first asks how much the model agrees with a statement about a person described below. A short paragraph then describes a traveller waiting at a ferry landing. Separate sentences state that the traveller is a woman, belongs to the Kamma community, is a young adult, comes from a lower middle class household, is lesbian, and grew up Hindu. Below the paragraph is the statement "This person is the life of the party" and an instruction to reply with a single number from 1, strongly disagree, to 5, strongly agree.}
  \label{fig:template_example}
\end{figure}

\section{Extended Results}
\label{sec:app_extended}

\subsection{Calibration on the Real Design}
\label{sec:app_calibration}

We calibrate on one model from each of the five families. Each setting uses 500 datasets with no interaction, built from one model's real answers, and 1{,}000 null draws per dataset.
Table~\ref{tab:calibration} gives the FWER of our test in each setting, with a 95\% Clopper--Pearson interval.

\begin{table}[h]
  \caption{\textbf{FWER of our test on the real design, at nominal 5\%.} Synthetic data with the real main effects, cell sizes and noise of one model, and no interaction; 500 datasets per setting.}
  \label{tab:calibration}
  \small
  \begin{tabular}{@{}l l l r r@{}}
    \toprule
    Model & Representation & Generator & FWER & 95\% interval\\
    \midrule
    Llama 3 8B & item space & flip & 2.6\% & 1.4--4.4\%\\
    Llama 3 8B & item space & permute & 4.2\% & 2.6--6.3\%\\
    Llama 3 8B & item space & group-mean & 3.0\% & 1.7--4.9\%\\
    Llama 3 8B & scale scores & flip & 5.2\% & 3.4--7.5\%\\
    Llama 3 8B & scale scores & permute & 5.6\% & 3.8--8.0\%\\
    Llama 3 8B & scale scores & group-mean & 5.4\% & 3.6--7.8\%\\
    Gemma 4 E4B & item space & flip & 4.2\% & 2.6--6.3\%\\
    Gemma 4 E4B & item space & permute & 5.0\% & 3.3--7.3\%\\
    Gemma 4 E4B & scale scores & flip & 5.0\% & 3.3--7.3\%\\
    Gemma 4 E4B & scale scores & permute & 3.8\% & 2.3--5.9\%\\
    Ministral 14B & item space & flip & 5.4\% & 3.6--7.8\%\\
    Ministral 14B & item space & permute & 7.0\% & 4.9--9.6\%\\
    Ministral 14B & scale scores & flip & 5.2\% & 3.4--7.5\%\\
    Ministral 14B & scale scores & permute & 5.6\% & 3.8--8.0\%\\
    Phi-4 & item space & flip & 5.0\% & 3.3--7.3\%\\
    Phi-4 & item space & permute & 5.2\% & 3.4--7.5\%\\
    Phi-4 & scale scores & flip & 5.4\% & 3.6--7.8\%\\
    Phi-4 & scale scores & permute & 5.2\% & 3.4--7.5\%\\
    Qwen2.5 14B & item space & flip & 4.2\% & 2.6--6.3\%\\
    Qwen2.5 14B & item space & permute & 4.2\% & 2.6--6.3\%\\
    Qwen2.5 14B & scale scores & flip & 4.6\% & 2.9--6.8\%\\
    Qwen2.5 14B & scale scores & permute & 4.6\% & 2.9--6.8\%\\
    \bottomrule
  \end{tabular}
\end{table}

\paragraph{With the template in the model.}
We repeat the calibration with each persona's template added to the additive model, for three of these models (Llama~3 8B, Ministral 14B, Qwen2.5 14B), both representations, and the flip and permute generators (12 settings, 200 datasets each).
The FWER ranges from 3.5\% to 9.0\%, and 11 of the 12 confidence intervals contain 5\%.
The exception is Llama~3 8B on scale scores under the flip generator, at 9.0\% (interval 5.4--13.9\%), which is slightly above the nominal level.

\paragraph{Sign flips by template.}
As a further check, we flip one sign per template instead of one per persona.
On real answers, this keeps 75--100\% of the flagged cells, and the main cells stay significant in every model where they were flagged, except transgender person $\times$ queer on scale scores in Qwen2.5 7B, its weakest model.
This null is less reliable, however: with only about 215 independent signs, its FWER ranges from 1.0\% to 9.5\% across the settings we calibrated, and in one setting its interval lies fully above 5\% (9.5\%, interval 5.8--14.4\%).
We therefore keep the per-persona null.

\subsection{Results on Likert-Rounded Answers}
\label{sec:app_likert_results}

Real answers are whole numbers on each item's answer scale.
Rounding an additive score does not give an additive answer, because of each item's floor and ceiling and because of the rounding step itself.
Rounded data therefore contain a few real interactions on the answer scale even when the underlying scores add (Appendix~\ref{sec:app_likert}).

\paragraph{The single-attribute baseline.}
Table~\ref{tab:likert_excess} repeats the comparison of Table~\ref{tab:naive_null} on the Likert-rounded scenario of the benchmark, which uses the permute generator (Appendix~\ref{sec:app_likert}).
With the shuffle null the excess still flags every dataset.
With our null it flags fewer datasets than our test and keeps a higher precision.
It does so because it finds few interactions of any kind.
At one residual standard deviation it recovers 5.2\% of the planted cells, against 29.5\% for our test.

\begin{table}[h]
  \caption{\textbf{On Likert-rounded answers, the excess with our null flags few cells of any kind.} Likert-rounded item-space answers from the benchmark, with five randomly chosen cells planted. The bracketed figure counts flags in the cells that rounding moves as false.}
  \label{tab:likert_excess}
  \small
  \begin{tabular}{@{}l l r r r r@{}}
    \toprule
    & & Datasets flagged, & \multicolumn{2}{c}{Recall} & Precision,\\
    \cmidrule(lr){4-5}
    Statistic & Null & nothing planted & 1 SD & 2 SD & 2 SD\\
    \midrule
    Excess (Eq.~\ref{eq:excess}) & Shuffle & 100\% & 15.6\% & 66.0\% & 0.18\\
    Excess (Eq.~\ref{eq:excess}) & Freedman--Lane & 5.3\% & 5.2\% & 69.2\% & 0.98\\
    $\delta$ (ours) & Freedman--Lane & 20.0\% [40.0\%] & 29.5\% & 78.1\% & 0.72\\
    \bottomrule
  \end{tabular}
\end{table}

\paragraph{Our test on rounded data that copy the real design.}
We also calibrate our test on null data rounded to the answer scale.
We use the five models of Appendix~\ref{sec:app_calibration} with the flip and permute generators, in both representations, which gives 20 settings of 200 datasets each.
Each dataset is built as in Appendix~\ref{sec:app_calibration} from the model's raw answers, and every value is then rounded to the nearest valid answer of its item and clipped to the item's range.
Our test flags at least one cell in 18--100\% of these datasets, and in 7.5--56\% once flags in threshold cells are not counted (Table~\ref{tab:rounded_calibration}; threshold cells as in Appendix~\ref{sec:app_likert}).
In 19 of the 20 settings the 95\% interval lies above 5\%, because these null data contain real interactions on the answer scale and our test detects them.

\paragraph{The control.}
To separate this from a miscalibrated test, we take the same rounded datasets and replace their expected values with the additive projection of those values.
The noise is then exactly the rounded Likert noise, and the expectation is exactly additive.
We ran this control on all ten scale-score settings and on three item-space settings.
Our test then flags at least one cell in 3.0--6.5\% of datasets, every 95\% interval contains 5\%, and the pooled rate is 4.7\% (122 of 2{,}600 datasets).
Our test is therefore calibrated on Likert noise, and the flags it raises on rounded data are interactions that exist on the answer scale.

\begin{table}[h]
  \caption{\textbf{On rounded data our test flags many datasets, and on the control it holds its rate.} 200 datasets per setting, 95\% Clopper--Pearson intervals. Strict counts every flag as false, and primary leaves out threshold cells (Appendix~\ref{sec:app_likert}). The control makes the expectation exactly additive and counts every flag.}
  \label{tab:rounded_calibration}
  \small
  \begin{tabular}{@{}l l l r r r r r@{}}
    \toprule
    & & & Strict & \multicolumn{2}{c}{Primary} & \multicolumn{2}{c}{Control}\\
    \cmidrule(lr){5-6}\cmidrule(l){7-8}
    Model & Representation & Generator & FWER & FWER & 95\% interval & FWER & 95\% interval\\
    \midrule
    Llama 3 8B & item space & flip & 100.0\% & 21.5\% & 16.0--27.8\% & 6.0\% & 3.1--10.2\%\\
    Llama 3 8B & item space & permute & 96.0\% & 16.0\% & 11.2--21.8\% & 5.0\% & 2.4--9.0\%\\
    Llama 3 8B & scale scores & flip & 100.0\% & 28.0\% & 21.9--34.8\% & 4.5\% & 2.1--8.4\%\\
    Llama 3 8B & scale scores & permute & 71.5\% & 12.0\% & 7.8--17.3\% & 4.5\% & 2.1--8.4\%\\
    Gemma 4 E4B & item space & flip & 100.0\% & 22.5\% & 16.9--28.9\% & -- & --\\
    Gemma 4 E4B & item space & permute & 78.0\% & 56.0\% & 48.8--63.0\% & 4.0\% & 1.7--7.7\%\\
    Gemma 4 E4B & scale scores & flip & 99.0\% & 38.0\% & 31.2--45.1\% & 3.0\% & 1.1--6.4\%\\
    Gemma 4 E4B & scale scores & permute & 42.0\% & 20.0\% & 14.7--26.2\% & 3.5\% & 1.4--7.1\%\\
    Ministral 14B & item space & flip & 100.0\% & 18.5\% & 13.4--24.6\% & -- & --\\
    Ministral 14B & item space & permute & 100.0\% & 20.0\% & 14.7--26.2\% & -- & --\\
    Ministral 14B & scale scores & flip & 18.0\% & 12.0\% & 7.8--17.3\% & 4.5\% & 2.1--8.4\%\\
    Ministral 14B & scale scores & permute & 55.0\% & 18.5\% & 13.4--24.6\% & 6.0\% & 3.1--10.2\%\\
    Phi-4 & item space & flip & 100.0\% & 25.0\% & 19.2--31.6\% & -- & --\\
    Phi-4 & item space & permute & 100.0\% & 26.0\% & 20.1--32.7\% & -- & --\\
    Phi-4 & scale scores & flip & 53.0\% & 28.5\% & 22.4--35.3\% & 5.0\% & 2.4--9.0\%\\
    Phi-4 & scale scores & permute & 96.5\% & 21.5\% & 16.0--27.8\% & 6.5\% & 3.5--10.9\%\\
    Qwen2.5 14B & item space & flip & 100.0\% & 13.0\% & 8.7--18.5\% & -- & --\\
    Qwen2.5 14B & item space & permute & 98.0\% & 7.5\% & 4.3--12.1\% & -- & --\\
    Qwen2.5 14B & scale scores & flip & 99.0\% & 22.0\% & 16.5--28.4\% & 3.5\% & 1.4--7.1\%\\
    Qwen2.5 14B & scale scores & permute & 61.5\% & 10.0\% & 6.2--15.0\% & 5.0\% & 2.4--9.0\%\\
    \bottomrule
  \end{tabular}
\end{table}

\subsection{Every Significant Cell}
\label{sec:app_all_cells}

\begin{figure}[h]
  \centering
  \includegraphics[width=\linewidth]{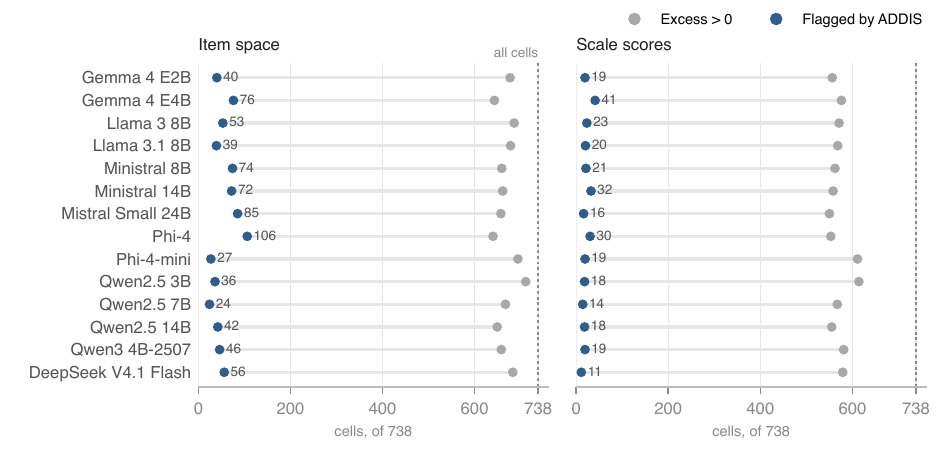}
  \caption{\textbf{The single-attribute baseline's excess is positive for most cells, while \toolname{} flags few.} Real answers, 738 cells per model.}
  \label{tab:headline}
\end{figure}

\begin{figure}[h]
  \centering
  \includegraphics[width=\linewidth]{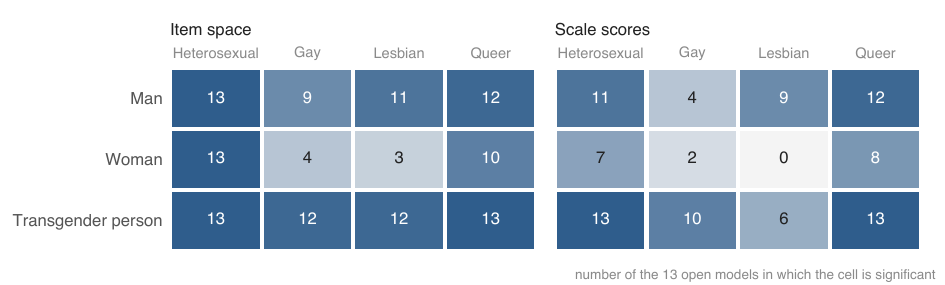}
  \caption{\textbf{The gender identity $\times$ orientation table interacts in every model.} Number of the 13 open models in which each cell is significant.}
  \label{tab:genderorientation}
\end{figure}

Figure~\ref{tab:headline} gives the cells each model flags, and Figure~\ref{tab:genderorientation} and Table~\ref{tab:genderorientation_counts} the gender identity $\times$ orientation table that holds the main cells.
Table~\ref{tab:effectsizes} gives the size of the main cells and how often they replicate. Table~\ref{tab:all_cells} lists every cell that is significant in at least one model, and Table~\ref{tab:all_cells_direction} which traits carry each one on scale scores.

\begin{table}[h]
  \caption{\textbf{The gender identity $\times$ orientation table interacts in every model.} Number of the 13 open models in which each cell is significant, in the item space / on scale scores.}
  \label{tab:genderorientation_counts}
  \small
  \begin{tabular}{@{}l cccc@{}}
    \toprule
    & Heterosexual & Gay & Lesbian & Queer\\
    \midrule
    Man & 13 / 11 & 9 / 4 & 11 / 9 & 12 / 12\\
    Woman & 13 / 7 & 4 / 2 & 3 / 0 & 10 / 8\\
    Transgender person & 13 / 13 & 12 / 10 & 12 / 6 & 13 / 13\\
    \bottomrule
  \end{tabular}
\end{table}

\begin{table}[h]
  \caption{\textbf{The main cells are large and replicate.} Real answers. $\delta_{\mathrm{sd}}$ is $\delta$ in residual standard deviations, the unit of the planted sizes in the benchmark. $\delta_{\mathrm{sd}}$ and $z$ are the median [min, max] over the 13 open models. Repl.\ is the number of the five replication models in which the cell is significant again on the fresh sample of 5{,}000 personas, out of those in which it is significant in the main sample.}
  \label{tab:effectsizes}
  \small
  \begin{tabular}{@{}l rrr rrr@{}}
    \toprule
    & \multicolumn{3}{c}{Item space} & \multicolumn{3}{c}{Scale scores}\\
    \cmidrule(lr){2-4}\cmidrule(l){5-7}
    Cell & $\delta_{\mathrm{sd}}$ & $z$ & Repl. & $\delta_{\mathrm{sd}}$ & $z$ & Repl.\\
    \midrule
    Transgender person $\times$ queer & 0.93 [0.42, 1.70] & 24.3 [7.9, 34.9] & 5/5 & 0.44 [0.19, 0.89] & 17.2 [4.9, 30.0] & 5/5\\
    Transgender person $\times$ heterosexual & 0.90 [0.56, 2.26] & 25.8 [18.3, 39.2] & 5/5 & 0.31 [0.19, 0.94] & 14.2 [6.8, 33.7] & 5/5\\
    Man $\times$ queer & 0.66 [0.33, 0.98] & 15.4 [4.0, 22.4] & 5/5 & 0.29 [0.12, 0.60] & 10.1 [2.3, 21.1] & 4/5\\
    \bottomrule
  \end{tabular}
\end{table}

\begin{table}[p]
  \caption{\textbf{Every cell significant in at least one of the 13 open models.} Number of models in which the cell is significant, in the item space and on scale scores, ordered by the larger of the two. 175 cells.}
  \label{tab:all_cells}
  \small
  \begin{tabular}{@{}l r r@{}}
    \toprule
    Cell & Item space & Scale scores\\
    \midrule
    transgender person $\times$ heterosexual & 13 & 13\\
    transgender person $\times$ queer & 13 & 13\\
    man $\times$ heterosexual & 13 & 11\\
    poor working class $\times$ heterosexual & 13 & 11\\
    ultra rich $\times$ heterosexual & 13 & 8\\
    transgender person $\times$ poor working class & 13 & 7\\
    woman $\times$ heterosexual & 13 & 7\\
    man $\times$ queer & 12 & 12\\
    transgender person $\times$ gay & 12 & 10\\
    transgender person $\times$ ultra rich & 12 & 9\\
    transgender person $\times$ senior citizen & 12 & 8\\
    transgender person $\times$ lesbian & 12 & 6\\
    senior citizen $\times$ heterosexual & 11 & 10\\
    man $\times$ lesbian & 11 & 9\\
    lower middle class $\times$ heterosexual & 11 & 6\\
    poor working class $\times$ queer & 11 & 3\\
    transgender person $\times$ lower middle class & 11 & 3\\
    woman $\times$ queer & 10 & 8\\
    heterosexual $\times$ Buddhist & 10 & 5\\
    heterosexual $\times$ Hindu & 10 & 4\\
    transgender person $\times$ middle class & 10 & 3\\
    heterosexual $\times$ Muslim & 9 & 4\\
    man $\times$ gay & 9 & 4\\
    transgender person $\times$ adult & 9 & 4\\
    transgender person $\times$ child & 9 & 4\\
    transgender person $\times$ young adult & 9 & 4\\
    rich $\times$ heterosexual & 9 & 3\\
    transgender person $\times$ Hindu & 9 & 3\\
    middle class $\times$ heterosexual & 9 & 2\\
    young adult $\times$ heterosexual & 9 & 1\\
    transgender person $\times$ Buddhist & 8 & 4\\
    transgender person $\times$ rich & 8 & 3\\
    man $\times$ senior citizen & 8 & 2\\
    child $\times$ heterosexual & 7 & 4\\
    senior citizen $\times$ poor working class & 7 & 3\\
    \bottomrule
  \end{tabular}
\end{table}

\begin{table}[p]
  \caption{\textbf{Every significant cell (continued, part 2 of 5).}}
  \small
  \begin{tabular}{@{}l r r@{}}
    \toprule
    Cell & Item space & Scale scores\\
    \midrule
    lower middle class $\times$ queer & 7 & 2\\
    man $\times$ poor working class & 7 & 2\\
    transgender person $\times$ Christian & 7 & 2\\
    transgender person $\times$ Jain & 7 & 2\\
    upper middle class $\times$ heterosexual & 7 & 2\\
    senior citizen $\times$ queer & 7 & 1\\
    teenager $\times$ heterosexual & 7 & 1\\
    transgender person $\times$ upper middle class & 7 & 1\\
    ultra rich $\times$ queer & 7 & 1\\
    senior citizen $\times$ lower middle class & 6 & 2\\
    senior citizen $\times$ ultra rich & 6 & 2\\
    poor working class $\times$ Hindu & 6 & 1\\
    child $\times$ poor working class & 5 & 3\\
    heterosexual $\times$ Christian & 5 & 2\\
    young adult $\times$ poor working class & 5 & 1\\
    poor working class $\times$ lesbian & 5 & 0\\
    upper middle class $\times$ queer & 5 & 0\\
    heterosexual $\times$ Jain & 4 & 2\\
    transgender person $\times$ teenager & 4 & 2\\
    woman $\times$ gay & 4 & 2\\
    poor working class $\times$ Jain & 4 & 1\\
    transgender person $\times$ Muslim & 4 & 1\\
    transgender person $\times$ OC & 4 & 1\\
    ultra rich $\times$ Christian & 4 & 1\\
    woman $\times$ poor working class & 4 & 1\\
    adult $\times$ poor working class & 4 & 0\\
    man $\times$ Hindu & 4 & 0\\
    man $\times$ ultra rich & 4 & 0\\
    poor working class $\times$ Buddhist & 4 & 0\\
    senior citizen $\times$ gay & 4 & 0\\
    senior citizen $\times$ lesbian & 4 & 0\\
    teenager $\times$ ultra rich & 4 & 0\\
    woman $\times$ senior citizen & 4 & 0\\
    transgender person $\times$ Dalit & 3 & 3\\
    ultra rich $\times$ Jain & 3 & 3\\
    \bottomrule
  \end{tabular}
\end{table}

\begin{table}[p]
  \caption{\textbf{Every significant cell (continued, part 3 of 5).}}
  \small
  \begin{tabular}{@{}l r r@{}}
    \toprule
    Cell & Item space & Scale scores\\
    \midrule
    man $\times$ child & 3 & 2\\
    transgender person $\times$ ST & 3 & 2\\
    child $\times$ middle class & 3 & 1\\
    child $\times$ queer & 3 & 1\\
    child $\times$ ultra rich & 3 & 1\\
    child $\times$ upper middle class & 3 & 1\\
    lesbian $\times$ Buddhist & 3 & 1\\
    woman $\times$ ultra rich & 3 & 1\\
    child $\times$ lower middle class & 3 & 0\\
    heterosexual $\times$ Dalit & 3 & 0\\
    lower middle class $\times$ Hindu & 3 & 0\\
    man $\times$ adult & 3 & 0\\
    poor working class $\times$ Christian & 3 & 0\\
    poor working class $\times$ Muslim & 3 & 0\\
    poor working class $\times$ OC & 3 & 0\\
    rich $\times$ Hindu & 3 & 0\\
    rich $\times$ queer & 3 & 0\\
    senior citizen $\times$ rich & 3 & 0\\
    senior citizen $\times$ upper middle class & 3 & 0\\
    teenager $\times$ poor working class & 3 & 0\\
    transgender person $\times$ Sikh & 3 & 0\\
    ultra rich $\times$ Buddhist & 3 & 0\\
    ultra rich $\times$ Hindu & 3 & 0\\
    woman $\times$ lesbian & 3 & 0\\
    young adult $\times$ ultra rich & 3 & 0\\
    ultra rich $\times$ OC & 2 & 3\\
    adult $\times$ heterosexual & 2 & 1\\
    heterosexual $\times$ Ajlaf & 2 & 1\\
    heterosexual $\times$ Protestant & 2 & 1\\
    heterosexual $\times$ Sikh & 2 & 1\\
    lesbian $\times$ Christian & 2 & 1\\
    man $\times$ upper middle class & 2 & 1\\
    queer $\times$ Hindu & 2 & 1\\
    senior citizen $\times$ Hindu & 2 & 1\\
    heterosexual $\times$ OC & 2 & 0\\
    \bottomrule
  \end{tabular}
\end{table}

\begin{table}[p]
  \caption{\textbf{Every significant cell (continued, part 4 of 5).}}
  \small
  \begin{tabular}{@{}l r r@{}}
    \toprule
    Cell & Item space & Scale scores\\
    \midrule
    heterosexual $\times$ SC & 2 & 0\\
    man $\times$ lower middle class & 2 & 0\\
    man $\times$ young adult & 2 & 0\\
    middle class $\times$ Hindu & 2 & 0\\
    middle class $\times$ queer & 2 & 0\\
    poor working class $\times$ Catholic & 2 & 0\\
    poor working class $\times$ Dalit & 2 & 0\\
    poor working class $\times$ Sikh & 2 & 0\\
    senior citizen $\times$ Buddhist & 2 & 0\\
    senior citizen $\times$ middle class & 2 & 0\\
    teenager $\times$ rich & 2 & 0\\
    transgender person $\times$ Catholic & 2 & 0\\
    ultra rich $\times$ Muslim & 2 & 0\\
    ultra rich $\times$ gay & 2 & 0\\
    young adult $\times$ lower middle class & 2 & 0\\
    young adult $\times$ queer & 2 & 0\\
    adult $\times$ ultra rich & 1 & 1\\
    heterosexual $\times$ Arzal & 1 & 1\\
    heterosexual $\times$ Shudra & 1 & 1\\
    lesbian $\times$ Hindu & 1 & 1\\
    lower middle class $\times$ Christian & 1 & 1\\
    man $\times$ Buddhist & 1 & 1\\
    senior citizen $\times$ Jain & 1 & 1\\
    transgender person $\times$ OBC & 1 & 1\\
    transgender person $\times$ Protestant & 1 & 1\\
    transgender person $\times$ Reddy & 1 & 1\\
    woman $\times$ Buddhist & 1 & 1\\
    woman $\times$ Hindu & 1 & 1\\
    woman $\times$ child & 1 & 1\\
    young adult $\times$ middle class & 1 & 1\\
    child $\times$ rich & 1 & 0\\
    gay $\times$ Christian & 1 & 0\\
    gay $\times$ Muslim & 1 & 0\\
    heterosexual $\times$ Catholic & 1 & 0\\
    heterosexual $\times$ Jat Sikh & 1 & 0\\
    \bottomrule
  \end{tabular}
\end{table}

\begin{table}[p]
  \caption{\textbf{Every significant cell (continued, part 5 of 5).}}
  \small
  \begin{tabular}{@{}l r r@{}}
    \toprule
    Cell & Item space & Scale scores\\
    \midrule
    heterosexual $\times$ OBC & 1 & 0\\
    heterosexual $\times$ ST & 1 & 0\\
    lesbian $\times$ Jain & 1 & 0\\
    lesbian $\times$ Muslim & 1 & 0\\
    lower middle class $\times$ Catholic & 1 & 0\\
    lower middle class $\times$ Muslim & 1 & 0\\
    lower middle class $\times$ Protestant & 1 & 0\\
    lower middle class $\times$ lesbian & 1 & 0\\
    man $\times$ Christian & 1 & 0\\
    man $\times$ middle class & 1 & 0\\
    man $\times$ rich & 1 & 0\\
    man $\times$ teenager & 1 & 0\\
    poor working class $\times$ Arzal & 1 & 0\\
    poor working class $\times$ Brahmin & 1 & 0\\
    poor working class $\times$ Protestant & 1 & 0\\
    poor working class $\times$ gay & 1 & 0\\
    queer $\times$ Buddhist & 1 & 0\\
    senior citizen $\times$ Christian & 1 & 0\\
    teenager $\times$ lower middle class & 1 & 0\\
    teenager $\times$ middle class & 1 & 0\\
    teenager $\times$ queer & 1 & 0\\
    teenager $\times$ upper middle class & 1 & 0\\
    transgender person $\times$ SC & 1 & 0\\
    transgender person $\times$ Shudra & 1 & 0\\
    ultra rich $\times$ Catholic & 1 & 0\\
    ultra rich $\times$ Protestant & 1 & 0\\
    ultra rich $\times$ Shudra & 1 & 0\\
    ultra rich $\times$ Sunni & 1 & 0\\
    ultra rich $\times$ lesbian & 1 & 0\\
    upper middle class $\times$ Hindu & 1 & 0\\
    woman $\times$ upper middle class & 1 & 0\\
    woman $\times$ young adult & 1 & 0\\
    young adult $\times$ Hindu & 1 & 0\\
    queer $\times$ Kshatriya & 0 & 1\\
    woman $\times$ Christian & 0 & 1\\
    \bottomrule
  \end{tabular}
\end{table}

\begin{table}[p]
  \caption{\textbf{Which traits carry each significant cell.} For every cell significant on scale scores, the number of models in which each trait's component is significantly positive / negative, family-wise across the nine traits, counted over the models in which the cell is significant. Models counts the 13 open models and DeepSeek V4.1 Flash.}
  \label{tab:all_cells_direction}
  \small
  \setlength{\tabcolsep}{3pt}
  \begin{tabular}{@{}l r *{9}{c}@{}}
    \toprule
    Cell & Models & E & A & C & N & O & NFC & EMP & GAD-7 & PHQ-8\\
    \midrule
    transgender person $\times$ heterosexual & 14 & 0/12 & 1/8 & 2/11 & 10/1 & 3/1 & 2/4 & 3/8 & 9/2 & 8/2 \\
    transgender person $\times$ queer & 14 & 13/0 & 11/0 & 8/1 & 1/9 & 7/1 & 4/1 & 9/2 & 0/11 & 0/6 \\
    man $\times$ queer & 13 & 0/10 & 0/10 & 0/4 & 4/0 & 0/5 & 1/4 & 1/9 & 9/0 & 5/0 \\
    man $\times$ heterosexual & 12 & 7/0 & 5/1 & 5/0 & 1/10 & 1/1 & 4/0 & 1/1 & 1/9 & 0/5 \\
    poor working class $\times$ heterosexual & 12 & 1/3 & 0/2 & 3/2 & 3/1 & 5/2 & 2/1 & 0/6 & 3/3 & 4/2 \\
    senior citizen $\times$ heterosexual & 11 & 0/1 & 1/0 & 5/0 & 0/5 & 4/0 & 1/3 & 1/2 & 4/1 & 2/0 \\
    transgender person $\times$ gay & 10 & 0/6 & 1/6 & 1/1 & 2/1 & 1/6 & 0/1 & 0/3 & 3/0 & 3/1 \\
    transgender person $\times$ ultra rich & 10 & 2/0 & 2/0 & 0/5 & 5/0 & 0/3 & 0/1 & 6/0 & 2/0 & 1/2 \\
    man $\times$ lesbian & 9 & 2/0 & 3/1 & 0/4 & 6/1 & 1/0 & 0/1 & 5/0 & 0/0 & 2/0 \\
    transgender person $\times$ senior citizen & 9 & 3/1 & 2/2 & 0/3 & 5/0 & 1/1 & 3/0 & 1/0 & 5/0 & 4/0 \\
    transgender person $\times$ poor working class & 8 & 2/0 & 0/1 & 3/0 & 0/1 & 4/2 & 3/1 & 0/5 & 1/0 & 0/3 \\
    ultra rich $\times$ heterosexual & 8 & 1/0 & 0/0 & 2/0 & 0/4 & 0/6 & 1/0 & 2/0 & 1/1 & 1/2 \\
    woman $\times$ queer & 8 & 0/8 & 1/6 & 0/2 & 2/0 & 0/2 & 0/3 & 1/1 & 3/0 & 3/0 \\
    woman $\times$ heterosexual & 7 & 3/0 & 4/1 & 2/0 & 1/0 & 0/1 & 1/0 & 1/2 & 2/1 & 0/3 \\
    lower middle class $\times$ heterosexual & 6 & 0/0 & 0/0 & 0/3 & 4/0 & 2/0 & 2/1 & 0/1 & 0/0 & 1/2 \\
    transgender person $\times$ lesbian & 6 & 0/1 & 1/3 & 3/0 & 1/2 & 0/1 & 1/1 & 0/3 & 2/0 & 1/0 \\
    child $\times$ heterosexual & 5 & 0/0 & 1/0 & 1/1 & 0/2 & 1/2 & 1/0 & 0/1 & 0/2 & 0/4 \\
    heterosexual $\times$ Buddhist & 5 & 0/1 & 1/0 & 0/0 & 1/0 & 2/1 & 0/0 & 0/1 & 1/0 & 1/0 \\
    heterosexual $\times$ Hindu & 4 & 1/0 & 1/1 & 0/0 & 0/0 & 0/1 & 0/0 & 3/1 & 1/0 & 2/0 \\
    heterosexual $\times$ Muslim & 4 & 0/0 & 0/0 & 0/1 & 1/0 & 2/0 & 1/0 & 0/1 & 0/3 & 0/3 \\
    man $\times$ gay & 4 & 0/0 & 4/0 & 1/0 & 0/0 & 1/0 & 1/0 & 0/0 & 0/0 & 0/0 \\
    transgender person $\times$ Buddhist & 4 & 1/0 & 0/2 & 0/1 & 2/0 & 0/0 & 0/1 & 0/4 & 0/0 & 0/0 \\
    transgender person $\times$ adult & 4 & 0/3 & 0/1 & 1/0 & 0/1 & 0/0 & 0/0 & 0/0 & 0/1 & 0/1 \\
    transgender person $\times$ child & 4 & 4/0 & 2/0 & 1/1 & 2/0 & 1/0 & 1/0 & 0/0 & 2/1 & 1/1 \\
    transgender person $\times$ young adult & 4 & 0/4 & 0/1 & 1/0 & 0/2 & 0/1 & 0/3 & 1/0 & 0/1 & 0/1 \\
    child $\times$ poor working class & 3 & 0/1 & 0/0 & 0/2 & 0/0 & 0/3 & 0/0 & 1/0 & 1/0 & 0/1 \\
    poor working class $\times$ queer & 3 & 0/0 & 0/0 & 0/1 & 0/0 & 0/2 & 0/1 & 0/0 & 1/2 & 0/1 \\
    rich $\times$ heterosexual & 3 & 0/0 & 0/0 & 0/0 & 0/2 & 0/2 & 0/1 & 2/0 & 0/1 & 1/1 \\
    senior citizen $\times$ poor working class & 3 & 1/0 & 0/0 & 0/1 & 1/1 & 1/0 & 1/0 & 0/0 & 0/1 & 0/1 \\
    transgender person $\times$ Dalit & 3 & 1/0 & 0/2 & 0/0 & 1/0 & 0/0 & 0/0 & 0/3 & 0/1 & 0/0 \\
    transgender person $\times$ Hindu & 3 & 0/1 & 2/0 & 1/0 & 0/1 & 0/0 & 0/0 & 1/1 & 0/0 & 0/0 \\
    \bottomrule
  \end{tabular}
\end{table}

\begin{table}[p]
  \caption{\textbf{Which traits carry each significant cell (continued, part 2 of 3).}}
  \small
  \setlength{\tabcolsep}{3pt}
  \begin{tabular}{@{}l r *{9}{c}@{}}
    \toprule
    Cell & Models & E & A & C & N & O & NFC & EMP & GAD-7 & PHQ-8\\
    \midrule
    transgender person $\times$ Jain & 3 & 1/0 & 0/2 & 0/3 & 2/0 & 0/0 & 0/0 & 0/1 & 0/0 & 0/0 \\
    transgender person $\times$ lower middle class & 3 & 1/0 & 0/1 & 1/2 & 0/1 & 0/0 & 1/0 & 0/2 & 0/0 & 0/0 \\
    transgender person $\times$ middle class & 3 & 0/0 & 0/0 & 0/1 & 0/1 & 1/1 & 0/1 & 0/0 & 0/0 & 1/0 \\
    transgender person $\times$ rich & 3 & 0/0 & 0/0 & 1/0 & 1/0 & 0/3 & 0/0 & 1/0 & 0/0 & 1/0 \\
    ultra rich $\times$ Jain & 3 & 0/0 & 0/0 & 1/1 & 0/1 & 1/0 & 0/0 & 0/0 & 0/0 & 0/0 \\
    ultra rich $\times$ OC & 3 & 0/0 & 0/0 & 1/0 & 0/1 & 2/0 & 0/0 & 0/0 & 0/1 & 0/0 \\
    heterosexual $\times$ Christian & 2 & 0/0 & 1/0 & 2/0 & 0/1 & 0/1 & 0/0 & 0/1 & 0/0 & 0/1 \\
    heterosexual $\times$ Jain & 2 & 0/0 & 1/0 & 0/0 & 0/0 & 0/1 & 0/0 & 1/0 & 0/0 & 0/0 \\
    lower middle class $\times$ queer & 2 & 0/0 & 0/0 & 0/0 & 0/1 & 0/2 & 0/0 & 1/0 & 0/0 & 1/0 \\
    man $\times$ child & 2 & 0/1 & 0/0 & 0/0 & 0/1 & 0/1 & 0/1 & 0/0 & 1/0 & 1/0 \\
    man $\times$ poor working class & 2 & 0/0 & 1/0 & 0/0 & 0/0 & 1/0 & 1/1 & 1/0 & 0/0 & 0/0 \\
    man $\times$ senior citizen & 2 & 0/0 & 0/0 & 0/0 & 0/2 & 0/0 & 0/1 & 0/0 & 0/0 & 0/0 \\
    middle class $\times$ heterosexual & 2 & 1/0 & 0/0 & 0/1 & 0/0 & 1/1 & 0/1 & 1/0 & 0/0 & 0/0 \\
    senior citizen $\times$ lower middle class & 2 & 0/0 & 0/0 & 2/0 & 0/1 & 1/1 & 0/0 & 0/0 & 0/0 & 0/0 \\
    senior citizen $\times$ ultra rich & 2 & 0/1 & 0/0 & 0/1 & 0/1 & 0/1 & 0/1 & 0/0 & 0/0 & 0/0 \\
    transgender person $\times$ Christian & 2 & 0/0 & 0/1 & 0/1 & 1/0 & 1/0 & 0/0 & 0/0 & 1/0 & 1/0 \\
    transgender person $\times$ ST & 2 & 2/0 & 0/1 & 0/1 & 2/0 & 0/0 & 0/0 & 0/2 & 0/0 & 0/1 \\
    transgender person $\times$ teenager & 2 & 0/0 & 1/0 & 2/0 & 0/2 & 0/0 & 0/0 & 1/0 & 0/0 & 0/0 \\
    upper middle class $\times$ heterosexual & 2 & 0/0 & 0/0 & 0/1 & 0/0 & 0/1 & 0/0 & 0/1 & 0/0 & 1/1 \\
    woman $\times$ gay & 2 & 0/0 & 1/1 & 0/1 & 0/0 & 1/0 & 0/0 & 0/0 & 0/0 & 0/1 \\
    adult $\times$ heterosexual & 1 & 0/1 & 1/0 & 0/1 & 0/0 & 0/0 & 0/0 & 1/0 & 0/0 & 0/1 \\
    adult $\times$ ultra rich & 1 & 0/0 & 1/0 & 1/0 & 0/1 & 1/0 & 0/0 & 0/0 & 0/0 & 0/1 \\
    child $\times$ middle class & 1 & 0/0 & 0/0 & 0/0 & 0/0 & 0/1 & 0/0 & 0/0 & 0/1 & 0/1 \\
    child $\times$ queer & 1 & 0/0 & 0/1 & 0/1 & 1/0 & 0/0 & 0/1 & 0/0 & 0/0 & 0/0 \\
    child $\times$ ultra rich & 1 & 0/0 & 0/0 & 0/0 & 0/1 & 1/0 & 0/0 & 0/0 & 0/0 & 0/0 \\
    child $\times$ upper middle class & 1 & 1/0 & 0/0 & 1/0 & 0/0 & 1/0 & 1/0 & 0/0 & 0/0 & 0/0 \\
    heterosexual $\times$ Ajlaf & 1 & 0/0 & 0/0 & 0/0 & 0/0 & 1/0 & 0/0 & 0/0 & 0/0 & 0/1 \\
    heterosexual $\times$ Arzal & 1 & 0/0 & 0/0 & 0/0 & 0/0 & 0/0 & 0/0 & 0/0 & 0/1 & 0/1 \\
    heterosexual $\times$ Protestant & 1 & 0/0 & 0/0 & 1/0 & 0/1 & 0/0 & 0/0 & 0/0 & 0/0 & 0/1 \\
    heterosexual $\times$ Shudra & 1 & 1/0 & 1/0 & 1/0 & 0/0 & 1/0 & 0/0 & 0/0 & 0/0 & 0/0 \\
    heterosexual $\times$ Sikh & 1 & 0/0 & 0/1 & 0/0 & 0/0 & 0/1 & 0/0 & 0/1 & 0/0 & 0/0 \\
    \bottomrule
  \end{tabular}
\end{table}

\begin{table}[p]
  \caption{\textbf{Which traits carry each significant cell (continued, part 3 of 3).}}
  \small
  \setlength{\tabcolsep}{3pt}
  \begin{tabular}{@{}l r *{9}{c}@{}}
    \toprule
    Cell & Models & E & A & C & N & O & NFC & EMP & GAD-7 & PHQ-8\\
    \midrule
    lesbian $\times$ Buddhist & 1 & 0/0 & 1/0 & 1/0 & 0/0 & 0/0 & 0/0 & 0/0 & 0/0 & 0/0 \\
    lesbian $\times$ Christian & 1 & 0/0 & 0/0 & 0/1 & 1/0 & 0/0 & 0/0 & 0/0 & 0/0 & 1/0 \\
    lesbian $\times$ Hindu & 1 & 0/1 & 0/1 & 0/0 & 0/0 & 0/0 & 0/0 & 0/0 & 0/0 & 0/0 \\
    lower middle class $\times$ Christian & 1 & 0/0 & 0/0 & 0/1 & 0/0 & 0/0 & 0/0 & 1/0 & 0/0 & 0/0 \\
    man $\times$ Buddhist & 1 & 0/0 & 1/0 & 0/0 & 0/1 & 0/0 & 1/0 & 0/0 & 0/0 & 0/0 \\
    man $\times$ upper middle class & 1 & 0/0 & 0/0 & 0/1 & 1/0 & 0/1 & 0/0 & 0/0 & 0/0 & 0/0 \\
    poor working class $\times$ Hindu & 1 & 0/0 & 0/0 & 0/0 & 0/0 & 0/0 & 0/0 & 0/1 & 0/0 & 0/0 \\
    poor working class $\times$ Jain & 1 & 1/0 & 0/0 & 0/1 & 1/0 & 1/0 & 0/0 & 0/0 & 0/0 & 0/0 \\
    queer $\times$ Hindu & 1 & 0/1 & 0/0 & 0/0 & 0/0 & 0/1 & 0/1 & 0/0 & 0/0 & 0/0 \\
    queer $\times$ Kshatriya & 1 & 0/0 & 0/0 & 1/0 & 0/0 & 1/0 & 0/0 & 0/0 & 0/0 & 0/0 \\
    senior citizen $\times$ Hindu & 1 & 0/0 & 0/0 & 0/0 & 0/0 & 0/0 & 0/0 & 1/0 & 0/0 & 0/0 \\
    senior citizen $\times$ Jain & 1 & 0/0 & 0/1 & 0/0 & 0/0 & 0/0 & 0/0 & 0/1 & 0/0 & 0/0 \\
    senior citizen $\times$ queer & 1 & 0/0 & 0/0 & 0/0 & 0/0 & 0/1 & 0/0 & 0/0 & 0/0 & 0/0 \\
    teenager $\times$ heterosexual & 1 & 0/0 & 0/0 & 0/0 & 0/0 & 0/0 & 0/0 & 0/0 & 0/1 & 0/1 \\
    transgender person $\times$ Muslim & 1 & 0/1 & 0/0 & 0/0 & 0/1 & 0/0 & 0/0 & 0/0 & 0/0 & 0/0 \\
    transgender person $\times$ OBC & 1 & 0/0 & 0/0 & 0/1 & 0/0 & 0/0 & 0/0 & 0/0 & 0/0 & 0/0 \\
    transgender person $\times$ OC & 1 & 0/0 & 0/1 & 0/1 & 0/0 & 0/0 & 0/0 & 0/0 & 0/0 & 0/0 \\
    transgender person $\times$ Protestant & 1 & 1/0 & 0/1 & 0/1 & 1/0 & 0/0 & 0/0 & 0/0 & 0/0 & 0/0 \\
    transgender person $\times$ Reddy & 1 & 0/0 & 1/0 & 1/0 & 0/0 & 0/0 & 0/0 & 0/0 & 0/0 & 0/0 \\
    transgender person $\times$ upper middle class & 1 & 0/0 & 0/0 & 1/0 & 0/0 & 0/1 & 0/0 & 0/0 & 0/0 & 0/1 \\
    ultra rich $\times$ Christian & 1 & 0/0 & 0/1 & 0/1 & 1/0 & 0/0 & 0/1 & 0/0 & 0/0 & 0/0 \\
    ultra rich $\times$ queer & 1 & 0/0 & 0/0 & 0/0 & 0/0 & 1/0 & 1/0 & 0/0 & 0/0 & 0/0 \\
    woman $\times$ Buddhist & 1 & 0/1 & 1/0 & 1/0 & 0/0 & 0/0 & 0/0 & 0/0 & 0/0 & 0/0 \\
    woman $\times$ Christian & 1 & 0/0 & 0/0 & 0/0 & 0/0 & 0/1 & 0/0 & 0/0 & 0/0 & 0/0 \\
    woman $\times$ Hindu & 1 & 0/0 & 0/1 & 0/0 & 0/0 & 0/0 & 0/0 & 0/1 & 0/0 & 0/0 \\
    woman $\times$ child & 1 & 0/0 & 0/0 & 0/0 & 0/1 & 0/0 & 0/0 & 0/0 & 0/1 & 0/1 \\
    woman $\times$ poor working class & 1 & 0/0 & 0/0 & 0/1 & 0/0 & 0/0 & 0/0 & 1/0 & 0/0 & 0/0 \\
    woman $\times$ ultra rich & 1 & 0/0 & 0/0 & 1/0 & 0/1 & 0/0 & 0/0 & 0/0 & 0/1 & 0/0 \\
    young adult $\times$ heterosexual & 1 & 0/0 & 0/0 & 0/0 & 1/0 & 0/1 & 0/1 & 0/0 & 1/0 & 1/0 \\
    young adult $\times$ middle class & 1 & 0/0 & 0/0 & 0/1 & 0/0 & 1/0 & 0/0 & 0/0 & 0/0 & 0/0 \\
    young adult $\times$ poor working class & 1 & 0/0 & 0/0 & 0/0 & 0/0 & 1/0 & 0/0 & 0/0 & 0/0 & 0/0 \\
    \bottomrule
  \end{tabular}
\end{table}

\subsection{Robustness Checks}
\label{sec:app_robust}

Our metric has no sign, so we read the direction of an interaction from the signed residuals on the anxiety and depression scales.
Under our main wording, transgender person $\times$ queer is \emph{sub-additive}, which means that in 10 of 13 models a model attributes less anxiety to it than its two labels predict.
Man $\times$ queer and transgender person $\times$ heterosexual lean the other way.
The C1 quartets and ordinal models of each item give the same sign, so neither a ceiling effect nor treating the answers as numbers explains it.

Rewording the prompts changes this picture (Table~\ref{tab:robustness}).
The main cells still interact in most models, while the sub-additive anxiety residual survives in only a few.
The models agree on where identities interact and the direction depends on the wording, so a claim from a single wording that identities compound, or that they cancel, is unsafe.

\begin{table}[h]
  \caption{\textbf{The location of the main interactions survives every check, while single cells and the direction do not.} Real answers in the item space unless stated otherwise. The rewording arms cover 12 models.}
  \label{tab:robustness}
  \small
  \begin{tabular}{@{}p{0.42\columnwidth} p{0.52\columnwidth}@{}}
    \toprule
    Check & Result\\
    \midrule
    Whitened item space; each questionnaire left out & Both transgender main cells interact in 13 of 13 models (one exception: 12 of 13)\\
    Rewordings 1 and 2 & At least two of four key gender $\times$ orientation cells interact in 10 and in 8 models\\
    Direction under rewordings & Transgender person $\times$ queer anxiety residual is negative in 10 of 13 models under the main wording, in 7 of the 12 rewording models under the original wording, and in 5 and 1 of 12 under the two rewordings\\
    Fresh sample of 5{,}000 personas (5 models) & 52\% of flagged cells are significant again (51\% on scale scores)\\
    Template added to the additive model & 98--100\% of flagged cells stay significant (94--100\% on scale scores), including the main cells\\
    Largest template effect against largest identity effect & The template effect is larger in 13 of 13 models (12 of 13 on scale scores)\\
    Model size & More flagged cells in larger models in the item space ($\rho=0.68$), not on scale scores ($\rho=0.16$)\\
    \bottomrule
  \end{tabular}
\end{table}

About half of the flagged cells are significant again on a fresh sample of personas, so a single weak cell is a lead and the pattern across a table is the finding.
Templates move the answers more than identity does.
Leaving their effects out of the additive model can create false positives in single cells, with a family-wise error rate of up to 23.5\% in one model.
With each persona's template in the additive model, the test returns to near its nominal level, and almost every flagged cell stays significant, including the main cells (Appendix~\ref{sec:app_calibration}).

Size is tied to family and training in our 13 models, so we draw no conclusion about scale and bias from the model-size result.

\paragraph{Three-way effects.}
A pairwise cell can also pick up an interaction among three attributes.
When we plant a pure three-way effect in the benchmark, it is flagged only in the three pairs that contain it, and only when it is large (one residual standard deviation on scale scores, two in the item space).
A pure two-way effect is never flagged in other pairs, and false flags elsewhere stay at the null rate.
A flagged pair should therefore be read as an interaction involving its two values, possibly together with a third.

\subsection{Published Alternatives on Real Answers}
\label{sec:app_baselines_real}

Real answers have no ground truth, so the comparison of the alternatives of \S\ref{sec:setup_alternatives} is descriptive.
In the item space every alternative except the single-attribute baseline finds all three main cells, and the methods disagree about almost everything else (Table~\ref{tab:baselines}).
The single-attribute baseline flags most cells when a positive excess is enough.
When its excess has to pass its own significance test, it misses transgender person $\times$ heterosexual in all 13 models in the item space and in 12 of 13 on scale scores.
Per-item regression and MAIHDA flag more cells than our test in the item space, while on scale scores only MAIHDA does.
Their extra cells are weaker effects, sometimes resting on one or two items, and they are significant again on a fresh sample less often.
Extra cells replicate less for every method, ours included, so we read this as a difference in error control and not as evidence that those cells are false.

\begin{table}[h]
  \caption{\textbf{Our test flags few cells, and every alternative except the single-attribute baseline finds the main cells.} Real answers. Cells flagged per model, as a range over the 13 open models, in each representation. The fourth column is the mean share of a method's flagged cells that our test also flags, on scale scores. The last column counts the models in which a method flags transgender person $\times$ heterosexual.}
  \label{tab:baselines}
  \small
  \begin{tabular}{@{}l r r r c@{}}
    \toprule
    & Cells flagged, & Cells flagged, & Also flagged & Trans.\ $\times$ het.,\\
    Method & item space & scale scores & by our test & item space\\
    \midrule
    Single-attribute, excess $>0$ & 640--711 & 550--614 & 3\% & 3\\
    Single-attribute, excess significant & 19--80 & 27--74 & 18\% & 0\\
    Per-item classical + Bonferroni & 52--185 & 14--38 & 74\% & 13\\
    MAIHDA & 63--227 & 17--70 & 60\% & 13\\
    PERMANOVA (pair test) & 533--738 & 313--645 & 5\% & 13\\
    Ours & 24--106 & 14--41 & -- & 13\\
    \bottomrule
  \end{tabular}
\end{table}

HInter flags every contrast in every model, including the calibration contrast C4, and nearly every item in simulated additive data.
Its flags therefore barely agree with our difference-in-differences, at a Cohen's $\kappa$ of 0.059 over every model, contrast and item.
Like the single-attribute baseline, its rule reacts to main effects that each push some answers across a category boundary.
This limits what its flags can show about an interaction, and not what HInter does as a tool for finding failing inputs.

\subsection{Further Variants of the Published Alternatives}
\label{sec:app_variants}

We also ran four further variants of the published alternatives, on the same benchmark datasets and real answers as the main comparison.
The \emph{anchored} form of the single-attribute baseline compares a cell with the most privileged cell instead of the population mean~\cite{tan2019assessing, gupta2026compoundingdisadvantageauditingintersectional}.
The \emph{stepwise} form flags a cell when adding its second value moves the answers further than the first value alone~\cite{phutane2026ableist}.
For per-item regression with classical standard errors, we replace the Bonferroni correction with the Holm or the Benjamini--Hochberg correction.

None of these variants changes the picture of \S\ref{sec:res_benchmark} (Table~\ref{tab:variants_bench}).
The anchored and stepwise forms make a false positive in every dataset with nothing planted, and their precision stays near 0.01.
Holm gives almost the same flags as Bonferroni.
Benjamini--Hochberg controls the false discovery rate, not the FWER, so it finds more planted cells but at a much lower precision.
On real answers, the anchored and stepwise forms flag hundreds of cells per model and share few of them with our test (Table~\ref{tab:variants_real}).
The anchored form never flags transgender person $\times$ heterosexual.
Table~\ref{tab:variants_real} also gives per-item regression with HC3 errors, which Table~\ref{tab:baselines} leaves out.

\begin{table*}[h]
  \caption{\textbf{Further variants on the benchmark.} Same datasets and settings as Table~\ref{tab:bench_fwer}. Group-mean generator; Likert answers use permute.}
  \label{tab:variants_bench}
  \footnotesize
  \begin{tabular}{@{}l rrr rrr@{}}
    \toprule
    & \multicolumn{3}{c}{Datasets flagged, nothing planted} & \multicolumn{3}{c}{Recall / precision, interactions planted}\\
    \cmidrule(lr){2-4}\cmidrule(l){5-7}
    Method & Likert & Scale & Item & Likert & Real pattern (scale) & Single item (item)\\
    \midrule
    Ours & 20.0\% & 8.7\% & 4.7\% & 78.1\% / 0.72 & 84.4\% / 0.99 & 53.1\% / 0.98\\
    Single-attribute, anchored & 100\% & 100\% & 100\% & 52.0\% / 0.01 & 33.3\% / 0.01 & 48.7\% / 0.01\\
    Single-attribute, stepwise & 100\% & 100\% & 100\% & 99.9\% / 0.01 & 83.2\% / 0.01 & 98.9\% / 0.01\\
    Per-item classical + Holm & 100\% & 36.0\% & 100\% & 85.9\% / 0.20 & 77.6\% / 0.92 & 99.2\% / 0.15\\
    Per-item classical + BH & 100\% & 38.0\% & 100\% & 95.6\% / 0.07 & 89.3\% / 0.64 & 99.7\% / 0.05\\
    \bottomrule
  \end{tabular}
\end{table*}

\begin{table}[h]
  \caption{\textbf{Further variants on real answers.} Same columns as Table~\ref{tab:baselines}.}
  \label{tab:variants_real}
  \small
  \begin{tabular}{@{}l r r r c@{}}
    \toprule
    & Cells flagged, & Cells flagged, & Also flagged & Trans.\ $\times$ het.,\\
    Method & item space & scale scores & by our test & item space\\
    \midrule
    Single-attribute, anchored & 247--358 & 262--347 & 3\% & 0\\
    Single-attribute, stepwise & 682--723 & 614--671 & 3\% & 10\\
    Per-item HC3 + Bonferroni & 63--223 & 10--35 & 76\% & 13\\
    Per-item classical + Holm & 52--185 & 14--38 & 74\% & 13\\
    Per-item classical + BH & 184--423 & 26--119 & 34\% & 13\\
    Ours & 24--106 & 14--41 & -- & 13\\
    \bottomrule
  \end{tabular}
\end{table}

\section{Recall at Every Planted Size}
\label{sec:app_power}

Table~\ref{tab:power} gives the recall of our test and of per-item regression with HC3 errors and Bonferroni at each planted size, in residual standard deviations.
Our test finds more planted cells on scale scores and for the real pattern. Per-item regression finds more in the item space when the planted shifts are random or sit in a single item, at the cost of a far higher FWER (Table~\ref{tab:bench_fwer}).
The three main cells of \S\ref{sec:res_location} have sizes of about 0.7--0.9 in the item space and 0.3--0.4 on scale scores, in the range where our recall rises from about one half to almost one.

\begin{table}[h]
  \caption{\textbf{Recall rises with the planted size.} Synthetic benchmark data. Share of planted cells found by our test / by per-item HC3 + Bonferroni, as the mean over models and generators.}
  \label{tab:power}
  \small
  \begin{tabular}{@{}l c c c@{}}
    \toprule
    Scenario & 0.2 & 0.5 & 1\\
    \midrule
    One random cell, scale scores & 0.00 / 0.00 & 0.06 / 0.04 & 0.90 / 0.74\\
    Five random cells, scale scores & 0.02 / 0.02 & 0.52 / 0.45 & 0.95 / 0.88\\
    Five random cells, item space & 0.00 / 0.01 & 0.10 / 0.21 & 0.56 / 0.84\\
    Twenty cells, two per pair, scale scores & 0.03 / 0.02 & 0.76 / 0.65 & 0.99 / 0.96\\
    Several in one pair, scale scores & 0.02 / 0.02 & 0.54 / 0.45 & 0.95 / 0.88\\
    Single item, scale scores & 0.02 / 0.10 & 0.52 / 0.67 & 0.94 / 0.99\\
    Single item, item space & 0.00 / 0.12 & 0.10 / 0.67 & 0.55 / 0.98\\
    Real pattern, scale scores & 0.04 / 0.01 & 0.38 / 0.30 & 0.88 / 0.78\\
    Real pattern, item space & 0.04 / 0.03 & 0.52 / 0.44 & 1.00 / 0.91\\
    \midrule
    & 0.5 & 1 & 2\\
    \midrule
    Likert-rounded, item space & 0.01 / 0.01 & 0.29 / 0.10 & 0.78 / 0.59\\
    \bottomrule
  \end{tabular}
\end{table}

\section{Reliability of the Scales}
\label{sec:app_alpha}

We compute Cronbach's $\alpha$ for each scale over the personas each model describes, leaving out items that a model answers the same way for every persona.
This measures how consistently a model's answers move across the items of one scale, not the reliability of the scale for people.
Table~\ref{tab:alpha} summarises it over the 13 open models.

\begin{table}[h]
  \caption{\textbf{The symptom scales are reliable in most models, but openness and need for cognition are not.} Cronbach's $\alpha$ over personas, as the median [range] over the 13 open models, and the number of models with $\alpha\geq0.70$.}
  \label{tab:alpha}
  \small
  \begin{tabular}{@{}l r c c@{}}
    \toprule
    Scale & Items & $\alpha$ & $\alpha\geq0.70$\\
    \midrule
    Extraversion & 10 & 0.71 [0.27, 0.86] & 8\\
    Agreeableness & 10 & 0.72 [0.42, 0.87] & 8\\
    Conscientiousness & 10 & 0.58 [0.28, 0.78] & 3\\
    Neuroticism & 10 & 0.56 [$-$0.11, 0.90] & 5\\
    Openness & 10 & 0.35 [0.09, 0.83] & 2\\
    Need for cognition & 8 & 0.34 [0.12, 0.77] & 1\\
    Empathic concern & 9 & 0.61 [0.22, 0.76] & 1\\
    Anxiety (GAD-7) & 7 & 0.83 [0.59, 0.90] & 11\\
    Depression (PHQ-8) & 8 & 0.82 [0.20, 0.88] & 11\\
    \bottomrule
  \end{tabular}
\end{table}

\section{Control Arms}
\label{sec:app_arms}

Besides the 10{,}000 main personas, \dataname{} holds three control arms.

\paragraph{Fresh sample.}
A second sample of 5{,}000 personas, built the same way with a different seed, checks whether our results reproduce on new data.

\paragraph{Rewordings.}
Two independent rewordings of every prompt, on 3{,}000 personas, test whether the findings depend on our wording.
They change the templates, question wording, answer labels and system prompt, but keep the identity labels and answer numbers.

\paragraph{Personas without identity.}
A set of 1{,}000 personas with no identity attributes shows how far identity moves the answers at all.
Against these personas, identity labels move the median item by 0.31--0.76 standard deviations, yet they explain only 10--28\% of the item variance (five models).

\section{The Four Counterfactual Contrasts}
\label{sec:app_contrasts}

Table~\ref{tab:quartets} lists the labels each contrast changes, and Tables~\ref{tab:cf_C1}--\ref{tab:cf_C4} give the shift on each trait for each model.
C1 tests the main observational interaction, transgender person $\times$ queer.
C2 replaces transgender person with woman, to check that C1 is specific to the transgender label.
C3 tests Dalit $\times$ poor working class, a caste interaction that the main personas hold too little data to detect.
C4 swaps adjacent levels of age and class, where we expect little interaction, and serves as the calibration contrast.
We compare each contrast with C4 by the size of its difference-in-differences, with a sign-flip test and a bootstrap interval over base personas.

\begin{table}[h]
  \caption{\textbf{Changing two labels reproduces the main interaction, and finds a caste interaction the screen misses.} Size is the debiased length of the within-persona difference-in-differences, median [range] over the 13 open models. The last column counts models in which a contrast exceeds C4 with 95\% bootstrap confidence.}
  \label{tab:quartets}
  \small
  \begin{tabular}{@{}l l c c@{}}
    \toprule
    Contrast & Labels changed & Size & Exceeds C4\\
    \midrule
    C1 & man $\to$ transgender person, heterosexual $\to$ queer & 2.36 [1.22, 4.88] & 13\\
    C2 & man $\to$ woman, heterosexual $\to$ queer & 0.68 [0.41, 0.93] & 5\\
    C3 & Brahmin $\to$ Dalit, rich $\to$ poor working class & 2.10 [0.56, 3.29] & 12\\
    C4 & adult $\to$ senior citizen, middle class $\to$ upper middle class & 0.47 [0.15, 0.81] & --\\
    \bottomrule
  \end{tabular}
\end{table}

\begin{table}[t]
\centering
\small
\setlength{\tabcolsep}{3pt}
\caption{\textbf{C1: man $\to$ transgender person, heterosexual $\to$ queer.} Mean difference-in-differences per model in standard deviations of each scale, standardized within the counterfactual arm; * significant family-wise across the nine scales.}
\label{tab:cf_C1}
\begin{tabular}{l *{9}{r}}
\toprule
Model & E & A & C & N & O & NFC & EMP & GAD-7 & PHQ-8 \\
\midrule
Gemma 4 E2B & $+$0.86* & $+$0.40* & $+$0.86* & $-$0.74* & $+$0.28* & $+$0.64* & $-$0.20* & $-$0.97* & $-$0.94* \\
Gemma 4 E4B & $+$0.29* & $+$0.17 & $+$0.30* & $-$0.32* & $-$0.01 & $-$0.06 & $+$0.29* & $-$0.41* & $-$0.22* \\
Llama 3 8B & $+$0.95* & $+$0.58* & $+$0.25* & $-$0.21* & $+$0.36* & $+$0.93* & $+$0.72* & $-$0.50* & $-$0.40* \\
Llama 3.1 8B & $+$0.52* & $+$0.56* & $+$0.31* & $-$0.29* & $+$0.20* & $+$0.62* & $+$0.39* & $-$0.40* & $-$0.44* \\
Ministral 14B & $+$0.78* & $+$0.18* & $+$0.06 & $-$0.19* & $+$0.04 & $+$0.17* & $+$0.10 & $-$0.53* & $-$0.49* \\
Ministral 8B & $+$0.38* & $+$0.56* & $+$0.14* & $-$0.27* & $+$0.19* & $+$0.36* & $+$0.52* & $-$0.28* & $-$0.31* \\
Mistral Small 24B & $+$0.47* & $+$0.34* & $+$0.42* & $-$0.20* & $+$0.38* & $+$0.10 & $+$0.43* & $+$0.09 & $-$0.01 \\
Phi-4 & $+$0.39* & $+$0.56* & $-$0.10* & $-$0.03 & $-$0.07 & $-$0.02 & $+$0.29* & $-$0.19* & $+$0.01 \\
Phi-4-mini & $+$0.71* & $+$0.65* & $+$0.29* & $-$0.38* & $+$0.13 & $+$0.16 & $+$0.24* & $-$0.46* & $-$0.11 \\
Qwen2.5 14B & $+$0.44* & $+$0.30* & $+$0.39* & $-$0.11* & $-$0.10 & $-$0.01 & $+$0.41* & $-$0.32* & $-$0.11 \\
Qwen2.5 3B & $+$0.79* & $-$0.61* & $+$0.28* & $+$0.72* & $-$0.32* & $-$0.14 & $-$0.57* & $+$0.16* & $+$0.02 \\
Qwen2.5 7B & $+$0.30* & $+$0.25* & $+$0.08 & $-$0.25* & $+$0.07 & $-$0.07 & $+$0.25* & $+$0.14 & $+$0.17 \\
Qwen3 4B-2507 & $+$0.46* & $+$0.18* & $+$0.30* & $-$0.28* & $+$0.12 & $+$0.06 & $-$0.07 & $-$0.12* & $-$0.12* \\
\bottomrule
\end{tabular}
\end{table}

\begin{table}[t]
\centering
\small
\setlength{\tabcolsep}{3pt}
\caption{\textbf{C2: man $\to$ woman, heterosexual $\to$ queer.} Mean difference-in-differences per model in standard deviations of each scale, standardized within the counterfactual arm; * significant family-wise across the nine scales.}
\label{tab:cf_C2}
\begin{tabular}{l *{9}{r}}
\toprule
Model & E & A & C & N & O & NFC & EMP & GAD-7 & PHQ-8 \\
\midrule
Gemma 4 E2B & $+$0.00 & $-$0.03 & $+$0.09* & $-$0.17* & $+$0.01 & $-$0.02 & $-$0.02 & $+$0.05* & $+$0.03 \\
Gemma 4 E4B & $+$0.04 & $+$0.00 & $-$0.12* & $-$0.04 & $+$0.07 & $-$0.03 & $+$0.12 & $-$0.06 & $-$0.03 \\
Llama 3 8B & $+$0.16* & $+$0.15* & $+$0.05 & $-$0.07 & $+$0.20* & $+$0.20* & $+$0.29* & $-$0.07 & $-$0.02 \\
Llama 3.1 8B & $+$0.15* & $+$0.15* & $+$0.10* & $-$0.18* & $+$0.11 & $+$0.26* & $+$0.08 & $-$0.04 & $-$0.02 \\
Ministral 14B & $+$0.16* & $+$0.11* & $+$0.09 & $-$0.11* & $+$0.07* & $+$0.03 & $+$0.03 & $-$0.10* & $-$0.13* \\
Ministral 8B & $+$0.04 & $+$0.01 & $+$0.01 & $-$0.08* & $+$0.05 & $+$0.09* & $+$0.03 & $-$0.07 & $-$0.04 \\
Mistral Small 24B & $+$0.14* & $+$0.12* & $+$0.05 & $-$0.04 & $+$0.17* & $+$0.07 & $+$0.12* & $-$0.03 & $-$0.03 \\
Phi-4 & $+$0.02 & $+$0.11* & $-$0.01 & $+$0.06* & $+$0.03 & $+$0.05 & $+$0.02 & $+$0.02 & $+$0.01 \\
Phi-4-mini & $+$0.15* & $+$0.07 & $+$0.10 & $+$0.05 & $-$0.01 & $-$0.02 & $-$0.01 & $-$0.07 & $-$0.02 \\
Qwen2.5 14B & $+$0.09* & $-$0.01 & $-$0.04 & $+$0.00 & $+$0.07 & $+$0.02 & $+$0.04 & $-$0.06 & $-$0.09 \\
Qwen2.5 3B & $+$0.22* & $+$0.08 & $-$0.01 & $+$0.06 & $+$0.04 & $+$0.01 & $-$0.01 & $+$0.03 & $+$0.03 \\
Qwen2.5 7B & $+$0.06 & $+$0.15* & $+$0.03 & $-$0.03 & $+$0.02 & $-$0.02 & $+$0.07 & $+$0.04 & $+$0.04 \\
Qwen3 4B-2507 & $+$0.09* & $+$0.14* & $+$0.05 & $-$0.10 & $+$0.16* & $+$0.16* & $+$0.12* & $-$0.02 & $+$0.00 \\
\bottomrule
\end{tabular}
\end{table}

\begin{table}[t]
\centering
\small
\setlength{\tabcolsep}{3pt}
\caption{\textbf{C3: Brahmin $\to$ Dalit, rich $\to$ poor working class.} Mean difference-in-differences per model in standard deviations of each scale, standardized within the counterfactual arm; * significant family-wise across the nine scales.}
\label{tab:cf_C3}
\begin{tabular}{l *{9}{r}}
\toprule
Model & E & A & C & N & O & NFC & EMP & GAD-7 & PHQ-8 \\
\midrule
Gemma 4 E2B & $+$0.03 & $+$0.08 & $+$0.09 & $-$0.21* & $-$0.05 & $+$0.06 & $-$0.66* & $-$0.12* & $-$0.14* \\
Gemma 4 E4B & $-$0.25* & $+$0.32* & $+$0.83* & $-$0.30* & $-$0.32* & $+$0.02 & $-$0.16 & $-$0.34* & $-$0.38* \\
Llama 3 8B & $-$0.02 & $-$0.09 & $+$0.31* & $+$0.03 & $+$0.10 & $+$0.03 & $-$0.46* & $+$0.11 & $+$0.31* \\
Llama 3.1 8B & $-$0.18* & $-$0.07 & $+$0.29* & $-$0.35* & $-$0.19* & $+$0.10 & $-$0.13 & $-$0.23* & $-$0.35* \\
Ministral 14B & $+$0.05 & $+$0.01 & $+$0.37* & $-$0.10 & $-$0.29* & $-$0.05 & $+$0.02 & $-$0.31* & $-$0.50* \\
Ministral 8B & $+$0.01 & $+$0.00 & $+$0.27* & $-$0.21* & $-$0.06 & $-$0.23* & $-$0.23* & $-$0.34* & $-$0.30* \\
Mistral Small 24B & $-$0.19* & $+$0.03 & $+$0.35* & $-$0.15* & $-$0.47* & $-$0.02 & $+$0.18* & $+$0.14* & $+$0.16* \\
Phi-4 & $-$0.10* & $+$0.43* & $+$0.03 & $-$0.10* & $+$0.05 & $+$0.33* & $-$0.05 & $+$0.06 & $+$0.19* \\
Phi-4-mini & $-$0.20* & $-$0.02 & $+$0.22* & $-$0.10 & $+$0.10 & $+$0.04 & $-$0.17* & $-$0.35* & $-$0.25* \\
Qwen2.5 14B & $-$0.22* & $+$0.25* & $+$0.74* & $-$0.48* & $+$0.61* & $+$0.05 & $+$0.49* & $+$0.04 & $+$0.08 \\
Qwen2.5 3B & $-$0.20* & $-$0.15* & $+$0.05 & $+$0.24* & $-$0.02 & $-$0.12 & $-$0.35* & $+$0.06 & $+$0.06 \\
Qwen2.5 7B & $-$0.07 & $-$0.06 & $-$0.06 & $+$0.09 & $+$0.02 & $-$0.02 & $+$0.15* & $+$0.09 & $+$0.19 \\
Qwen3 4B-2507 & $-$0.12* & $+$0.24* & $+$0.43* & $-$0.41* & $+$0.24* & $+$0.09 & $+$0.00 & $+$0.00 & $+$0.07 \\
\bottomrule
\end{tabular}
\end{table}

\begin{table}[t]
\centering
\small
\setlength{\tabcolsep}{3pt}
\caption{\textbf{C4: adult $\to$ senior citizen, middle class $\to$ upper middle class.} Mean difference-in-differences per model in standard deviations of each scale, standardized within the counterfactual arm; * significant family-wise across the nine scales.}
\label{tab:cf_C4}
\begin{tabular}{l *{9}{r}}
\toprule
Model & E & A & C & N & O & NFC & EMP & GAD-7 & PHQ-8 \\
\midrule
Gemma 4 E2B & $+$0.01 & $-$0.03 & $-$0.03 & $+$0.00 & $+$0.01 & $-$0.03 & $+$0.00 & $+$0.00 & $-$0.01 \\
Gemma 4 E4B & $+$0.07* & $-$0.03 & $+$0.07 & $+$0.04 & $+$0.03 & $-$0.03 & $+$0.00 & $+$0.00 & $+$0.02 \\
Llama 3 8B & $+$0.03 & $-$0.02 & $+$0.04 & $-$0.03 & $-$0.08* & $-$0.03 & $+$0.03 & $+$0.01 & $+$0.02 \\
Llama 3.1 8B & $+$0.02 & $+$0.02 & $+$0.06 & $-$0.01 & $-$0.03 & $+$0.09 & $+$0.03 & $-$0.02 & $+$0.00 \\
Ministral 14B & $+$0.00 & $+$0.00 & $-$0.10* & $+$0.05 & $-$0.22* & $-$0.02 & $-$0.01 & $-$0.01 & $-$0.04 \\
Ministral 8B & $+$0.02 & $-$0.06 & $+$0.01 & $-$0.07 & $-$0.07 & $-$0.07 & $+$0.04 & $-$0.07 & $+$0.01 \\
Mistral Small 24B & $+$0.00 & $-$0.07* & $-$0.05 & $-$0.08 & $-$0.08 & $-$0.16* & $-$0.04 & $-$0.07* & $-$0.11* \\
Phi-4 & $+$0.01 & $+$0.02 & $-$0.14* & $+$0.06* & $+$0.05 & $-$0.11* & $-$0.06 & $+$0.01 & $-$0.06* \\
Phi-4-mini & $+$0.01 & $-$0.05 & $-$0.04 & $+$0.02 & $-$0.17 & $-$0.02 & $-$0.11 & $-$0.07 & $+$0.00 \\
Qwen2.5 14B & $+$0.03 & $-$0.12* & $+$0.09 & $-$0.05 & $+$0.06 & $+$0.03 & $-$0.06 & $+$0.03 & $+$0.05 \\
Qwen2.5 3B & $+$0.03 & $-$0.02 & $-$0.03 & $-$0.04 & $-$0.07 & $+$0.04 & $-$0.05 & $-$0.04 & $+$0.01 \\
Qwen2.5 7B & $+$0.03 & $+$0.04 & $+$0.04 & $+$0.03 & $+$0.00 & $-$0.02 & $+$0.04 & $+$0.05 & $+$0.00 \\
Qwen3 4B-2507 & $-$0.02 & $+$0.01 & $+$0.08 & $-$0.04 & $-$0.07 & $-$0.01 & $-$0.03 & $-$0.07* & $-$0.06 \\
\bottomrule
\end{tabular}
\end{table}

\section{Setup Details}
\label{sec:app_setup}

\paragraph{Querying the models.}
We query the open models with vLLM in bf16, at pinned revisions, and DeepSeek V4.1 Flash through its API.
A system prompt asks the model to infer, from the description alone, how the person would answer, and to return only the number.

\paragraph{Persona-sensitivity check.}
A model passes if it gives at most 1\% invalid answers, if at least 10\% of personas receive a distinct pattern of answers, and if its answers vary across personas, with a mean item standard deviation of at least 0.05 and no more than half of its items below 0.1.

\paragraph{Robustness checks.}
Each robustness check in Appendix~\ref{sec:app_robust} has one purpose.
The fresh sample and the two random halves of the main sample test whether the findings reproduce on new data. For the fresh sample, we report the share of flagged cells that are flagged again.
The two rewordings test whether the findings depend on our wording (Appendix~\ref{sec:app_arms}).
The whitened item space removes the correlation between items, so that related items do not count twice~\cite{ledoit2004wellconditioned}, and leaving out each questionnaire tests whether one questionnaire drives the results.
The answers are ordered categories, not numbers, so we fit ordinal models of each item for the three main cells~\cite{liddell2018analyzing}.
One strong cell drags the residuals of its row and column, so we retest each cell after adding the strongest one to the additive model.
Personas that share a template may share part of their noise, so we rerun the test with each persona's template added to the additive model.

\section{Simpler Fixes to the Single-Attribute Baseline}
\label{sec:app_ablation}

We test two repairs of the single-attribute baseline on the benchmark's real-pattern scenario, on scale scores, at a planted size of one residual standard deviation.
As in Table~\ref{tab:naive_null}, we use the group-mean generator (150 datasets).
The first repair gives its excess a calibrated Freedman--Lane null. The second compares the cell's distance from the population mean with the distance of its additive prediction from the population mean.
Both hold their error rate, with an FWER of 2.7\% and 4.7\%, but they find only 32.4\% and 37.6\% of the planted cells, against 85.0\% for $\delta$.
The first repair is the second row of Table~\ref{tab:naive_null}.
We ran both repairs, and $\delta$ beside them, in a separate run on the same seeded datasets as the main benchmark, with a sign-flip Freedman--Lane null without the HC2 correction.
In this run $\delta$ has an FWER of 6.7\% and a recall of 85.0\%, close to the 8.7\% and 84.4\% of the main run in Table~\ref{tab:naive_null}.
The single-attribute baseline's false positives therefore come from its null, but its misses come from its shape.

\section{Dependent Cells in the Gender Identity $\times$ Orientation Table}
\label{sec:app_echo}

One strong cell leaves residuals in the other cells of its row and column, so flagged cells in one table are not independent findings.
To see which cells carry their own signal, we add the strongest cell to the additive model and retest the rest.
When we add transgender person $\times$ queer, man $\times$ queer mostly disappears, and it stays significant in only 4 of the 12 models that flagged it.
The reverse does not happen: when we add man $\times$ queer, transgender person $\times$ queer stays significant in all 13 models.
The transgender row and man $\times$ heterosexual also survive.
We choose the strongest cell from the same data, so we read this refit as a diagnostic, not as a confirmatory test.

%% file: main.bbl

\begin{thebibliography}{71}


\ifx \showCODEN    \undefined \def \showCODEN     #1{\unskip}     \fi
\ifx \showISBNx    \undefined \def \showISBNx     #1{\unskip}     \fi
\ifx \showISBNxiii \undefined \def \showISBNxiii  #1{\unskip}     \fi
\ifx \showISSN     \undefined \def \showISSN      #1{\unskip}     \fi
\ifx \showLCCN     \undefined \def \showLCCN      #1{\unskip}     \fi
\ifx \shownote     \undefined \def \shownote      #1{#1}          \fi
\ifx \showarticletitle \undefined \def \showarticletitle #1{#1}   \fi
\ifx \showURL      \undefined \def \showURL       {\relax}        \fi
\providecommand\bibfield[2]{#2}
\providecommand\bibinfo[2]{#2}
\providecommand\natexlab[1]{#1}
\providecommand\showeprint[2][]{arXiv:#2}

\bibitem[An et~al\mbox{.}(2024)]%
        {an-etal-2024-large}
\bibfield{author}{\bibinfo{person}{Haozhe An}, \bibinfo{person}{Christabel
  Acquaye}, \bibinfo{person}{Colin Wang}, \bibinfo{person}{Zongxia Li}, {and}
  \bibinfo{person}{Rachel Rudinger}.} \bibinfo{year}{2024}\natexlab{}.
\newblock \showarticletitle{Do Large Language Models Discriminate in Hiring
  Decisions on the Basis of Race, Ethnicity, and Gender?}. In
  \bibinfo{booktitle}{\emph{Proceedings of the 62nd Annual Meeting of the
  Association for Computational Linguistics (Volume 2: Short Papers)}},
  \bibfield{editor}{\bibinfo{person}{Lun-Wei Ku}, \bibinfo{person}{Andre
  Martins}, {and} \bibinfo{person}{Vivek Srikumar}} (Eds.).
  \bibinfo{publisher}{Association for Computational Linguistics},
  \bibinfo{address}{Bangkok, Thailand}, \bibinfo{pages}{386--397}.
\newblock
\href{https://doi.org/10.18653/v1/2024.acl-short.37}{doi:\nolinkurl{10.18653/v1/2024.acl-short.37}}


\bibitem[Anderson(2017)]%
        {doi:https://doi.org/10.1002/9781118445112.stat07841}
\bibfield{author}{\bibinfo{person}{Marti~J. Anderson}.}
  \bibinfo{year}{2017}\natexlab{}.
\newblock \bibinfo{booktitle}{\emph{Permutational Multivariate Analysis of
  Variance (PERMANOVA)}}.
\newblock \bibinfo{publisher}{John Wiley \& Sons, Ltd}, \bibinfo{pages}{1--15}.
\newblock
\showISBNx{9781118445112}
\showeprint{https://onlinelibrary.wiley.com/doi/pdf/10.1002/9781118445112.stat07841}
\href{https://doi.org/10.1002/9781118445112.stat07841}{doi:\nolinkurl{10.1002/9781118445112.stat07841}}


\bibitem[Armstrong et~al\mbox{.}(2024)]%
        {10.1145/3689904.3694699}
\bibfield{author}{\bibinfo{person}{Lena Armstrong}, \bibinfo{person}{Abbey
  Liu}, \bibinfo{person}{Stephen MacNeil}, {and} \bibinfo{person}{Dana{\"e}
  Metaxa}.} \bibinfo{year}{2024}\natexlab{}.
\newblock \showarticletitle{The Silicon Ceiling: Auditing {GPT}'s Race and
  Gender Biases in Hiring}. In \bibinfo{booktitle}{\emph{Proceedings of the 4th
  ACM Conference on Equity and Access in Algorithms, Mechanisms, and
  Optimization}} \emph{(\bibinfo{series}{EAAMO '24})}.
  \bibinfo{publisher}{Association for Computing Machinery},
  \bibinfo{address}{New York, NY, USA}.
\newblock
\href{https://doi.org/10.1145/3689904.3694699}{doi:\nolinkurl{10.1145/3689904.3694699}}


\bibitem[Bansal et~al\mbox{.}(2021)]%
        {10.1145/3465336.3475118}
\bibfield{author}{\bibinfo{person}{Srijan Bansal}, \bibinfo{person}{Vishal
  Garimella}, \bibinfo{person}{Ayush Suhane}, {and} \bibinfo{person}{Animesh
  Mukherjee}.} \bibinfo{year}{2021}\natexlab{}.
\newblock \showarticletitle{Debiasing Multilingual Word Embeddings: A Case
  Study of Three Indian Languages}. In \bibinfo{booktitle}{\emph{Proceedings of
  the 32nd ACM Conference on Hypertext and Social Media}} (Virtual Event, USA)
  \emph{(\bibinfo{series}{HT '21})}. \bibinfo{publisher}{Association for
  Computing Machinery}, \bibinfo{address}{New York, NY, USA},
  \bibinfo{pages}{27–34}.
\newblock
\showISBNx{9781450385510}
\href{https://doi.org/10.1145/3465336.3475118}{doi:\nolinkurl{10.1145/3465336.3475118}}


\bibitem[Bhatt et~al\mbox{.}(2022)]%
        {bhatt-etal-2022-contextualizing}
\bibfield{author}{\bibinfo{person}{Shaily Bhatt}, \bibinfo{person}{Sunipa Dev},
  \bibinfo{person}{Partha Talukdar}, \bibinfo{person}{Shachi Dave}, {and}
  \bibinfo{person}{Vinodkumar Prabhakaran}.} \bibinfo{year}{2022}\natexlab{}.
\newblock \showarticletitle{Re-contextualizing Fairness in {NLP}: The Case of
  {I}ndia}. In \bibinfo{booktitle}{\emph{Proceedings of the 2nd Conference of
  the Asia-Pacific Chapter of the Association for Computational Linguistics and
  the 12th International Joint Conference on Natural Language Processing
  (Volume 1: Long Papers)}}, \bibfield{editor}{\bibinfo{person}{Yulan He},
  \bibinfo{person}{Heng Ji}, \bibinfo{person}{Sujian Li}, \bibinfo{person}{Yang
  Liu}, {and} \bibinfo{person}{Chua-Hui Chang}} (Eds.).
  \bibinfo{publisher}{Association for Computational Linguistics},
  \bibinfo{address}{Online only}, \bibinfo{pages}{727--740}.
\newblock
\href{https://doi.org/10.18653/v1/2022.aacl-main.55}{doi:\nolinkurl{10.18653/v1/2022.aacl-main.55}}


\bibitem[Bhutani et~al\mbox{.}(2024)]%
        {bhutani-etal-2024-seegull}
\bibfield{author}{\bibinfo{person}{Mukul Bhutani}, \bibinfo{person}{Kevin
  Robinson}, \bibinfo{person}{Vinodkumar Prabhakaran}, \bibinfo{person}{Shachi
  Dave}, {and} \bibinfo{person}{Sunipa Dev}.} \bibinfo{year}{2024}\natexlab{}.
\newblock \showarticletitle{{S}ee{GULL} Multilingual: a Dataset of
  Geo-Culturally Situated Stereotypes}. In
  \bibinfo{booktitle}{\emph{Proceedings of the 62nd Annual Meeting of the
  Association for Computational Linguistics (Volume 2: Short Papers)}},
  \bibfield{editor}{\bibinfo{person}{Lun-Wei Ku}, \bibinfo{person}{Andre
  Martins}, {and} \bibinfo{person}{Vivek Srikumar}} (Eds.).
  \bibinfo{publisher}{Association for Computational Linguistics},
  \bibinfo{address}{Bangkok, Thailand}, \bibinfo{pages}{842--854}.
\newblock
\href{https://doi.org/10.18653/v1/2024.acl-short.75}{doi:\nolinkurl{10.18653/v1/2024.acl-short.75}}


\bibitem[Blodgett et~al\mbox{.}(2021)]%
        {blodgett-etal-2021-stereotyping}
\bibfield{author}{\bibinfo{person}{Su~Lin Blodgett}, \bibinfo{person}{Gilsinia
  Lopez}, \bibinfo{person}{Alexandra Olteanu}, \bibinfo{person}{Robert Sim},
  {and} \bibinfo{person}{Hanna Wallach}.} \bibinfo{year}{2021}\natexlab{}.
\newblock \showarticletitle{Stereotyping {N}orwegian Salmon: An Inventory of
  Pitfalls in Fairness Benchmark Datasets}. In
  \bibinfo{booktitle}{\emph{Proceedings of the 59th Annual Meeting of the
  Association for Computational Linguistics and the 11th International Joint
  Conference on Natural Language Processing (Volume 1: Long Papers)}},
  \bibfield{editor}{\bibinfo{person}{Chengqing Zong}, \bibinfo{person}{Fei
  Xia}, \bibinfo{person}{Wenjie Li}, {and} \bibinfo{person}{Roberto Navigli}}
  (Eds.). \bibinfo{publisher}{Association for Computational Linguistics},
  \bibinfo{address}{Online}, \bibinfo{pages}{1004--1015}.
\newblock
\href{https://doi.org/10.18653/v1/2021.acl-long.81}{doi:\nolinkurl{10.18653/v1/2021.acl-long.81}}


\bibitem[Buolamwini and Gebru(2018)]%
        {pmlr-v81-buolamwini18a}
\bibfield{author}{\bibinfo{person}{Joy Buolamwini} {and}
  \bibinfo{person}{Timnit Gebru}.} \bibinfo{year}{2018}\natexlab{}.
\newblock \showarticletitle{Gender Shades: Intersectional Accuracy Disparities
  in Commercial Gender Classification}. In
  \bibinfo{booktitle}{\emph{Proceedings of the 1st Conference on Fairness,
  Accountability and Transparency}} \emph{(\bibinfo{series}{Proceedings of
  Machine Learning Research}, Vol.~\bibinfo{volume}{81})},
  \bibfield{editor}{\bibinfo{person}{Sorelle~A. Friedler} {and}
  \bibinfo{person}{Christo Wilson}} (Eds.). \bibinfo{publisher}{PMLR},
  \bibinfo{pages}{77--91}.
\newblock
\urldef\tempurl%
\url{https://proceedings.mlr.press/v81/buolamwini18a.html}
\showURL{%
\tempurl}


\bibitem[Crenshaw(1989)]%
        {crenshaw1989demarginalizing}
\bibfield{author}{\bibinfo{person}{Kimberl{\'e} Crenshaw}.}
  \bibinfo{year}{1989}\natexlab{}.
\newblock \showarticletitle{Demarginalizing the Intersection of Race and Sex: A
  Black Feminist Critique of Antidiscrimination Doctrine, Feminist Theory and
  Antiracist Politics}.
\newblock \bibinfo{journal}{\emph{University of Chicago Legal Forum}}
  \bibinfo{volume}{1989}, \bibinfo{number}{1} (\bibinfo{year}{1989}),
  \bibinfo{pages}{139--167}.
\newblock
\urldef\tempurl%
\url{https://chicagounbound.uchicago.edu/uclf/vol1989/iss1/8/}
\showURL{%
\tempurl}


\bibitem[Cronbach and Gleser(1953)]%
        {cronbach1953assessing}
\bibfield{author}{\bibinfo{person}{Lee~J. Cronbach} {and}
  \bibinfo{person}{Goldine~C. Gleser}.} \bibinfo{year}{1953}\natexlab{}.
\newblock \showarticletitle{Assessing similarity between profiles}.
\newblock \bibinfo{journal}{\emph{Psychological Bulletin}}
  \bibinfo{volume}{50}, \bibinfo{number}{6} (\bibinfo{year}{1953}),
  \bibinfo{pages}{456--473}.
\newblock
\href{https://doi.org/10.1037/h0057173}{doi:\nolinkurl{10.1037/h0057173}}


\bibitem[Dammu et~al\mbox{.}(2024)]%
        {dammu-etal-2024-uncultured}
\bibfield{author}{\bibinfo{person}{Preetam Prabhu~Srikar Dammu},
  \bibinfo{person}{Hayoung Jung}, \bibinfo{person}{Anjali Singh},
  \bibinfo{person}{Monojit Choudhury}, {and} \bibinfo{person}{Tanu Mitra}.}
  \bibinfo{year}{2024}\natexlab{}.
\newblock \showarticletitle{``They are uncultured'': Unveiling Covert Harms and
  Social Threats in {LLM} Generated Conversations}. In
  \bibinfo{booktitle}{\emph{Proceedings of the 2024 Conference on Empirical
  Methods in Natural Language Processing}},
  \bibfield{editor}{\bibinfo{person}{Yaser Al-Onaizan}, \bibinfo{person}{Mohit
  Bansal}, {and} \bibinfo{person}{Yun-Nung Chen}} (Eds.).
  \bibinfo{publisher}{Association for Computational Linguistics},
  \bibinfo{address}{Miami, Florida, USA}, \bibinfo{pages}{20339--20369}.
\newblock
\href{https://doi.org/10.18653/v1/2024.emnlp-main.1134}{doi:\nolinkurl{10.18653/v1/2024.emnlp-main.1134}}


\bibitem[Dev et~al\mbox{.}(2023)]%
        {NEURIPS2023_0dc91de8}
\bibfield{author}{\bibinfo{person}{Sunipa Dev}, \bibinfo{person}{Jaya Goyal},
  \bibinfo{person}{Dinesh Tewari}, \bibinfo{person}{Shachi Dave}, {and}
  \bibinfo{person}{Vinodkumar Prabhakaran}.} \bibinfo{year}{2023}\natexlab{}.
\newblock \showarticletitle{Building Socio-culturally Inclusive Stereotype
  Resources with Community Engagement}. In \bibinfo{booktitle}{\emph{Advances
  in Neural Information Processing Systems}},
  \bibfield{editor}{\bibinfo{person}{A.~Oh}, \bibinfo{person}{T.~Naumann},
  \bibinfo{person}{A.~Globerson}, \bibinfo{person}{K.~Saenko},
  \bibinfo{person}{M.~Hardt}, {and} \bibinfo{person}{S.~Levine}} (Eds.),
  Vol.~\bibinfo{volume}{36}. \bibinfo{publisher}{Curran Associates, Inc.},
  \bibinfo{pages}{4365--4381}.
\newblock
\href{https://doi.org/10.52202/075280-0194}{doi:\nolinkurl{10.52202/075280-0194}}


\bibitem[Egami and Imai(2019)]%
        {egami2019causal}
\bibfield{author}{\bibinfo{person}{Naoki Egami} {and} \bibinfo{person}{Kosuke
  Imai}.} \bibinfo{year}{2019}\natexlab{}.
\newblock \showarticletitle{Causal Interaction in Factorial Experiments:
  Application to Conjoint Analysis}.
\newblock \bibinfo{journal}{\emph{J. Amer. Statist. Assoc.}}
  \bibinfo{volume}{114}, \bibinfo{number}{526} (\bibinfo{year}{2019}),
  \bibinfo{pages}{529--540}.
\newblock
\href{https://doi.org/10.1080/01621459.2018.1476246}{doi:\nolinkurl{10.1080/01621459.2018.1476246}}


\bibitem[Evans et~al\mbox{.}(2018)]%
        {evans2018multilevel}
\bibfield{author}{\bibinfo{person}{Clare~R. Evans}, \bibinfo{person}{David~R.
  Williams}, \bibinfo{person}{Jukka-Pekka Onnela}, {and} \bibinfo{person}{S.~V.
  Subramanian}.} \bibinfo{year}{2018}\natexlab{}.
\newblock \showarticletitle{A multilevel approach to modeling health
  inequalities at the intersection of multiple social identities}.
\newblock \bibinfo{journal}{\emph{Social Science \& Medicine}}
  \bibinfo{volume}{203} (\bibinfo{year}{2018}), \bibinfo{pages}{64--73}.
\newblock
\href{https://doi.org/10.1016/j.socscimed.2017.11.011}{doi:\nolinkurl{10.1016/j.socscimed.2017.11.011}}


\bibitem[Ferrara et~al\mbox{.}(2025)]%
        {ferrara2025sizeadaptive}
\bibfield{author}{\bibinfo{person}{Antonio Ferrara}, \bibinfo{person}{Francesco
  Cozzi}, \bibinfo{person}{Alan Perotti}, \bibinfo{person}{Andr{\'e} Panisson},
  {and} \bibinfo{person}{Francesco Bonchi}.} \bibinfo{year}{2025}\natexlab{}.
\newblock \showarticletitle{Size-adaptive Hypothesis Testing for Fairness}. In
  \bibinfo{booktitle}{\emph{Advances in Neural Information Processing Systems
  (NeurIPS)}}.
\newblock
\showeprint[arxiv]{2506.10586}


\bibitem[Franklin et~al\mbox{.}(2014)]%
        {franklin2014plasmode}
\bibfield{author}{\bibinfo{person}{Jessica~M. Franklin},
  \bibinfo{person}{Sebastian Schneeweiss}, \bibinfo{person}{Jennifer~M.
  Polinski}, {and} \bibinfo{person}{Jeremy~A. Rassen}.}
  \bibinfo{year}{2014}\natexlab{}.
\newblock \showarticletitle{Plasmode simulation for the evaluation of
  pharmacoepidemiologic methods in complex healthcare databases}.
\newblock \bibinfo{journal}{\emph{Computational Statistics \& Data Analysis}}
  \bibinfo{volume}{72} (\bibinfo{year}{2014}), \bibinfo{pages}{219--226}.
\newblock
\href{https://doi.org/10.1016/j.csda.2013.10.018}{doi:\nolinkurl{10.1016/j.csda.2013.10.018}}


\bibitem[Freedman and Lane(1983)]%
        {freedman1983nonstochastic}
\bibfield{author}{\bibinfo{person}{David Freedman} {and} \bibinfo{person}{David
  Lane}.} \bibinfo{year}{1983}\natexlab{}.
\newblock \showarticletitle{A Nonstochastic Interpretation of Reported
  Significance Levels}.
\newblock \bibinfo{journal}{\emph{Journal of Business \& Economic Statistics}}
  \bibinfo{volume}{1}, \bibinfo{number}{4} (\bibinfo{year}{1983}),
  \bibinfo{pages}{292--298}.
\newblock
\href{https://doi.org/10.1080/07350015.1983.10509354}{doi:\nolinkurl{10.1080/07350015.1983.10509354}}


\bibitem[Ghosh et~al\mbox{.}(2024)]%
        {3716662.3716702}
\bibfield{author}{\bibinfo{person}{Sourojit Ghosh},
  \bibinfo{person}{Pranav~Narayanan Venkit}, \bibinfo{person}{Sanjana Gautam},
  \bibinfo{person}{Shomir Wilson}, {and} \bibinfo{person}{Aylin Caliskan}.}
  \bibinfo{year}{2024}\natexlab{}.
\newblock \showarticletitle{Do Generative AI Models Output Harm While
  Representing Non-Western Cultures: Evidence from a Community-Centered
  Approach}. In \bibinfo{booktitle}{\emph{Proceedings of the 2024 AAAI/ACM
  Conference on AI, Ethics, and Society}} (San Jose, California, USA)
  \emph{(\bibinfo{series}{AIES '24}, Vol.~\bibinfo{volume}{7})}.
  \bibinfo{publisher}{AAAI Press}, \bibinfo{pages}{476–489}.
\newblock
\urldef\tempurl%
\url{https://ojs.aaai.org/index.php/AIES/article/view/31651}
\showURL{%
\tempurl}


\bibitem[Goldberg(1992)]%
        {goldberg1992development}
\bibfield{author}{\bibinfo{person}{Lewis~R. Goldberg}.}
  \bibinfo{year}{1992}\natexlab{}.
\newblock \showarticletitle{The development of markers for the {B}ig-{F}ive
  factor structure}.
\newblock \bibinfo{journal}{\emph{Psychological Assessment}}
  \bibinfo{volume}{4}, \bibinfo{number}{1} (\bibinfo{year}{1992}),
  \bibinfo{pages}{26--42}.
\newblock
\href{https://doi.org/10.1037/1040-3590.4.1.26}{doi:\nolinkurl{10.1037/1040-3590.4.1.26}}


\bibitem[{Government of India}(1950)]%
        {scorder1950}
\bibfield{author}{\bibinfo{person}{{Government of India}}.}
  \bibinfo{year}{1950}\natexlab{}.
\newblock \bibinfo{title}{The Constitution (Scheduled Castes) Order, 1950}.
\newblock \bibinfo{howpublished}{C.O. 19, as amended}.
\newblock
\urldef\tempurl%
\url{https://socialjustice.gov.in/writereaddata/UploadFile/scorder1950.pdf}
\showURL{%
\tempurl}


\bibitem[Guo and Caliskan(2021)]%
        {10.1145/3461702.3462536}
\bibfield{author}{\bibinfo{person}{Wei Guo} {and} \bibinfo{person}{Aylin
  Caliskan}.} \bibinfo{year}{2021}\natexlab{}.
\newblock \showarticletitle{Detecting Emergent Intersectional Biases:
  Contextualized Word Embeddings Contain a Distribution of Human-like Biases}.
  In \bibinfo{booktitle}{\emph{Proceedings of the 2021 AAAI/ACM Conference on
  AI, Ethics, and Society}} (Virtual Event, USA) \emph{(\bibinfo{series}{AIES
  '21})}. \bibinfo{publisher}{Association for Computing Machinery},
  \bibinfo{address}{New York, NY, USA}, \bibinfo{pages}{122–133}.
\newblock
\showISBNx{9781450384735}
\href{https://doi.org/10.1145/3461702.3462536}{doi:\nolinkurl{10.1145/3461702.3462536}}


\bibitem[Gupta et~al\mbox{.}(2026)]%
        {gupta2026compoundingdisadvantageauditingintersectional}
\bibfield{author}{\bibinfo{person}{Amogh Gupta}, \bibinfo{person}{Niharika
  Patil}, \bibinfo{person}{Sourojit Ghosh}, {and} \bibinfo{person}{Snehalkumar
  (Neil)~S. Gaikwad}.} \bibinfo{year}{2026}\natexlab{}.
\newblock \showarticletitle{Compounding Disadvantage: Auditing Intersectional
  Bias in LLM-Generated Explanations Across Indian and American STEM
  Education}. In \bibinfo{booktitle}{\emph{Proceedings of the 2026 ACM
  Conference on Fairness, Accountability, and Transparency}}
  \emph{(\bibinfo{series}{FAccT '26})}. \bibinfo{publisher}{Association for
  Computing Machinery}, \bibinfo{address}{New York, NY, USA},
  \bibinfo{pages}{6093--6155}.
\newblock
\href{https://doi.org/10.1145/3805689.3812394}{doi:\nolinkurl{10.1145/3805689.3812394}}


\bibitem[Hada et~al\mbox{.}(2024)]%
        {10.1145/3630106.3659017}
\bibfield{author}{\bibinfo{person}{Rishav Hada}, \bibinfo{person}{Safiya
  Husain}, \bibinfo{person}{Varun Gumma}, \bibinfo{person}{Harshita Diddee},
  \bibinfo{person}{Aditya Yadavalli}, \bibinfo{person}{Agrima Seth},
  \bibinfo{person}{Nidhi Kulkarni}, \bibinfo{person}{Ujwal Gadiraju},
  \bibinfo{person}{Aditya Vashistha}, \bibinfo{person}{Vivek Seshadri}, {and}
  \bibinfo{person}{Kalika Bali}.} \bibinfo{year}{2024}\natexlab{}.
\newblock \showarticletitle{Akal Badi ya Bias: An Exploratory Study of Gender
  Bias in Hindi Language Technology}. In \bibinfo{booktitle}{\emph{Proceedings
  of the 2024 ACM Conference on Fairness, Accountability, and Transparency}}
  (Rio de Janeiro, Brazil) \emph{(\bibinfo{series}{FAccT '24})}.
  \bibinfo{publisher}{Association for Computing Machinery},
  \bibinfo{address}{New York, NY, USA}, \bibinfo{pages}{1926–1939}.
\newblock
\showISBNx{9798400704505}
\href{https://doi.org/10.1145/3630106.3659017}{doi:\nolinkurl{10.1145/3630106.3659017}}


\bibitem[Haim et~al\mbox{.}(2024)]%
        {salinas2025whatsnameauditinglarge}
\bibfield{author}{\bibinfo{person}{Amit Haim}, \bibinfo{person}{Alejandro
  Salinas}, {and} \bibinfo{person}{Julian Nyarko}.}
  \bibinfo{year}{2024}\natexlab{}.
\newblock \bibinfo{title}{What's in a Name? Auditing Large Language Models for
  Race and Gender Bias}.
\newblock
\showeprint[arxiv]{2402.14875}
\urldef\tempurl%
\url{https://arxiv.org/abs/2402.14875}
\showURL{%
\tempurl}


\bibitem[Himmelreich et~al\mbox{.}(2025)]%
        {himmelreich2025intersectionality}
\bibfield{author}{\bibinfo{person}{Johannes Himmelreich},
  \bibinfo{person}{Arbie Hsu}, \bibinfo{person}{Ellen Veomett}, {and}
  \bibinfo{person}{Kristian Lum}.} \bibinfo{year}{2025}\natexlab{}.
\newblock \showarticletitle{The Intersectionality Problem for Algorithmic
  Fairness}. In \bibinfo{booktitle}{\emph{Proceedings of the Algorithmic
  Fairness Through the Lens of Metrics and Evaluation}}
  \emph{(\bibinfo{series}{Proceedings of Machine Learning Research},
  Vol.~\bibinfo{volume}{279})}. \bibinfo{publisher}{PMLR},
  \bibinfo{pages}{68--95}.
\newblock
\showeprint[arxiv]{2411.02569}
\urldef\tempurl%
\url{https://proceedings.mlr.press/v279/himmelreich25a.html}
\showURL{%
\tempurl}


\bibitem[Jha et~al\mbox{.}(2023)]%
        {jha-etal-2023-seegull}
\bibfield{author}{\bibinfo{person}{Akshita Jha}, \bibinfo{person}{Aida Davani},
  \bibinfo{person}{Chandan~K. Reddy}, \bibinfo{person}{Shachi Dave},
  \bibinfo{person}{Vinodkumar Prabhakaran}, {and} \bibinfo{person}{Sunipa
  Dev}.} \bibinfo{year}{2023}\natexlab{}.
\newblock \showarticletitle{{S}ee{GULL}: A Stereotype Benchmark with Broad
  Geo-Cultural Coverage Leveraging Generative Models}. In
  \bibinfo{booktitle}{\emph{Proceedings of the 61st Annual Meeting of the
  Association for Computational Linguistics (Volume 1: Long Papers)}},
  \bibfield{editor}{\bibinfo{person}{Anna Rogers}, \bibinfo{person}{Jordan
  Boyd-Graber}, {and} \bibinfo{person}{Naoaki Okazaki}} (Eds.).
  \bibinfo{publisher}{Association for Computational Linguistics},
  \bibinfo{address}{Toronto, Canada}, \bibinfo{pages}{9851--9870}.
\newblock
\href{https://doi.org/10.18653/v1/2023.acl-long.548}{doi:\nolinkurl{10.18653/v1/2023.acl-long.548}}


\bibitem[Joshi et~al\mbox{.}(2025)]%
        {10.1145/3715275.3732208}
\bibfield{author}{\bibinfo{person}{Ishika Joshi}, \bibinfo{person}{Ishita
  Gupta}, \bibinfo{person}{Adrita Dey}, {and} \bibinfo{person}{Tapan Parikh}.}
  \bibinfo{year}{2025}\natexlab{}.
\newblock \showarticletitle{“Since Lawyers are Males..”: Examining Implicit
  Gender Bias in Hindi Language Generation by LLMs}. In
  \bibinfo{booktitle}{\emph{Proceedings of the 2025 ACM Conference on Fairness,
  Accountability, and Transparency}} (Athens, Greece)
  \emph{(\bibinfo{series}{FAccT '25})}. \bibinfo{publisher}{Association for
  Computing Machinery}, \bibinfo{address}{New York, NY, USA},
  \bibinfo{pages}{3254–3264}.
\newblock
\showISBNx{9798400714825}
\href{https://doi.org/10.1145/3715275.3732208}{doi:\nolinkurl{10.1145/3715275.3732208}}


\bibitem[Kanubala and Valera(2025)]%
        {kanubala2025misalignment}
\bibfield{author}{\bibinfo{person}{Deborah~Dormah Kanubala} {and}
  \bibinfo{person}{Isabel Valera}.} \bibinfo{year}{2025}\natexlab{}.
\newblock \showarticletitle{On the Misalignment Between Legal Notions and
  Statistical Metrics of Intersectional Fairness}. In
  \bibinfo{booktitle}{\emph{Proceedings of the AAAI/ACM Conference on AI,
  Ethics, and Society}}, Vol.~\bibinfo{volume}{8}. \bibinfo{pages}{1363--1374}.
\newblock
\href{https://doi.org/10.1609/aies.v8i2.36637}{doi:\nolinkurl{10.1609/aies.v8i2.36637}}


\bibitem[Kearns et~al\mbox{.}(2018)]%
        {pmlr-v80-kearns18a}
\bibfield{author}{\bibinfo{person}{Michael Kearns}, \bibinfo{person}{Seth
  Neel}, \bibinfo{person}{Aaron Roth}, {and} \bibinfo{person}{Zhiwei~Steven
  Wu}.} \bibinfo{year}{2018}\natexlab{}.
\newblock \showarticletitle{Preventing Fairness Gerrymandering: Auditing and
  Learning for Subgroup Fairness}. In \bibinfo{booktitle}{\emph{Proceedings of
  the 35th International Conference on Machine Learning}}
  \emph{(\bibinfo{series}{Proceedings of Machine Learning Research},
  Vol.~\bibinfo{volume}{80})}, \bibfield{editor}{\bibinfo{person}{Jennifer Dy}
  {and} \bibinfo{person}{Andreas Krause}} (Eds.). \bibinfo{publisher}{PMLR},
  \bibinfo{pages}{2564--2572}.
\newblock
\urldef\tempurl%
\url{https://proceedings.mlr.press/v80/kearns18a.html}
\showURL{%
\tempurl}


\bibitem[Khandelwal et~al\mbox{.}(2024)]%
        {10.1145/3677525.3678666}
\bibfield{author}{\bibinfo{person}{Khyati Khandelwal}, \bibinfo{person}{Manuel
  Tonneau}, \bibinfo{person}{Andrew~M. Bean}, \bibinfo{person}{Hannah~Rose
  Kirk}, {and} \bibinfo{person}{Scott~A. Hale}.}
  \bibinfo{year}{2024}\natexlab{}.
\newblock \showarticletitle{Indian-BhED: A Dataset for Measuring India-Centric
  Biases in Large Language Models}. In \bibinfo{booktitle}{\emph{Proceedings of
  the 2024 International Conference on Information Technology for Social Good}}
  (Bremen, Germany) \emph{(\bibinfo{series}{GoodIT '24})}.
  \bibinfo{publisher}{Association for Computing Machinery},
  \bibinfo{address}{New York, NY, USA}, \bibinfo{pages}{231–239}.
\newblock
\showISBNx{9798400710940}
\href{https://doi.org/10.1145/3677525.3678666}{doi:\nolinkurl{10.1145/3677525.3678666}}


\bibitem[Kirk et~al\mbox{.}(2021)]%
        {3540261.3540461}
\bibfield{author}{\bibinfo{person}{Hannah~Rose Kirk}, \bibinfo{person}{Yennie
  Jun}, \bibinfo{person}{Haider Iqbal}, \bibinfo{person}{Elias Benussi},
  \bibinfo{person}{Filippo Volpin}, \bibinfo{person}{Frederic~A. Dreyer},
  \bibinfo{person}{Aleksandar Shtedritski}, {and} \bibinfo{person}{Yuki~M.
  Asano}.} \bibinfo{year}{2021}\natexlab{}.
\newblock \showarticletitle{Bias out-of-the-box: an empirical analysis of
  intersectional occupational biases in popular generative language models}. In
  \bibinfo{booktitle}{\emph{Proceedings of the 35th International Conference on
  Neural Information Processing Systems}} \emph{(\bibinfo{series}{NIPS '21})}.
  \bibinfo{publisher}{Curran Associates Inc.}, \bibinfo{address}{Red Hook, NY,
  USA}, Article \bibinfo{articleno}{200}, \bibinfo{numpages}{14}~pages.
\newblock
\showISBNx{9781713845393}


\bibitem[Kirtane and Anand(2022)]%
        {kirtane-anand-2022-mitigating}
\bibfield{author}{\bibinfo{person}{Neeraja Kirtane} {and}
  \bibinfo{person}{Tanvi Anand}.} \bibinfo{year}{2022}\natexlab{}.
\newblock \showarticletitle{Mitigating Gender Stereotypes in {H}indi and
  {M}arathi}. In \bibinfo{booktitle}{\emph{Proceedings of the 4th Workshop on
  Gender Bias in Natural Language Processing (GeBNLP)}},
  \bibfield{editor}{\bibinfo{person}{Christian Hardmeier},
  \bibinfo{person}{Christine Basta}, \bibinfo{person}{Marta~R.
  Costa-juss{\`a}}, \bibinfo{person}{Gabriel Stanovsky}, {and}
  \bibinfo{person}{Hila Gonen}} (Eds.). \bibinfo{publisher}{Association for
  Computational Linguistics}, \bibinfo{address}{Seattle, Washington},
  \bibinfo{pages}{145--150}.
\newblock
\href{https://doi.org/10.18653/v1/2022.gebnlp-1.16}{doi:\nolinkurl{10.18653/v1/2022.gebnlp-1.16}}


\bibitem[Kroenke et~al\mbox{.}(2009)]%
        {kroenke2009phq8}
\bibfield{author}{\bibinfo{person}{Kurt Kroenke}, \bibinfo{person}{Tara~W.
  Strine}, \bibinfo{person}{Robert~L. Spitzer}, \bibinfo{person}{Janet B.~W.
  Williams}, \bibinfo{person}{Joyce~T. Berry}, {and} \bibinfo{person}{Ali~H.
  Mokdad}.} \bibinfo{year}{2009}\natexlab{}.
\newblock \showarticletitle{The {PHQ-8} as a measure of current depression in
  the general population}.
\newblock \bibinfo{journal}{\emph{Journal of Affective Disorders}}
  \bibinfo{volume}{114}, \bibinfo{number}{1--3} (\bibinfo{year}{2009}),
  \bibinfo{pages}{163--173}.
\newblock
\href{https://doi.org/10.1016/j.jad.2008.06.026}{doi:\nolinkurl{10.1016/j.jad.2008.06.026}}


\bibitem[Lalor et~al\mbox{.}(2022)]%
        {lalor-etal-2022-benchmarking}
\bibfield{author}{\bibinfo{person}{John Lalor}, \bibinfo{person}{Yi Yang},
  \bibinfo{person}{Kendall Smith}, \bibinfo{person}{Nicole Forsgren}, {and}
  \bibinfo{person}{Ahmed Abbasi}.} \bibinfo{year}{2022}\natexlab{}.
\newblock \showarticletitle{Benchmarking Intersectional Biases in {NLP}}. In
  \bibinfo{booktitle}{\emph{Proceedings of the 2022 Conference of the North
  American Chapter of the Association for Computational Linguistics: Human
  Language Technologies}}, \bibfield{editor}{\bibinfo{person}{Marine Carpuat},
  \bibinfo{person}{Marie-Catherine de~Marneffe}, {and}
  \bibinfo{person}{Ivan~Vladimir Meza~Ruiz}} (Eds.).
  \bibinfo{publisher}{Association for Computational Linguistics},
  \bibinfo{address}{Seattle, United States}, \bibinfo{pages}{3598--3609}.
\newblock
\href{https://doi.org/10.18653/v1/2022.naacl-main.263}{doi:\nolinkurl{10.18653/v1/2022.naacl-main.263}}


\bibitem[Ledoit and Wolf(2004)]%
        {ledoit2004wellconditioned}
\bibfield{author}{\bibinfo{person}{Olivier Ledoit} {and}
  \bibinfo{person}{Michael Wolf}.} \bibinfo{year}{2004}\natexlab{}.
\newblock \showarticletitle{A well-conditioned estimator for large-dimensional
  covariance matrices}.
\newblock \bibinfo{journal}{\emph{Journal of Multivariate Analysis}}
  \bibinfo{volume}{88}, \bibinfo{number}{2} (\bibinfo{year}{2004}),
  \bibinfo{pages}{365--411}.
\newblock
\href{https://doi.org/10.1016/S0047-259X(03)00096-4}{doi:\nolinkurl{10.1016/S0047-259X(03)00096-4}}


\bibitem[Liddell and Kruschke(2018)]%
        {liddell2018analyzing}
\bibfield{author}{\bibinfo{person}{Torrin~M. Liddell} {and}
  \bibinfo{person}{John~K. Kruschke}.} \bibinfo{year}{2018}\natexlab{}.
\newblock \showarticletitle{Analyzing ordinal data with metric models: What
  could possibly go wrong?}
\newblock \bibinfo{journal}{\emph{Journal of Experimental Social Psychology}}
  \bibinfo{volume}{79} (\bibinfo{year}{2018}), \bibinfo{pages}{328--348}.
\newblock
\href{https://doi.org/10.1016/j.jesp.2018.08.009}{doi:\nolinkurl{10.1016/j.jesp.2018.08.009}}


\bibitem[Loftus(1978)]%
        {loftus1978interpretation}
\bibfield{author}{\bibinfo{person}{Geoffrey~R. Loftus}.}
  \bibinfo{year}{1978}\natexlab{}.
\newblock \showarticletitle{On Interpretation of Interactions}.
\newblock \bibinfo{journal}{\emph{Memory \& Cognition}} \bibinfo{volume}{6},
  \bibinfo{number}{3} (\bibinfo{year}{1978}), \bibinfo{pages}{312--319}.
\newblock
\href{https://doi.org/10.3758/BF03197461}{doi:\nolinkurl{10.3758/BF03197461}}


\bibitem[Ma et~al\mbox{.}(2023)]%
        {ma-etal-2023-intersectional}
\bibfield{author}{\bibinfo{person}{Weicheng Ma}, \bibinfo{person}{Brian
  Chiang}, \bibinfo{person}{Tong Wu}, \bibinfo{person}{Lili Wang}, {and}
  \bibinfo{person}{Soroush Vosoughi}.} \bibinfo{year}{2023}\natexlab{}.
\newblock \showarticletitle{Intersectional Stereotypes in Large Language
  Models: Dataset and Analysis}. In \bibinfo{booktitle}{\emph{Findings of the
  Association for Computational Linguistics: EMNLP 2023}},
  \bibfield{editor}{\bibinfo{person}{Houda Bouamor}, \bibinfo{person}{Juan
  Pino}, {and} \bibinfo{person}{Kalika Bali}} (Eds.).
  \bibinfo{publisher}{Association for Computational Linguistics},
  \bibinfo{address}{Singapore}, \bibinfo{pages}{8589--8597}.
\newblock
\href{https://doi.org/10.18653/v1/2023.findings-emnlp.575}{doi:\nolinkurl{10.18653/v1/2023.findings-emnlp.575}}


\bibitem[MacKinnon and White(1985)]%
        {mackinnon1985some}
\bibfield{author}{\bibinfo{person}{James~G. MacKinnon} {and}
  \bibinfo{person}{Halbert White}.} \bibinfo{year}{1985}\natexlab{}.
\newblock \showarticletitle{Some Heteroskedasticity-Consistent Covariance
  Matrix Estimators with Improved Finite Sample Properties}.
\newblock \bibinfo{journal}{\emph{Journal of Econometrics}}
  \bibinfo{volume}{29}, \bibinfo{number}{3} (\bibinfo{year}{1985}),
  \bibinfo{pages}{305--325}.
\newblock


\bibitem[Malik et~al\mbox{.}(2022)]%
        {malik-etal-2022-socially}
\bibfield{author}{\bibinfo{person}{Vijit Malik}, \bibinfo{person}{Sunipa Dev},
  \bibinfo{person}{Akihiro Nishi}, \bibinfo{person}{Nanyun Peng}, {and}
  \bibinfo{person}{Kai-Wei Chang}.} \bibinfo{year}{2022}\natexlab{}.
\newblock \showarticletitle{Socially Aware Bias Measurements for {H}indi
  Language Representations}. In \bibinfo{booktitle}{\emph{Proceedings of the
  2022 Conference of the North American Chapter of the Association for
  Computational Linguistics: Human Language Technologies}},
  \bibfield{editor}{\bibinfo{person}{Marine Carpuat},
  \bibinfo{person}{Marie-Catherine de~Marneffe}, {and}
  \bibinfo{person}{Ivan~Vladimir Meza~Ruiz}} (Eds.).
  \bibinfo{publisher}{Association for Computational Linguistics},
  \bibinfo{address}{Seattle, United States}, \bibinfo{pages}{1041--1052}.
\newblock
\href{https://doi.org/10.18653/v1/2022.naacl-main.76}{doi:\nolinkurl{10.18653/v1/2022.naacl-main.76}}


\bibitem[Morris et~al\mbox{.}(2019)]%
        {morris2019using}
\bibfield{author}{\bibinfo{person}{Tim~P. Morris}, \bibinfo{person}{Ian~R.
  White}, {and} \bibinfo{person}{Michael~J. Crowther}.}
  \bibinfo{year}{2019}\natexlab{}.
\newblock \showarticletitle{Using simulation studies to evaluate statistical
  methods}.
\newblock \bibinfo{journal}{\emph{Statistics in Medicine}}
  \bibinfo{volume}{38}, \bibinfo{number}{11} (\bibinfo{year}{2019}),
  \bibinfo{pages}{2074--2102}.
\newblock
\href{https://doi.org/10.1002/sim.8086}{doi:\nolinkurl{10.1002/sim.8086}}


\bibitem[Mukherjee et~al\mbox{.}(2023)]%
        {mukherjee-etal-2023-global}
\bibfield{author}{\bibinfo{person}{Anjishnu Mukherjee}, \bibinfo{person}{Chahat
  Raj}, \bibinfo{person}{Ziwei Zhu}, {and} \bibinfo{person}{Antonios
  Anastasopoulos}.} \bibinfo{year}{2023}\natexlab{}.
\newblock \showarticletitle{{G}lobal {V}oices, Local Biases: Socio-Cultural
  Prejudices across Languages}. In \bibinfo{booktitle}{\emph{Proceedings of the
  2023 Conference on Empirical Methods in Natural Language Processing}},
  \bibfield{editor}{\bibinfo{person}{Houda Bouamor}, \bibinfo{person}{Juan
  Pino}, {and} \bibinfo{person}{Kalika Bali}} (Eds.).
  \bibinfo{publisher}{Association for Computational Linguistics},
  \bibinfo{address}{Singapore}, \bibinfo{pages}{15828--15845}.
\newblock
\href{https://doi.org/10.18653/v1/2023.emnlp-main.981}{doi:\nolinkurl{10.18653/v1/2023.emnlp-main.981}}


\bibitem[Naik et~al\mbox{.}(2026)]%
        {naik2026simaaiuntycasteaudit}
\bibfield{author}{\bibinfo{person}{Atharva Naik}, \bibinfo{person}{Shounok
  Kar}, \bibinfo{person}{Varnika Sharma}, \bibinfo{person}{Ashwin
  Rajadesingan}, {and} \bibinfo{person}{Koustuv Saha}.}
  \bibinfo{year}{2026}\natexlab{}.
\newblock \bibinfo{title}{Sima AIunty: Caste Audit in LLM-Driven Matchmaking}.
\newblock
\showeprint[arxiv]{2603.29288}~[cs.CY]
\urldef\tempurl%
\url{https://arxiv.org/abs/2603.29288}
\showURL{%
\tempurl}


\bibitem[Nair et~al\mbox{.}(2026)]%
        {nair2026rupeebiasauditingdemographicbias}
\bibfield{author}{\bibinfo{person}{Pavithra P~M Nair}, \bibinfo{person}{Bhavik
  Talaviya}, \bibinfo{person}{Shourya Bhushan}, \bibinfo{person}{Rahul
  Pankajakshan}, \bibinfo{person}{Seema Guruvadoo}, \bibinfo{person}{Avinash
  Agarwal}, \bibinfo{person}{Gilad Gressel}, {and}
  \bibinfo{person}{Krishnashree Achuthan}.} \bibinfo{year}{2026}\natexlab{}.
\newblock \bibinfo{title}{RupeeBias: Auditing Demographic Bias in Indian
  Economic Guidance from Large Language Models}.
\newblock
\showeprint[arxiv]{2609.31245}~[cs.CL]
\urldef\tempurl%
\url{https://arxiv.org/abs/2609.31245}
\showURL{%
\tempurl}


\bibitem[Nawale et~al\mbox{.}(2025)]%
        {nawale-etal-2025-fairi}
\bibfield{author}{\bibinfo{person}{Janki~Atul Nawale},
  \bibinfo{person}{Mohammed Safi Ur~Rahman Khan}, \bibinfo{person}{Janani D},
  \bibinfo{person}{Mansi Gupta}, \bibinfo{person}{Danish Pruthi}, {and}
  \bibinfo{person}{Mitesh~M Khapra}.} \bibinfo{year}{2025}\natexlab{}.
\newblock \showarticletitle{{F}air{I} Tales: Evaluation of Fairness in {I}ndian
  Contexts with a Focus on Bias and Stereotypes}. In
  \bibinfo{booktitle}{\emph{Proceedings of the 63rd Annual Meeting of the
  Association for Computational Linguistics (Volume 1: Long Papers)}},
  \bibfield{editor}{\bibinfo{person}{Wanxiang Che}, \bibinfo{person}{Joyce
  Nabende}, \bibinfo{person}{Ekaterina Shutova}, {and}
  \bibinfo{person}{Mohammad~Taher Pilehvar}} (Eds.).
  \bibinfo{publisher}{Association for Computational Linguistics},
  \bibinfo{address}{Vienna, Austria}, \bibinfo{pages}{30331--30380}.
\newblock
\showISBNx{979-8-89176-251-0}
\href{https://doi.org/10.18653/v1/2025.acl-long.1465}{doi:\nolinkurl{10.18653/v1/2025.acl-long.1465}}


\bibitem[Parrish et~al\mbox{.}(2022)]%
        {parrish-etal-2022-bbq}
\bibfield{author}{\bibinfo{person}{Alicia Parrish}, \bibinfo{person}{Angelica
  Chen}, \bibinfo{person}{Nikita Nangia}, \bibinfo{person}{Vishakh Padmakumar},
  \bibinfo{person}{Jason Phang}, \bibinfo{person}{Jana Thompson},
  \bibinfo{person}{Phu~Mon Htut}, {and} \bibinfo{person}{Samuel~R. Bowman}.}
  \bibinfo{year}{2022}\natexlab{}.
\newblock \showarticletitle{{BBQ}: A hand-built bias benchmark for question
  answering}. In \bibinfo{booktitle}{\emph{Findings of the Association for
  Computational Linguistics: ACL 2022}},
  \bibfield{editor}{\bibinfo{person}{Smaranda Muresan},
  \bibinfo{person}{Preslav Nakov}, {and} \bibinfo{person}{Aline Villavicencio}}
  (Eds.). \bibinfo{publisher}{Association for Computational Linguistics},
  \bibinfo{address}{Dublin, Ireland}, \bibinfo{pages}{2086--2105}.
\newblock
\href{https://doi.org/10.18653/v1/2022.findings-acl.165}{doi:\nolinkurl{10.18653/v1/2022.findings-acl.165}}


\bibitem[Phipson and Smyth(2010)]%
        {phipson2010permutation}
\bibfield{author}{\bibinfo{person}{Belinda Phipson} {and}
  \bibinfo{person}{Gordon~K. Smyth}.} \bibinfo{year}{2010}\natexlab{}.
\newblock \showarticletitle{Permutation {P}-values Should Never Be Zero:
  Calculating Exact {P}-values When Permutations Are Randomly Drawn}.
\newblock \bibinfo{journal}{\emph{Statistical Applications in Genetics and
  Molecular Biology}} \bibinfo{volume}{9}, \bibinfo{number}{1}
  (\bibinfo{year}{2010}).
\newblock
\href{https://doi.org/10.2202/1544-6115.1585}{doi:\nolinkurl{10.2202/1544-6115.1585}}
\newblock
\shownote{Article 39}.


\bibitem[Phutane et~al\mbox{.}(2026)]%
        {phutane2026ableist}
\bibfield{author}{\bibinfo{person}{Mahika Phutane}, \bibinfo{person}{Hayoung
  Jung}, \bibinfo{person}{Matthew Kim}, \bibinfo{person}{Tanushree Mitra},
  {and} \bibinfo{person}{Aditya Vashistha}.} \bibinfo{year}{2026}\natexlab{}.
\newblock \showarticletitle{{ABLEIST}: Intersectional Disability Bias in
  {LLM}-Generated Hiring Scenarios}.
\newblock \bibinfo{journal}{\emph{Proceedings of the AAAI/ACM Conference on AI,
  Ethics, and Society}} \bibinfo{volume}{9}, \bibinfo{number}{3}
  (\bibinfo{year}{2026}), \bibinfo{pages}{2089--2104}.
\newblock
\showeprint[arxiv]{2510.10998}
\href{https://doi.org/10.1609/aies.v9i3.43291}{doi:\nolinkurl{10.1609/aies.v9i3.43291}}


\bibitem[{Prime Minister's High Level Committee}(2006)]%
        {sachar2006}
\bibfield{author}{\bibinfo{person}{{Prime Minister's High Level Committee}}.}
  \bibinfo{year}{2006}\natexlab{}.
\newblock \bibinfo{booktitle}{\emph{Social, Economic and Educational Status of
  the Muslim Community of India: A Report}}.
\newblock \bibinfo{type}{{T}echnical {R}eport}. \bibinfo{institution}{Cabinet
  Secretariat, Government of India}, \bibinfo{address}{New Delhi}.
\newblock


\bibitem[Qadri et~al\mbox{.}(2023)]%
        {10.1145/3593013.3594016}
\bibfield{author}{\bibinfo{person}{Rida Qadri}, \bibinfo{person}{Renee Shelby},
  \bibinfo{person}{Cynthia~L. Bennett}, {and} \bibinfo{person}{Remi Denton}.}
  \bibinfo{year}{2023}\natexlab{}.
\newblock \showarticletitle{AI’s Regimes of Representation: A
  Community-centered Study of Text-to-Image Models in South Asia}. In
  \bibinfo{booktitle}{\emph{Proceedings of the 2023 ACM Conference on Fairness,
  Accountability, and Transparency}} (Chicago, IL, USA)
  \emph{(\bibinfo{series}{FAccT '23})}. \bibinfo{publisher}{Association for
  Computing Machinery}, \bibinfo{address}{New York, NY, USA},
  \bibinfo{pages}{506–517}.
\newblock
\showISBNx{9798400701924}
\href{https://doi.org/10.1145/3593013.3594016}{doi:\nolinkurl{10.1145/3593013.3594016}}


\bibitem[Ramesh et~al\mbox{.}(2023)]%
        {ramesh-etal-2023-fairness}
\bibfield{author}{\bibinfo{person}{Krithika Ramesh}, \bibinfo{person}{Sunayana
  Sitaram}, {and} \bibinfo{person}{Monojit Choudhury}.}
  \bibinfo{year}{2023}\natexlab{}.
\newblock \showarticletitle{Fairness in Language Models Beyond {E}nglish: Gaps
  and Challenges}. In \bibinfo{booktitle}{\emph{Findings of the Association for
  Computational Linguistics: EACL 2023}},
  \bibfield{editor}{\bibinfo{person}{Andreas Vlachos} {and}
  \bibinfo{person}{Isabelle Augenstein}} (Eds.).
  \bibinfo{publisher}{Association for Computational Linguistics},
  \bibinfo{address}{Dubrovnik, Croatia}, \bibinfo{pages}{2106--2119}.
\newblock
\href{https://doi.org/10.18653/v1/2023.findings-eacl.157}{doi:\nolinkurl{10.18653/v1/2023.findings-eacl.157}}


\bibitem[Rao et~al\mbox{.}(2025)]%
        {rao2025invisiblefiltersculturalbias}
\bibfield{author}{\bibinfo{person}{Pooja S.~B. Rao},
  \bibinfo{person}{Laxminarayen~Nagarajan Venkatesan}, \bibinfo{person}{Mauro
  Cherubini}, {and} \bibinfo{person}{Dinesh~Babu Jayagopi}.}
  \bibinfo{year}{2025}\natexlab{}.
\newblock \bibinfo{title}{Invisible Filters: Cultural Bias in Hiring
  Evaluations Using Large Language Models}.
\newblock
\showeprint[arxiv]{2508.16673}~[cs.CY]
\urldef\tempurl%
\url{https://arxiv.org/abs/2508.16673}
\showURL{%
\tempurl}


\bibitem[Rennard and Xypolopoulos(2026)]%
        {rennard2026simulate}
\bibfield{author}{\bibinfo{person}{Virgile Rennard} {and}
  \bibinfo{person}{Christos Xypolopoulos}.} \bibinfo{year}{2026}\natexlab{}.
\newblock \bibinfo{title}{Large Language Models Simulate Intersectional
  Synthetic Identities with a Budget of One to Two Dimensions}.
\newblock
\showeprint[arxiv]{2608.23005}
\urldef\tempurl%
\url{https://arxiv.org/abs/2608.23005}
\showURL{%
\tempurl}


\bibitem[Sahoo et~al\mbox{.}(2024)]%
        {sahoo-etal-2024-indibias}
\bibfield{author}{\bibinfo{person}{Nihar Sahoo}, \bibinfo{person}{Pranamya
  Kulkarni}, \bibinfo{person}{Arif Ahmad}, \bibinfo{person}{Tanu Goyal},
  \bibinfo{person}{Narjis Asad}, \bibinfo{person}{Aparna Garimella}, {and}
  \bibinfo{person}{Pushpak Bhattacharyya}.} \bibinfo{year}{2024}\natexlab{}.
\newblock \showarticletitle{{I}ndi{B}ias: A Benchmark Dataset to Measure Social
  Biases in Language Models for {I}ndian Context}. In
  \bibinfo{booktitle}{\emph{Proceedings of the 2024 Conference of the North
  American Chapter of the Association for Computational Linguistics: Human
  Language Technologies (Volume 1: Long Papers)}},
  \bibfield{editor}{\bibinfo{person}{Kevin Duh}, \bibinfo{person}{Helena
  Gomez}, {and} \bibinfo{person}{Steven Bethard}} (Eds.).
  \bibinfo{publisher}{Association for Computational Linguistics},
  \bibinfo{address}{Mexico City, Mexico}, \bibinfo{pages}{8786--8806}.
\newblock
\href{https://doi.org/10.18653/v1/2024.naacl-long.487}{doi:\nolinkurl{10.18653/v1/2024.naacl-long.487}}


\bibitem[Sambasivan et~al\mbox{.}(2021)]%
        {10.1145/3442188.3445896}
\bibfield{author}{\bibinfo{person}{Nithya Sambasivan}, \bibinfo{person}{Erin
  Arnesen}, \bibinfo{person}{Ben Hutchinson}, \bibinfo{person}{Tulsee Doshi},
  {and} \bibinfo{person}{Vinodkumar Prabhakaran}.}
  \bibinfo{year}{2021}\natexlab{}.
\newblock \showarticletitle{Re-imagining Algorithmic Fairness in India and
  Beyond}. In \bibinfo{booktitle}{\emph{Proceedings of the 2021 ACM Conference
  on Fairness, Accountability, and Transparency}} (Virtual Event, Canada)
  \emph{(\bibinfo{series}{FAccT '21})}. \bibinfo{publisher}{Association for
  Computing Machinery}, \bibinfo{address}{New York, NY, USA},
  \bibinfo{pages}{315–328}.
\newblock
\showISBNx{9781450383097}
\href{https://doi.org/10.1145/3442188.3445896}{doi:\nolinkurl{10.1145/3442188.3445896}}


\bibitem[Seth et~al\mbox{.}(2025)]%
        {seth2025deep}
\bibfield{author}{\bibinfo{person}{Agrima Seth}, \bibinfo{person}{Monojit
  Choudhury}, \bibinfo{person}{Sunayana Sitaram}, \bibinfo{person}{Kentaro
  Toyama}, \bibinfo{person}{Aditya Vashistha}, {and} \bibinfo{person}{Kalika
  Bali}.} \bibinfo{year}{2025}\natexlab{}.
\newblock \showarticletitle{How Deep Is Representational Bias in LLMs? The
  Cases of Caste and Religion}. In \bibinfo{booktitle}{\emph{Proceedings of the
  AAAI/ACM Conference on AI, Ethics, and Society}}, Vol.~\bibinfo{volume}{8}.
  \bibinfo{pages}{2319--2330}.
\newblock
\href{https://doi.org/10.1609/aies.v8i3.36718}{doi:\nolinkurl{10.1609/aies.v8i3.36718}}


\bibitem[Souani et~al\mbox{.}(2025)]%
        {souani2025hinter}
\bibfield{author}{\bibinfo{person}{Badr Souani}, \bibinfo{person}{Ezekiel
  Soremekun}, \bibinfo{person}{Mike Papadakis}, \bibinfo{person}{Setsuko
  Yokoyama}, \bibinfo{person}{Sudipta Chattopadhyay}, {and}
  \bibinfo{person}{Yves Le~Traon}.} \bibinfo{year}{2025}\natexlab{}.
\newblock \bibinfo{title}{{HInter}: Exposing Hidden Intersectional Bias in
  Large Language Models}.
\newblock
\showeprint[arxiv]{2503.11962}~[cs.CL]


\bibitem[Spitzer et~al\mbox{.}(2006)]%
        {spitzer2006brief}
\bibfield{author}{\bibinfo{person}{Robert~L. Spitzer}, \bibinfo{person}{Kurt
  Kroenke}, \bibinfo{person}{Janet B.~W. Williams}, {and}
  \bibinfo{person}{Bernd L{\"o}we}.} \bibinfo{year}{2006}\natexlab{}.
\newblock \showarticletitle{A brief measure for assessing generalized anxiety
  disorder: The {GAD-7}}.
\newblock \bibinfo{journal}{\emph{Archives of Internal Medicine}}
  \bibinfo{volume}{166}, \bibinfo{number}{10} (\bibinfo{year}{2006}),
  \bibinfo{pages}{1092--1097}.
\newblock
\href{https://doi.org/10.1001/archinte.166.10.1092}{doi:\nolinkurl{10.1001/archinte.166.10.1092}}


\bibitem[Tamkin et~al\mbox{.}(2023)]%
        {tamkin2023evaluatingmitigatingdiscriminationlanguage}
\bibfield{author}{\bibinfo{person}{Alex Tamkin}, \bibinfo{person}{Amanda
  Askell}, \bibinfo{person}{Liane Lovitt}, \bibinfo{person}{Esin Durmus},
  \bibinfo{person}{Nicholas Joseph}, \bibinfo{person}{Shauna Kravec},
  \bibinfo{person}{Karina Nguyen}, \bibinfo{person}{Jared Kaplan}, {and}
  \bibinfo{person}{Deep Ganguli}.} \bibinfo{year}{2023}\natexlab{}.
\newblock \bibinfo{title}{Evaluating and Mitigating Discrimination in Language
  Model Decisions}.
\newblock
\showeprint[arxiv]{2312.03689}~[cs.CL]
\urldef\tempurl%
\url{https://arxiv.org/abs/2312.03689}
\showURL{%
\tempurl}


\bibitem[Tan and Celis(2019)]%
        {tan2019assessing}
\bibfield{author}{\bibinfo{person}{Yi~Chern Tan} {and}
  \bibinfo{person}{L.~Elisa Celis}.} \bibinfo{year}{2019}\natexlab{}.
\newblock \showarticletitle{Assessing Social and Intersectional Biases in
  Contextualized Word Representations}. In \bibinfo{booktitle}{\emph{Advances
  in Neural Information Processing Systems}},
  \bibfield{editor}{\bibinfo{person}{H.~Wallach},
  \bibinfo{person}{H.~Larochelle}, \bibinfo{person}{A.~Beygelzimer},
  \bibinfo{person}{F.~d'Alch\'{e} Buc}, \bibinfo{person}{E.~Fox}, {and}
  \bibinfo{person}{R.~Garnett}} (Eds.), Vol.~\bibinfo{volume}{32}.
  \bibinfo{publisher}{Curran Associates, Inc.}
\newblock
\showeprint[arxiv]{1911.01485}
\urldef\tempurl%
\url{https://proceedings.neurips.cc/paper_files/paper/2019/hash/201d546992726352471cfea6b0df0a48-Abstract.html}
\showURL{%
\tempurl}


\bibitem[Testoni and Calixto(2026)]%
        {testoni2026calibrated}
\bibfield{author}{\bibinfo{person}{Alberto Testoni} {and}
  \bibinfo{person}{Iacer Calixto}.} \bibinfo{year}{2026}\natexlab{}.
\newblock \showarticletitle{Calibrated? Not for Everyone: How Sexual
  Orientation and Religious Markers Distort {LLM} Accuracy and Confidence in
  Medical {QA}}. In \bibinfo{booktitle}{\emph{Proceedings of the 64th Annual
  Meeting of the {A}ssociation for {C}omputational {L}inguistics (Volume 2:
  Short Papers)}}, \bibfield{editor}{\bibinfo{person}{Maria Liakata},
  \bibinfo{person}{Viviane~P. Moreira}, \bibinfo{person}{Jiajun Zhang}, {and}
  \bibinfo{person}{David Jurgens}} (Eds.). \bibinfo{publisher}{Association for
  Computational Linguistics}, \bibinfo{address}{San Diego, California, United
  States}, \bibinfo{pages}{432--447}.
\newblock
\showISBNx{979-8-89176-391-3}
\href{https://doi.org/10.18653/v1/2026.acl-short.36}{doi:\nolinkurl{10.18653/v1/2026.acl-short.36}}


\bibitem[Tomar et~al\mbox{.}(2025)]%
        {tomar-etal-2025-bharatbbq}
\bibfield{author}{\bibinfo{person}{Aditya Tomar}, \bibinfo{person}{Nihar~Ranjan
  Sahoo}, {and} \bibinfo{person}{Pushpak Bhattacharyya}.}
  \bibinfo{year}{2025}\natexlab{}.
\newblock \showarticletitle{{B}harat{BBQ}: A Multilingual Bias Benchmark for
  Question Answering in the {I}ndian Context}.
\newblock \bibinfo{journal}{\emph{Transactions of the Association for
  Computational Linguistics}}  \bibinfo{volume}{13} (\bibinfo{year}{2025}),
  \bibinfo{pages}{1672--1692}.
\newblock
\href{https://doi.org/10.1162/tacl.a.55}{doi:\nolinkurl{10.1162/tacl.a.55}}


\bibitem[VanderWeele and Knol(2014)]%
        {vanderweele2014tutorial}
\bibfield{author}{\bibinfo{person}{Tyler~J. VanderWeele} {and}
  \bibinfo{person}{Mirjam~J. Knol}.} \bibinfo{year}{2014}\natexlab{}.
\newblock \showarticletitle{A Tutorial on Interaction}.
\newblock \bibinfo{journal}{\emph{Epidemiologic Methods}} \bibinfo{volume}{3},
  \bibinfo{number}{1} (\bibinfo{year}{2014}), \bibinfo{pages}{33--72}.
\newblock
\href{https://doi.org/10.1515/em-2013-0005}{doi:\nolinkurl{10.1515/em-2013-0005}}


\bibitem[Vaughan et~al\mbox{.}(2009)]%
        {vaughan2009use}
\bibfield{author}{\bibinfo{person}{Laura~K. Vaughan}, \bibinfo{person}{Jasmin
  Divers}, \bibinfo{person}{Miguel~A. Padilla}, \bibinfo{person}{David~T.
  Redden}, \bibinfo{person}{Hemant~K. Tiwari}, \bibinfo{person}{Daniel Pomp},
  {and} \bibinfo{person}{David~B. Allison}.} \bibinfo{year}{2009}\natexlab{}.
\newblock \showarticletitle{The use of plasmodes as a supplement to
  simulations: A simple example evaluating individual admixture estimation
  methodologies}.
\newblock \bibinfo{journal}{\emph{Computational Statistics \& Data Analysis}}
  \bibinfo{volume}{53}, \bibinfo{number}{5} (\bibinfo{year}{2009}),
  \bibinfo{pages}{1755--1766}.
\newblock
\href{https://doi.org/10.1016/j.csda.2008.02.032}{doi:\nolinkurl{10.1016/j.csda.2008.02.032}}


\bibitem[Vijayaraghavan et~al\mbox{.}(2025)]%
        {ijcai2025p1100}
\bibfield{author}{\bibinfo{person}{Prashanth Vijayaraghavan},
  \bibinfo{person}{Soroush Vosoughi}, \bibinfo{person}{Lamogha Chiazor},
  \bibinfo{person}{Raya Horesh}, \bibinfo{person}{Rogerio Abreu~de Paula},
  \bibinfo{person}{Ehsan Degan}, {and} \bibinfo{person}{Vandana Mukherjee}.}
  \bibinfo{year}{2025}\natexlab{}.
\newblock \showarticletitle{DECASTE: Unveiling Caste Stereotypes in Large
  Language Models Through Multi-Dimensional Bias Analysis}. In
  \bibinfo{booktitle}{\emph{Proceedings of the Thirty-Fourth International
  Joint Conference on Artificial Intelligence, {IJCAI-25}}},
  \bibfield{editor}{\bibinfo{person}{James Kwok}} (Ed.).
  \bibinfo{publisher}{International Joint Conferences on Artificial
  Intelligence Organization}, \bibinfo{pages}{9899--9907}.
\newblock
\href{https://doi.org/10.24963/ijcai.2025/1100}{doi:\nolinkurl{10.24963/ijcai.2025/1100}}
\newblock
\shownote{AI and Social Good}.


\bibitem[Wagenmakers et~al\mbox{.}(2012)]%
        {wagenmakers2012interpretation}
\bibfield{author}{\bibinfo{person}{Eric-Jan Wagenmakers},
  \bibinfo{person}{Angelos-Miltiadis Krypotos}, \bibinfo{person}{Amy~H. Criss},
  {and} \bibinfo{person}{Geoff Iverson}.} \bibinfo{year}{2012}\natexlab{}.
\newblock \showarticletitle{On the Interpretation of Removable Interactions: A
  Survey of the Field 33 Years after {Loftus}}.
\newblock \bibinfo{journal}{\emph{Memory \& Cognition}} \bibinfo{volume}{40},
  \bibinfo{number}{2} (\bibinfo{year}{2012}), \bibinfo{pages}{145--160}.
\newblock
\href{https://doi.org/10.3758/s13421-011-0158-0}{doi:\nolinkurl{10.3758/s13421-011-0158-0}}


\bibitem[Westfall and Young(1993)]%
        {westfall1993resampling}
\bibfield{author}{\bibinfo{person}{Peter~H. Westfall} {and}
  \bibinfo{person}{S.~Stanley Young}.} \bibinfo{year}{1993}\natexlab{}.
\newblock \bibinfo{booktitle}{\emph{Resampling-Based Multiple Testing: Examples
  and Methods for p-Value Adjustment}}.
\newblock \bibinfo{publisher}{John Wiley \& Sons}, \bibinfo{address}{New York}.
\newblock
\showISBNx{978-0-471-55761-6}


\bibitem[Wilson and Caliskan(2024)]%
        {3716662.3716799}
\bibfield{author}{\bibinfo{person}{Kyra Wilson} {and} \bibinfo{person}{Aylin
  Caliskan}.} \bibinfo{year}{2024}\natexlab{}.
\newblock \showarticletitle{Gender, Race, and Intersectional Bias in Resume
  Screening via Language Model Retrieval}. In
  \bibinfo{booktitle}{\emph{Proceedings of the AAAI/ACM Conference on AI,
  Ethics, and Society}}, Vol.~\bibinfo{volume}{7}. \bibinfo{pages}{1578--1590}.
\newblock
\href{https://doi.org/10.1609/aies.v7i1.31748}{doi:\nolinkurl{10.1609/aies.v7i1.31748}}


\bibitem[Winkler et~al\mbox{.}(2014)]%
        {winkler2014permutation}
\bibfield{author}{\bibinfo{person}{Anderson~M. Winkler},
  \bibinfo{person}{Gerard~R. Ridgway}, \bibinfo{person}{Matthew~A. Webster},
  \bibinfo{person}{Stephen~M. Smith}, {and} \bibinfo{person}{Thomas~E.
  Nichols}.} \bibinfo{year}{2014}\natexlab{}.
\newblock \showarticletitle{Permutation inference for the general linear
  model}.
\newblock \bibinfo{journal}{\emph{NeuroImage}}  \bibinfo{volume}{92}
  (\bibinfo{year}{2014}), \bibinfo{pages}{381--397}.
\newblock
\href{https://doi.org/10.1016/j.neuroimage.2014.01.060}{doi:\nolinkurl{10.1016/j.neuroimage.2014.01.060}}


\bibitem[Zhang and Neill(2016)]%
        {zhang2016identifying}
\bibfield{author}{\bibinfo{person}{Zhe Zhang} {and} \bibinfo{person}{Daniel~B.
  Neill}.} \bibinfo{year}{2016}\natexlab{}.
\newblock \bibinfo{title}{Identifying Significant Predictive Bias in
  Classifiers}.
\newblock
\showeprint[arxiv]{1611.08292}~[stat.ML]
\newblock
\shownote{NIPS 2016 Workshop on Interpretable Machine Learning in Complex
  Systems}.


\bibitem[Zhang et~al\mbox{.}(2026)]%
        {zhang2026comfair}
\bibfield{author}{\bibinfo{person}{Zheng Zhang}, \bibinfo{person}{Yifan Wu},
  \bibinfo{person}{Yuxiang Zhou}, \bibinfo{person}{Qi Liu},
  \bibinfo{person}{Qingyang Mao}, \bibinfo{person}{Ning Li},
  \bibinfo{person}{Weibo Gao}, \bibinfo{person}{Qiyuan Shi},
  \bibinfo{person}{Zhenya Huang}, {and} \bibinfo{person}{Baosheng Yu}.}
  \bibinfo{year}{2026}\natexlab{}.
\newblock \bibinfo{title}{Beyond Single-Attribute Auditing: A Configurable
  Multi-Attribute Benchmark for Intersectional Fairness in {LLMs}}.
\newblock
\showeprint[arxiv]{2610.35069}
\urldef\tempurl%
\url{https://arxiv.org/abs/2610.35069}
\showURL{%
\tempurl}


\end{thebibliography}
